\documentclass[10pt]{article}
\usepackage[dvipsnames]{xcolor}
\usepackage{bigdelim} %
\usepackage[preprint]{tmlr}

\usepackage[T1]{fontenc}
\usepackage{bm}

\usepackage{natbib}
\usepackage{cancel} %

\usepackage[pdfusetitle]{hyperref}
\usepackage{url}
\usepackage{doi}

\hypersetup{
	colorlinks,
	linkcolor={},
	citecolor={blue!45!black},
	urlcolor={blue!80!black}
}

\usepackage{booktabs}
\usepackage{threeparttable}
\usepackage{colortbl} %
\usepackage{multirow}
\usepackage{tabularx}
\usepackage{ifthen}

\newcolumntype{L}[1]{>{\ifthenelse{\equal{#1}{}}{}{\hsize=#1\hsize}\raggedright\arraybackslash}X}%
\newcolumntype{M}[1]{>{\ifthenelse{\equal{#1}{}}{}{\hsize=#1\hsize}\hsize=#1\hsize\centering\arraybackslash}X}%
\usepackage{longtable}

\usepackage{tocloft}
\usepackage{caption} %
\usepackage{subcaption} %

\usepackage{tikz}
\usetikzlibrary{arrows}
\usetikzlibrary{backgrounds}
\usetikzlibrary{calc}
\usetikzlibrary{fit}
\usetikzlibrary{positioning}

\let\classAND\AND{}
\let\AND\relax
\usepackage{algorithm,algorithmic}
{}
\let\AND\classAND{}
\AtBeginEnvironment{algorithmic}{\let\AND\algoAND}

\usepackage{amssymb} %
\usepackage{calc} %
\usepackage{pifont} %
\definecolor{cmarkcolor}{RGB}{0, 128, 0}
\definecolor{acmarkcolor}{RGB}{52, 152, 219}
\definecolor{xmarkcolor}{RGB}{200, 0, 0}
\newcommand{\cmark}{{\color{cmarkcolor}\ding{51}}} %
\newcommand{\acmark}{{\color{acmarkcolor}(\ding{51})}} %
\newcommand{\xmark}{{\color{xmarkcolor}\ding{55}}} %
\newcommand{\sxmark}{{\color{gray}\tiny\ding{55}}} %
\newcommand{\scmark}{{\color{black}\ding{51}}} %
\newcommand{\maybeany}{{\color{gray}\ding{58}}} %

\newcommand{\argmax}{\operatornamewithlimits{argmax}}
\newcommand{\St}{{\mathcal{S}}}
\newcommand{\Ac}{{\mathcal{A}}}
\newcommand{\T}{{P}} %
\newcommand{\R}{{R}} %
\newcommand{\traj}{\tau}

\usepackage{glossaries}
\newcommand{\glsintro}[1]{\glsreset{#1}\gls{#1}}
\newcommand{\Glsintro}[1]{\glsreset{#1}\Gls{#1}}
\glsdisablehyper{} %
\newacronym{RL}{RL}{reinforcement learning}
\newacronym{ML}{ML}{machine learning}
\newacronym{MLE}{MLE}{maximum likelihood estimate}
\newacronym{SGD}{SGD}{stochastic gradient descent}
\newacronym{MDP}{MDP}{Markov decision process}
\newacronym{RLHF}{RLHF}{reinforcement learning from human feedback}
\newacronym{DPO}{DPO}{direct policy optimization}
\newacronym{LLM}{LLM}{large language model}
\newacronym{RLAIF}{RLAIF}{reinforcement learning from AI feedback}
\newacronym{PbRL}{PbRL}{preference-based reinforcement learning}
\newacronym{LPbRL}{LPbRL}{logistic preference-based reinforcement learning}
\newacronym{SSRL}{SSRL}{semi-supervised reinforcement learning}
\newacronym{HCI}{HCI}{human-computer interaction}
\newacronym{NLP}{NLP}{natural language processing}
\newacronym{MLP}{MLP}{multilayer perceptron}

\usepackage[nameinlink,capitalize]{cleveref}

\newcommand{\theabstract}{%
	Reinforcement learning from human feedback (RLHF) is a variant of reinforcement learning (RL) that learns from human feedback instead of relying on an engineered reward function.
	Building on prior work on the related setting of preference-based reinforcement learning (PbRL), it stands at the intersection of artificial intelligence and human-computer interaction.
	This positioning provides a promising approach to enhance the performance and adaptability of intelligent systems while also improving the alignment of their objectives with human values.
	The success in training large language models (LLMs) has impressively demonstrated this potential in recent years, where RLHF has played a decisive role in directing the model's capabilities towards human objectives.
	This article provides an overview of the fundamentals of RLHF, exploring how RL agents interact with human feedback.
	While recent focus has been on RLHF for LLMs, our survey covers the technique across multiple domains.
	We provide our most comprehensive coverage in control and robotics, where many fundamental techniques originate, alongside a dedicated LLM section.
	We examine the core principles that underpin RLHF, how algorithms and human feedback work together, and the main research trends in the field.
	Our goal is to give researchers and practitioners a clear understanding of this rapidly growing field.
}
\hypersetup{pdfsubject={\theabstract}}

\def\month{06}  %
\def\year{2025} %
\def\openreview{\url{https://openreview.net/forum?id=f7OkIurx4b}} %

\newtoggle{final}
\toggletrue{final}
\iftoggle{final}{
	\usepackage[disable]{todonotes}

}{
	\setlength{\marginparwidth}{2cm}
	\usepackage[size=tiny]{todonotes}

	\usepackage{silence}
	\WarningFilter{latex}{Marginpar on page}
}
\begin{document}

\title{A Survey of Reinforcement Learning from Human Feedback}
\hypersetup{pdftitle={A Survey of Reinforcement Learning from Human Feedback}}

\author{%
	\name{}Timo Kaufmann \email{}timo.kaufmann@ifi.lmu.de \\
	\addr{}LMU Munich, MCML Munich
	\AND{}%
	\name{}Paul Weng \email{} paul.weng@dukekunshan.edu.cn \\
	\addr{}Digital Innovation Research Center, Duke Kunshan University
	\AND{}%
	\name{}Viktor Bengs\email{}viktor.bengs@dfki.de\\
	\addr{}German Research Center for Artificial Intelligence (DFKI)
	\AND{}%
	\name{}Eyke Hüllermeier \email{} eyke@lmu.de\\
	\addr{}LMU Munich, MCML Munich, DFKI Kaiserslautern
}
\hypersetup{pdfauthor={Timo Kaufmann, Paul Weng, Viktor Bengs, Eyke Hüllermeier}}
\maketitle
\begin{abstract}
	\noindent\theabstract{}
\end{abstract}
\newcommand{\kwA}{Preference-Based Reinforcement Learning}
\newcommand{\kwB}{Human Feedback}
\newcommand{\kwC}{Preference Learning}
\newcommand{\kwD}{Interactive Learning}
\newcommand{\kwE}{Reward Learning}
\newcommand{\kwF}{Reward Modeling}
\newcommand{\kwG}{Agent Alignment}
\newcommand{\kwH}{Human-in-the-Loop}
\newcommand{\kwI}{Human-Computer Interaction}
\newcommand{\kwJ}{Human-Compatible AI}

\setcounter{tocdepth}{2} %
\tableofcontents
\hypersetup{linkcolor=red!40!black}
\glsresetall{}

\clearpage

\section{Introduction}\label{sec:introduction}

In \gls{RL}, an agent navigates an environment and learns through trial and error.
The optimality of these decisions is determined solely by reward signals, which must be carefully designed based on performance measurements to ensure the agent learns the desired behavior.
However, designing effective reward functions presents significant challenges.
In real-world applications, success is often difficult to formally define and measure.
Moreover, sparse success signals are not well suited for agent learning, often necessitating \emph{reward shaping}~\citep{ng1999policy}, where the reward signal is transformed into one that is more suitable for learning.
Unfortunately, this transformation can introduce spurious correlations -- rewarding behaviors that correlate with desired outcomes without directly achieving the true objective.
This can lead to \emph{reward hacking}~\citep{skalse2022defining}, where agents exploit such correlations to maximize rewards while producing undesired outcomes.

\Gls{RLHF} has emerged as a practical solution to these challenges by incorporating human feedback directly into the learning process.
Unlike traditional \gls{RL}, where objectives are fixed a priori, \gls{RLHF} allows humans to define and iteratively refine objectives during training.
This approach addresses the limitations of classical \gls{RL} while promoting better alignment between agent behavior and human values and objectives, enabling the development of more ethically sound, socially responsible, and practically useful AI systems.

\Gls{RLHF} has seen a number of successful applications, methodological advances, and theoretical insights since the last comparable survey~\citep{wirth2017survey}.
The applications span various domains, including \gls{LLM} fine-tuning~\citep{openai2022introducing}, image generation~\citep{lee2023aligning},
music generation~\citep{cideron2024musicrl},
continuous control~\citep{christiano2017deep}, games~\citep{ibarz2018reward}, and robotics~\citep{hejna2022fewshot}.
Alongside these applications, significant methodological advances have emerged.
These include
fusing multiple feedback types to leverage their relative strengths (\cref{sec:combination-of-feedback-types}),
enhancing query efficiency through active learning and query synthesis (\cref{sec:labelcollection-activelearning}),
incorporating psychological insights to improve feedback quality (\cref{sec:labeling-psychologyaware}),
using techniques such as meta-learning to quickly adapt learned preferences to new tasks using prior data (\cref{sec:prior-data}),
and using available preference data more efficiently through approaches such as data augmentation and semi-supervised learning (\cref{sec:using-data-more-efficiently}).
Finally, novel theoretical results (\cref{sec:theory}) have provided insights as well as new questions about the fundamental mathematical problems underlying \gls{RLHF}.

While we focus on the core principles underlying \gls{RLHF}, \gls{LLM} fine-tuning requires additional considerations due to the unique characteristics of \glspl{LLM} and their operational scale.
We address these considerations without delving deeply into \gls{LLM}-specific details.
Notably, applying \gls{RLHF} to \glspl{LLM} is not our primary focus.
While \gls{RLHF} for \glspl{LLM} remains in scope, we survey \gls{RLHF} broadly, examining techniques and applications across domains with particular attention to control applications.
Nonetheless, many techniques and insights discussed herein apply to \glspl{LLM}, and we address some unique aspects of \gls{LLM} fine-tuning.
For readers primarily interested in \gls{LLM} fine-tuning, \cref{sec:rlhf-for-llms} provides an \gls{LLM}-focused introduction that complements this survey and guides the application of the insights presented here to \gls{LLM} fine-tuning.
Note that the literature focused on \gls{RLHF} for \glspl{LLM} in particular is vast and rapidly growing, and we cannot provide a comprehensive overview of this area.

This survey therefore provides an overview of current \gls{RLHF} research, classifies existing approaches, describes their main characteristics, and briefly surveys application areas.
In the remainder of this section, we begin by examining the motivation (\cref{sec:introduction-whyhumanfeedback}) and origins (\cref{sec:origins-of-rlhf}) of \gls{RLHF}, defining the scope of this survey (\cref{sec:scope-of-the-survey}), and outlining the structure of subsequent sections (\cref{sec:introduction-outline}).

\subsection{Why Human Feedback?}\label{sec:introduction-whyhumanfeedback}

In conventional \gls{RL}, the agent's objective is defined by a reward function that it aims to maximize~\citep{sutton2018reinforcement}.
Specifying this reward function can be challenging, particularly in complex domains:
What would be a suitable reward function for a robot assisting humans in a household environment or for autonomous vehicles navigating through a busy urban environment?
Moreover, even reward functions that initially seem well-defined can lead to unexpected behaviors due to distributional shifts or over-optimization, raising practical and safety concerns.
Learning the agent's objective from human feedback circumvents reward engineering challenges and fosters robust training, with the reward function being dynamically refined and adjusted to distributional shifts as the agent learns.

\paragraph{Interactive Feedback vs.\ Demonstrations}

The field of inverse \gls{RL} aims to infer reward functions from human demonstrations~\citep{arora2021survey}.
While this can partially resolve reward engineering challenges, it faces inherent difficulties:
(i) it is generally not possible to robustly identify rewards from demonstrations~\citep{cao2021identifiability,mindermann2018occam},
(ii) it is only applicable in scenarios where good demonstrations can be obtained,
(iii) it struggles to outperform the demonstrator, and
(iv) humans often do not demonstrate the behavior they would prefer a machine to adopt~\citep{basu2017you}.
Interactive feedback, by contrast, can
use active queries to differentiate between aspects relevant or irrelevant to the human preference,
is much easier to provide than demonstrations,
does not require near-optimal performance from human evaluators, and
elicits preferences about the behavior that a human would prefer from the machine.
Interactive feedback also complements demonstrations effectively, refining capabilities learned through initial training methods like behavioral cloning and preventing overfitting to demonstrated behavior~\citep{abramson2022improving}.

\paragraph{Avoiding Reward Engineering}

Reward engineering in \gls{RL} presents significant challenges, as accurately specifying reward functions is notoriously difficult~\citep{amodei2016concrete,knox2023reward}.
These challenges can be mitigated by using human feedback, which enables training agents for tasks that are hard to define manually and represents a step towards addressing safety issues arising from misaligned rewards~\citep{skalse2022defining}.
Safety issues related to a misalignment between the agent's and the human's objectives are studied as the AI alignment problem~\citep{gabriel2020artificial}, in particular agent alignment and value alignment~\citep{kirchner2022researching}.

Excessive optimization of poorly specified rewards frequently produces unintended behaviors.
Examples of such behaviors include exploiting flaws in the environment simulation for higher rewards~\citep{lehman2020surprising,baker2020emergent} or engaging in more general \emph{reward hacking}~\citep{skalse2022defining}, where the behavior maximizes the specified reward but deviates from the intended objective.
This is evident in cases where agents focus on intermediate rewards without achieving the actual goal~\citep{clark2016faulty} or prematurely exit games to avoid negative rewards~\citep{saunders2018trial}.
The root of these issues is that the reward function does not properly reflect the actual learning task.
While such problems appear trivial in game environments, they become critical in safety-critical contexts such as healthcare and autonomous driving, where misaligned rewards could lead to care robots causing injuries or self-driving cars compromising road safety.

\Gls{RLHF} presents a promising approach to enhance alignment~\citep{leike2018scalable} by enabling agents to learn from human feedback, which is often more closely aligned with the true objective than manually specified rewards.
Nonetheless, the effectiveness of \gls{RLHF} in resolving these alignment issues is debated~\citep{christiano2023thoughts}.
Examples of possible pitfalls raised in this debate are that an agent may be incentivized to manipulate the human teacher to provide feedback that is easier to optimize~\citep{armstrong2020pitfalls,carroll2023characterizing} or that the agent may learn to exploit errors in human judgment~\citep{ngo2024alignment}.
We refer the interested reader to the survey by \citet{casper2023open} for a more detailed discussion of these issues.
Despite these concerns, \gls{RLHF}
represents an important early step towards aligning agents with human values and serves as a foundation to build on to further improve agent alignment.

\subsection{The Origins of RLHF}\label{sec:origins-of-rlhf}

Learning behavior from human feedback has long been studied as a subfield of \gls{RL}, but methods and terminology have evolved over time.
Early methods focused on learning directly from human rewards~\citep{knox2012learning,isbell2001social,knox2009interactively}, from action advice \citep{maclin2005giving}, or from action critique \citep{judah2010reinforcement}.
Notable examples include TAMER~\citep{knox2009interactively,warnell2018deep}, which interprets human feedback as samples of the optimal action-value function, and COACH~\citep{macglashan2017interactive,arumugam2019deep}, which interprets human feedback in a policy-dependent way, i.e., as samples of the advantage function.
This survey, however, focuses on more indirect approaches to inferring the objective from human feedback.

Modern \glsintro{RLHF} originates from \glsintro{PbRL}, independently introduced by \citet{akrour2011preferencebased} and \citet{cheng2011preferencebased}.
\Gls{PbRL} infers the objective from qualitative feedback, such as pairwise preferences between behaviors or between actions given states, instead of quantitative feedback in the form of numerical rewards.
The term \gls{RLHF} was later coined as an alternative~\citep{askell2021general,ouyang2022training,openai2022introducing},
 competing with the short-lived term reinforcement learning from human preferences (RLHP)~\citep{menick2022teaching},
though initially referring to the same concept of learning behavior from relative feedback.

\begin{table}
	\center{}
	\caption{Feedback types classified as belonging to PbRL, SSRL, and RLHF as defined in this survey.}\label{tbl:terminology}
	\begin{tabular}{lccc}
		\toprule
		Feedback Type & PbRL & SSRL & RLHF \\
		\midrule
		Binary trajectory comparisons & \cmark{} & \xmark{} & \cmark{} \\
		Trajectory rankings & \cmark{} & \xmark{} & \cmark{} \\
		State preferences & \cmark{} & \xmark{} & \cmark{} \\
		Action preferences & \cmark{} & \xmark{} & \cmark{} \\
		Binary critique & \xmark{} & \cmark{} & \cmark{} \\
		Scalar feedback & \xmark{} & \cmark{} & \cmark{} \\
		Corrections & \xmark{} & \xmark{} & \cmark{} \\
		Action advice & \xmark{} & \xmark{} & \cmark{} \\
		Implicit feedback & \xmark{} & \xmark{} & \cmark{} \\
		Natural language & \xmark{} & \xmark{} & \cmark{} \\
		\bottomrule
	\end{tabular}
\end{table}
Disentangling \gls{PbRL} and \gls{RLHF} is challenging due to their overlapping usage in the literature.
For instance, \citet{christiano2017deep} themselves are using the term \gls{PbRL}, yet are often cited as a seminal reference for \gls{RLHF}~\citep{daniels-koch2022expertise,ouyang2022training}.
This demonstrates that the terms are often used interchangeably.
Practically, \gls{RLHF} is often associated with reward modeling and deep \gls{RL}, while \gls{PbRL} is often linked to direct policy optimization in traditional \gls{RL} settings.
This is underlined by \citet{jeon2020rewardrational}, who characterize \gls{PbRL} as limited to direct policy learning from preferences.
This is in contrast with other sources, however, who include reward learning within the scope of \gls{PbRL}~\citep{christiano2017deep,wirth2017survey}.
Additionally, \gls{PbRL} is rarely used in the language modeling domain, while \gls{RLHF} is common in both language modeling and classical \gls{RL} domains such as continuous control.

Despite overlapping and sometimes conflicting usage, in this work we view \gls{RLHF} as a generalization of \gls{PbRL}.
While both involve human feedback to define \gls{RL} objectives, \gls{PbRL} primarily focuses on relative feedback, such as binary comparisons and rankings.
\Gls{RLHF} not only includes these aspects but also extends to a wider range of feedback types~\citep{metz2023rlhfblender,yuan2024unirlhf}.
\Cref{tbl:terminology} gives an exemplary overview of our interpretation of these terms.

Another concept, \glsintro{SSRL}, introduced by \citet{christiano2016semisupervised} and discussed by \citet{amodei2016concrete}, refers to an \gls{RL} setting where an agent receives feedback on a subset of its experiences.
The initial discussions of \gls{SSRL} focused on absolute feedback on subsets of the agent's experiences, making the concept complementary to \gls{PbRL}.
In contrast to \gls{PbRL} and \gls{RLHF}, the term \gls{SSRL} seems to be used less in the recent literature.

We adopt the view that \gls{RLHF} broadly encompasses all approaches using human feedback to define \gls{RL} objectives, including both \gls{PbRL} and \gls{SSRL}.
As the definitions and distinctions between these terms are not universally agreed upon, these distinctions are based on our interpretation of the current predominant usage of these terms in the literature.

\subsection{Scope of the Survey}\label{sec:scope-of-the-survey}

This section outlines the criteria guiding our selection of \gls{RLHF} approaches we cover.
We concentrate on methods where a learned reward model serves as the sole source of objective information, learned through interactive, online, scalable, and asynchronous human feedback.
We describe each of these criteria in more detail below.
\begin{description}
	\item[Reward Modeling]
	We focus on approaches that learn a reward model from human feedback and then use this model to train a policy.
	Although it is possible to directly optimize a policy from human feedback~\citep{wirth2017survey}, thereby performing \gls{RLHF} without reward learning, this approach was rarely practiced for a long time and has only recently gained renewed interest, especially in the domain of language model fine-tuning (see \cref{subsec:direct-methods}).
	The decomposition into reward learning and policy training offers many conceptual and practical benefits.
	These include the direct applicability of supervised learning techniques for the reward model and the ability to evaluate the reward model in isolation.
	Additionally, the decomposition naturally leads to a form of semi-supervised learning, enabling the agent to use labeled episodes for reward model training while leveraging unlabeled episodes to refine its behavior and explore the environment.
	\item[Human Defined]
	While many approaches include humans in the \gls{RL} loop, we focus on approaches where human feedback is the only source of truth about the objective.
	This excludes approaches to reward shaping, feature engineering, and other forms of human guidance that are supplementary to a given objective.
	\item[Interactive and Online]
	Our scope encompasses methods that collect human feedback interactively during the learning process, rather than relying solely on pre-collected demonstrations.
	This excludes pure imitation learning, learning from demonstration, and standalone inverse \gls{RL} approaches.
	However, hybrid methods that combine inverse \gls{RL} with interactive reward refinement fall within our scope, as discussed in \cref{sec:feedback-initializations,par:reward-model-initialization}.
	\item[Scalable and Asynchronous]
	We consider methods that incorporate human feedback without requiring synchronous human-agent interaction.
	The agent must be capable of continuing its learning process while awaiting human input, and humans need not provide continuous supervision.
	This distinguishes \gls{RLHF} from more direct methods of incorporating a human into the \gls{RL} loop, and we believe that this is key for practicality and efficiency.
\end{description}

Beyond these methodological criteria, we primarily cover works from 2017 through 2024\footnote{%
We updated preprints to their peer-reviewed versions where applicable, leading to references outside this window.%
}, building upon the comprehensive treatment of earlier works by \citet{wirth2017survey}.
We selectively revisit earlier foundational contributions that continue to inform current practice or have fundamentally shaped the field.

While \gls{RLHF} has recently gained prominence through \gls{LLM} applications, this survey adopts a broader perspective, examining \gls{RLHF} across multiple domains with particular depth in control and robotics.
We dedicate \cref{sec:rlhf-for-llms} to \glspl{LLM}, addressing domain-specific challenges including large action spaces, distribution shift, and KL regularization.
Many techniques from control domains transfer naturally to the \gls{LLM} setting, which is frequently constrained to a single turn, though not all techniques are equally applicable.
Given the rapidly evolving nature of \gls{LLM} research, our coverage of techniques specific to this domain is less exhaustive than our treatment of control and robotics applications.

This survey can be seen as the canonical continuation of \citet{wirth2017survey}, examining the evolution from \gls{PbRL} to \gls{RLHF} and the accompanying methodological advances.
We provide both a thorough description of the basics and an in-depth discussion of current advances and trends.
We refer to \cref{sec:prior-surveys} for an overview of prior and related surveys and their relation to this work.

\subsection{Outline}\label{sec:introduction-outline}

In the next section, we begin with an introduction to the basics by revisiting the most important concepts from the standard \gls{RL} setting, which are also naturally important in \gls{RLHF} (\cref{sec:preliminaries}).
We then dive into the \gls{RLHF} topic by outlining the most studied scenario of reward model learning from pairwise preferences.
Using this introductory and illustrative example scenario, we explain the basic framework of \gls{RLHF} alongside its three main components of (human) feedback, label collection (feedback acquisition), and reward model learning.
These three main components will essentially form the structure of our survey.
In \cref{sec:feedback}, we turn our attention to the human feedback component and provide an overview of the different types of feedback as well as their key attributes.
The important concepts in terms of label collection are then explained in \cref{sec:labelcollection}, followed by learning the reward model in \cref{sec:reward_learning}.
We also discuss policy learning in \cref{sec:policy-learning}, since some \gls{RLHF} methods adapt standard \gls{RL} training methods to the \gls{RLHF} setting.
\Cref{sec:theory} is devoted to an overview of recent progress on the theoretical side of \gls{RLHF}, including approaches involving a theoretical guarantee, as well as theoretical insights into the relationship between standard \gls{RL} and \gls{RLHF}.
Finally, \cref{sec:application} highlights some interesting practical applications of \gls{RLHF} and the existing benchmarks before \cref{sec:conclusion} concludes the survey by pointing out some possible avenues for future work.

\section{Preliminaries}\label{sec:preliminaries}

In this section, we review the basic setting and the most important concepts of \gls{RL} and \gls{RLHF}.
In the course of this review, we will establish the notation that will be used throughout the survey.
We first introduce what is probably the most studied \gls{RLHF} scenario, i.e., learning a reward model from binary trajectory comparisons.
Based on this introductory and illustrative example scenario, we explain the basic framework of \gls{RLHF} with its main components and briefly discuss the respective roles of these components in the learning process.
We will also briefly touch on active learning, which strongly connects to the feedback collection component.

\paragraph{Notation}
For any integer $n \in \mathbb{N}$, we denote by $[n]$ the set $\{1, 2, \ldots, n\}$.
For any set $S$, $\Delta(S)$ denotes the set of probability distributions over $S$.
We use $\mathbb{P}(E)$ for denoting the probability of some event $E$, while $\mathbb{E}[X]$ is used to denote the expected value of a random variable $X$.
In some cases, we will write $\mathbb{E}_P[\cdot]$ or similar variants to emphasize that the distribution for the expected value is governed by the probability distribution $P \in \Delta(S)$.
Moreover, we will write $X \sim P$ if a random variable $X$ is distributed according to a probability distribution $P$.

\subsection{Reinforcement Learning}\label{sec:preliminaries-rl}

\Glsintro{RL}~\citep{sutton2018reinforcement} is the setting of learning behavior from rewarded interaction with an environment.
Such a learning environment is formalized as an \gls{MDP}, which is a model for sequential decision-making.
In an \gls{MDP}, an agent iteratively observes its current state, takes an action that causes the transition to a new state, and finally receives a reward that depends on the action's effectiveness.
Formally, an \gls{MDP} is defined as a tuple $(\St, \Ac, \T, \R, d_0, \gamma)$ where
\begin{itemize}
	\item $\St$ is a set of states (the \emph{state space}),
	\item $\Ac$ is a set of actions (the \emph{action space}),
	\item $\T : \St \times \Ac \to \Delta(\St)$ is a transition function (the \emph{transition dynamics}),
	\item $\R : \St \times \Ac \to \mathbb R$ is a reward function,
	\item $d_0 \in \Delta(\St)$ is a distribution over initial states,
	\item and $\gamma \in [0, 1]$ is a discount factor.
\end{itemize}
The transition function $\T$ defines the dynamics of the environment:
For any state $s$ and action $a$, the value $\T(s, a)(s')$, also sometimes denoted $\T(s' \mid s, a)$, is the probability of reaching the state $s'$ after executing $a$ in $s$.
In light of this, we will also sometimes refer to the transition function simply as the \emph{transition dynamics.}
For a given state and action, the transition is conditionally independent of all previous states and actions, which is known as the \emph{Markov property} and the reason for the naming as an \gls{MDP}.
The value $\R(s, a) \in \mathbb R$ provides an immediate evaluation after performing action $a$ in state $s$, which is also called the (instantaneous) reward.
It is also possible that the instantaneous reward is $0$ for some states, and one only receives a reward in specific states, for example, in so-called \emph{terminal} states for which the transition function is zero.
When both the state space $\St$ and the action space $\Ac$ are finite, we call the \gls{MDP} a \emph{tabular} \gls{MDP}.

In an \gls{MDP}, an \emph{$H$-step trajectory} $\traj$ is a sequence of $H \in \mathbb{N} \setminus \{0\}$ state-action pairs ending in a terminal state.
Formally, it is given by $\traj = (s_0, a_0, s_1, a_1, \ldots, s_H)$.
A trajectory $\tau$'s \emph{return} $\R(\traj)$ is the accumulated (discounted) rewards collected along this trajectory:
\begin{align}\label{def:return}
	\R(\traj) = \sum_{h=0}^{H-1} \gamma^h \R(s_h, a_h)
	\,\text.
\end{align}
Note that we here use the same notation for the return and the reward function; however, both have different signatures (trajectory vs.\ state-action pair).
We can also define the return $\R(\sigma)$ of a segment $\sigma$ in a similar manner.
The return is well defined even if the horizon $H$ is infinite as long as \( \gamma < 1 \).
If the \gls{MDP} is a tabular \gls{MDP} and any trajectory has finite length, i.e., $H$ is necessarily finite, we call the \gls{MDP} \emph{finite} and otherwise \emph{infinite}.

\begin{figure}
	\begin{subfigure}[T]{0.5\textwidth}
		\center{}
		\definecolor{bgcolor}{HTML}{3c8e4b}
\begin{tikzpicture}[%
	every text node part/.style={align=center},%
	>=stealth'%
]
	\node[inner sep=0,outer sep=0,text=white] (environment-title) {Environment};
	\node[rectangle,minimum width=2.2cm,below=0.3cm of environment-title,fill=white,text=black] (dynamics) {\small Dynamics};
	\node[,rectangle,minimum width=2.2cm,below=0.2cm of dynamics,fill=white,text=black] (objective) {\small Objective};
	\begin{scope}[on background layer]
		\node[rectangle,rounded corners,fit={(environment-title) (dynamics) (objective)}, inner ysep=0.2cm,fill=bgcolor] (environment) {};
	\end{scope}

	\node[rectangle,text depth=2.06cm,inner ysep=0.2cm,minimum width=1.2cm,rounded corners,left=2.5cm of environment.north west,anchor=north east,fill=bgcolor,text=white] (agent) {Agent};
	\draw[thick,->] (agent.east |- environment.155) to node[above=-0.05cm,pos=0.04,anchor=south west] {\footnotesize Action \( a_t \)} (environment.155);
	\draw[thick,->] (dynamics) to node[below=-0.05cm,pos=0.05,anchor=north east] {\footnotesize State \( s_{t+1} \)} (agent.east |- dynamics);
	\draw[thick,->] (objective) to node[below=-0.05cm,pos=0.05,anchor=north east] {\footnotesize Reward \( r_{t + 1} \)} (agent.east |- objective);
\end{tikzpicture}
		\caption{The standard \gls{RL} setting.}\label{fig:rl-loop}
	\end{subfigure}
	\begin{subfigure}[T]{0.5\textwidth}
		\center{}
		\definecolor{bgcolor}{HTML}{3c8e4b}
\begin{tikzpicture}[%
	every text node part/.style={align=center},%
	>=stealth'%
]
	\node[inner sep=0,outer sep=0,text=white] (agent-title) {Agent};
	\node[rectangle,minimum width=3.50cm,below=0.3cm of agent-title,fill=white,text=black] (policy) {\small Policy};
	\node[rectangle,minimum width=3.50cm,below=0.7cm of policy,fill=white,text=black] (rewardmodel) {\small Reward Model};
	\begin{scope}[on background layer]
		\node[rectangle,rounded corners,fit={(agent-title) (policy) (rewardmodel)}, inner ysep=0.2cm,fill=bgcolor] (agent) {};
	\end{scope}

	\node[inner sep=0,outer sep=0,text=white,right=3.5cm of agent-title] (environment-title) {Environment};
	\node[rectangle,minimum width=2.2cm,below=0.3cm of environment-title,fill=white,text=black] (dynamics) {\small Dynamics};
	\begin{scope}[on background layer]
		\node[rectangle,rounded corners,fit={(environment-title) (dynamics)}, inner ysep=0.2cm,fill=bgcolor] (environment) {};
	\end{scope}
	\draw[thick,->] (agent.east |- environment.180) to node[above=-0.05cm,pos=0.04,anchor=south west] {\footnotesize Action \( a_t \)} (environment.180);

	\coordinate (midway) at ($(environment.west |- policy) + (-1.45,0)$);
	\draw[thick,-|] (environment.west |- policy) -- (midway) node[below=-0.05cm,pos=0.46] {\footnotesize State \( s_{t+1} \)}; %
	\draw[thick,->] (midway) -- (policy -| agent.east) node[below=-0.05cm,pos=0.53] {\phantom{S}\footnotesize \( s_t \)};

	\draw[thick,->] (rewardmodel.70) to node[right=-0.05cm] {\color{white}\footnotesize Reward \( \hat r_{t + 1} \)} (policy.south -| rewardmodel.70);
	\draw[thick,->] (policy.south -| rewardmodel.110) to node[left=-0.05cm] {\color{white}\footnotesize Action \( a_t \)} (rewardmodel.110);

	\node[rectangle,minimum height=0.7cm,minimum width=2cm,rounded corners,below=1.15cm of environment.south,anchor=south,fill=bgcolor,text=white] (evaluator) {Labeler};

	\draw[thick,->] (rewardmodel.4) to[->] node[above=-0.05cm,midway] {\footnotesize Query \( q_i \)} (evaluator.west |- rewardmodel.4);
	\draw[thick,->] (evaluator.west |- rewardmodel.-4) to[->] node[below=-0.05cm,midway] {\footnotesize Label {\( l_i \)}} (rewardmodel.-4);
\end{tikzpicture}
		\caption{\gls{RLHF} with reward modeling.}\label{fig:reward-modeling}
	\end{subfigure}
	\caption{%
		Contrasting the standard \gls{RL} setting with \gls{RLHF} in its most common formulation, using a reward model.
		In each step, the policy commits to an action \( a_t \) and receives the next state \( s_{t + 1} \) and either the true reward \( r_{t + 1} \) or an estimate \( \hat r_{t + 1} \) in return.
		In contrast to the standard \gls{RL} setting, the true reward function is not known in the \gls{RLHF} setting but instead learned from human feedback.
		This reward learning process is decoupled from policy learning and can happen fully asynchronously.
		The dataset consists of a set of queries \( q_i \) (e.g., pairs of trajectory fragments) and their labels \( l_i \) (e.g., a preference for one of the fragments).
	}\label{fig:rl-rlhf-contrast}
\end{figure}

A \emph{policy} specifies how to select actions in a state, either deterministically or stochastically.
In the former case, a policy is simply a mapping $\pi: \St \to \Ac$ from states to actions, while in the latter, it is a mapping $\pi: \St \to \Delta(\Ac)$ from states to probability distributions over actions.
Since the deterministic case is a special case of the stochastic one, we assume the latter case in the following.

The basic interaction loop is depicted in \cref{fig:rl-loop}:
The agent chooses an action \( a_t \sim \pi(s_t) \) based on its policy and the current state.
As a consequence, the environment transitions to the new state \( s_{t + 1} \sim \T(s_t, a_t) \), governed by the transition dynamics.
The agent observes this new state and the reward \( r_{t + 1} = R(s_t, a_t) \), after which the interaction cycle repeats.

In this setting, the \gls{RL} agent aims at learning a policy that maximizes the expected return
\[
	J(\pi) = \mathbb E_{d_0, \T, \pi}[\R(\traj)]
	\,\text,
\]
where the expectation is with respect to policy $\pi$, transition function $\T$, and initial distribution $d_0$.
To solve this problem, two families of \gls{RL} approaches have been considered:
\emph{model-based} \gls{RL} and \emph{model-free} \gls{RL}.
The methods in the first family learn a model (i.e., $\T, \R$) of the underlying \gls{MDP} to help solve the \gls{RL} problem, while the methods in the second directly try to obtain a good policy without learning an \gls{MDP} model.
The second family can be further decomposed into two main categories: \emph{value-based} methods and \emph{policy search} methods.
In deep \gls{RL}, both value functions and policies are approximated with neural networks.

Value-based methods (e.g., DQN and its variants~\citep{mnih2015humanlevel,hessel2018rainbow}) aim at learning the $Q$-function $Q^*$ of an optimal policy.
The $Q$-function of a policy $\pi$ is defined by:
\begin{align*}
	Q_\pi(s, a) = \mathbb E_{\T, \pi}\left[\sum_{h=0}^{H-1} \gamma^h \R(s_h, a_h)\right]
	\,\text,
\end{align*}
where $s_0 = s$ and $a_0 = a$, and in the expectation,
$a_h \sim \pi(\cdot \mid s_h)$ as well as $s_{h} \sim \T(\cdot \mid s_{h-1}, a_{h-1})$ for $h \in [H-1]$.
A policy can be naturally designed from a $Q$-function by choosing an action in a greedy manner in each state: $\pi(s) = \arg\max_a Q(s, a)$.
Note that for a deterministic optimal policy $\pi^*$ it holds that $J(\pi^*) = \mathbb E_{d_0}[Q^*(s, \pi^*(s))]$.

Similar to the action-value function $Q$, we can also define the state-value function
\begin{align*}
	V_\pi(s ) =  \mathbb E_{\T, \pi}\left[\sum_{h=0}^{H-1} \gamma^h \R(s_h, a_h) \, | \, s_0 = s \right]
	\,\text.
\end{align*}
Its value for some state $s$ is the expected return when starting in that state and then always using the policy $\pi$.
It is related to the $Q$-function by means of
\[ V_\pi(s) = \mathbb{E}_{a \sim \pi(s)}\left[Q_\pi(s,a)\right] \]
for any state $s \in \St$.

In contrast, policy search methods directly aim at finding a good policy in some parametrized policy space.
The most data-efficient algorithms in this class of methods follow an actor-critic scheme where both an actor (i.e., a policy) and a critic (i.e., usually its $Q$-value function) are learned at the same time.
Typical representative methods here are PPO~\citep{schulman2017proximal}, TD3~\citep{fujimoto2018addressing}, or SAC~\citep{haarnoja2018soft}.

\Gls{RL} algorithms can further be classified as either \emph{on-policy} or \emph{off-policy}.
In an on-policy algorithm, such as PPO, only the recently generated transitions are used for training.
In contrast, in an off-policy algorithm, such as DQN (or its variants), TD3, or SAC, the agent can be updated with transitions not necessarily generated by its current policy.
While on-policy training is usually more stable, off-policy training enables more data-efficient learning by reusing samples from a replay buffer that stores past transitions.

\subsection{Preference-Based MDPs}

In contrast to standard \gls{RL} as described in the previous section, \gls{RLHF} does not assume that a reward signal is available.
It instead assumes the existence of an \emph{oracle} (e.g., \emph{human labeler}) that can provide information about the reward in a specific indirect manner.
More precisely, in \gls{RLHF}, the agent can make queries $q_i$ to the oracle, which in practice means asking for human feedback, and in response, the agent receives a label $l_i$, which in general gives a hint about the reward.
In principle, the query can be made asynchronously to the actual conventional \gls{RL} cycle.
See \cref{fig:reward-modeling} for an illustration.

In the most common setting, the oracle can compare two (segments of) trajectories, but various other cases have been considered, as we shall see later on.
For the former case, \gls{RLHF} is based on the setting of preference-based \glspl{MDP}~\citep{gilbert2017optimizing,wirth2017survey}, which can be defined as an \gls{MDP} model without reward function, but where comparisons of trajectories are available.

\subsection{Reward Learning}\label{subsec:reward_learning_basic}

\Gls{RLHF} approaches can be divided into two categories, depending on whether a utility-based approach is used for reward modeling or whether an alternative criterion that is detached from a utility concept is used~\citep{gilbert2016modelfree,gilbert2016quantile,wirth2017survey}.
Most works fall into the first category, on which this overview focuses.
Such approaches assume a human-dependent utility function that can be used as a reward function in order to apply standard \gls{RL} methods.
Next, we will describe the commonly used approach for reward learning for the common setting of binary trajectory comparisons.

The prevalent approach to learning a utility function from observations of pairwise comparisons is based on the Bradley-Terry model~\citep{bradley1952rank}, which stipulates a probabilistic model for the oracle (human labeler):
\begin{align*}
	\mathbb P(\traj_1 \succ \traj_2) = \frac{1}{1 + \exp(\R(\traj_2) - \R(\traj_1))}
	\,\text,
\end{align*}
where $\succ$ means ``preferred to'' and $\R(\traj)$ corresponds to the utility (i.e., return in the context of \gls{RL}) of $\traj$.
Note that this utility function is a kind of surrogate function for the true reward function, which is (tacitly) assumed to induce the same optimal policy as the true reward function.
For a given data set $\mathcal D = \{\traj^i_1 \succ \traj^i_2 \mid i \in [N]\}$, a utility function $\R_\psi$ parameterized by $\psi$ can then be learned by the maximum likelihood principle (or equivalently using a cross-entropy loss):
\begin{align}\label{eq:ml}
	\max_\psi \prod_{i=1}^N \frac{1}{1 + \exp(\R_\psi(\traj^i_2) - \R_\psi(\traj^i_1))}
	\,\text.
\end{align}
In the context of \gls{RL}, since $\R_\psi(\traj) = \sum_{h=0}^{H-1} \gamma^h R_\psi(s_h, a_h)$,~\eqref{eq:ml} can then directly be used to train a function approximator (e.g., single or ensemble of neural networks) to approximate $R$.

This approach accommodates the case of a noisy or unreliable oracle, in which case the Bradley-Terry model can be understood as the generative model of the oracle's answers (or labels provided by the human labeler).
When the oracle is reliable, more direct methods based on preference elicitation for recovering the reward function have been studied~\citep{regan2009regretbased,regan2011robust,weng2013interactive,gilbert2015reducing,sadigh2017active,wilde2018learning}.
In this survey, we will focus on the general case where the oracle may be noisy.

Note that, in contrast to the typical way of preference learning, the learned reward function is used to train an \gls{RL} agent and not directly to compare trajectories.
This discrepancy between the objective function used in reward learning and how the learned rewards are actually used may lead to suboptimal policies~\citep{lindner2021information}.

\subsection{Reinforcement Learning from Human Feedback}\label{sec:RLHF_setting_intro}

In the \gls{RLHF} setting as illustrated in \cref{fig:reward-modeling}, the learning agent needs to solve an \gls{RL} task without having access to a reward function.
To this end, the agent usually simultaneously learns an approximation of the reward function (via the assumed utility function) and an \gls{RL} policy.
Therefore, a generic \gls{RLHF} algorithm consists of repeating two phases: (1) reward learning and (2) \gls{RL} training.
The first phase can itself be decomposed into two main steps:
(i) generate queries to ask the oracle,
(ii) train a reward function approximator with the answers provided by the oracle.
The \gls{RL} training part is more conventional and is usually directly based on running a deep \gls{RL} algorithm using the currently trained reward function approximator.

\begin{algorithm}
	\caption{Generic RLHF Algorithm in an Actor-Critic Scheme.}\label{alg:generic}
	\begin{algorithmic}[1]
		\STATE{} Initialize parameters $\theta$ (policy), $\phi$ (critic), and $\psi$ (reward)
		\STATE{} Initialize replay buffer $\mathcal B$ with randomly-generated trajectories
		\FOR{$i=1, \ldots, N$}
		\STATE{} // Reward learning
		\STATE{} Generate queries from $\mathcal B$\label{alg:generateQueries}
		\STATE{} Update $\mathcal D$ with answers to queries from the oracle
		\STATE{} Update $\psi$ using $\mathcal D$ (e.g., to maximize \cref{eq:ml})

		\STATE{} // RL training
		\STATE{} Update $\mathcal B$ with new trajectories generated with $\pi_\theta$\label{alg:RLbegin}
		\STATE{} Update $\theta$ (actor) using $\mathcal B$ and $\R_\psi$
		\STATE{} Update $\phi$ (critic) using $\mathcal B$ and $\R_\psi$\label{alg:RLend}
		\ENDFOR{}
	\end{algorithmic}
\end{algorithm}
This basic generic algorithm is summarized in \cref{alg:generic}, where an off-policy actor-critic scheme is assumed to be used for the \gls{RL} training part, but other \gls{RL} policy learning approaches can, of course, also be used here.
For an on-policy algorithm, such as PPO~\citep{schulman2017proximal}, only the recently generated transitions are used for training.
For a DQN-like algorithm, lines~\ref{alg:RLbegin} to~\ref{alg:RLend} would be replaced by a loop where transitions are generated by a behavior policy based on the current estimate of the $Q$-function (e.g., $\varepsilon$-greedy algorithm) and the $Q$ network is updated using mini-batches of transitions sampled from the replay buffer $\mathcal B$.

An efficient \gls{RLHF} algorithm needs to overcome several difficulties which are specific to this setting:
\begin{itemize}
	\item The oracle may provide various types of feedback (see \cref{sec:feedback}).
	The questions of what information some given feedback provides and how observed feedback can be exploited need to be answered (see \cref{sec:reward_learning}).
	\item Informative queries need to be generated to minimize the efforts of the oracle, which is crucial when it is a human (see \cref{sec:labelcollection}).
	Active learning techniques (see next subsection) can be adapted to face this challenge.
	\item The \gls{RL} agent is actually trained in a non-stationary environment since the reward approximator is concurrently updated.
	The \gls{RL} training part therefore needs to account for this factor, e.g., using non-vanishing learning rates (see \cref{sec:policy-learning}).
	\item There is also the question of how the agent's performance can be meaningfully evaluated, especially if the reward function is not known (see \cref{sec:application}).
	\item Collecting feedback directly from humans introduces its own challenges, such as the question of a suitable user interface and the associated issues of delay between query and feedback observation, or the feedback variability and reliability (see \cref{sec:labelcollection}).
	This may explain why many studies evaluate novel \gls{RLHF} algorithms with simulated feedback.
\end{itemize}

A standard \gls{RL} algorithm can be run in the \gls{RL} training part, as done in most previous work in \gls{RLHF} (although this may not be the best approach).
This suggests that any improvements in a standard deep \gls{RL} method (e.g., auxiliary losses~\citep{gelada2019deepmdp}, planning in learned model~\citep{hafner2020dream}, curriculum learning~\citep{narvekar2020curriculum}, or data augmentation~\citep{laskin2020reinforcement,lee2020network,lin2020invariant}) may potentially be transferred to the \gls{RLHF} setting.
In addition, most previous work in \gls{RLHF} directly uses trajectories stored in replay buffer $\mathcal B$ to synthesize queries.
An interesting research direction to explore in \gls{RLHF} would be to specifically generate trajectories in order to be able to synthesize more informative queries (instead of only generating trajectories that are beneficial for \gls{RL} training).
This would lead to tackling a novel exploration-exploitation dilemma:
Shall we visit state-action pairs that may be bad but may help better learn the reward function, or shall we visit state-action pairs that we currently think are good?
This is further discussed in \cref{sec:gathering-better-data}.

In \gls{RLHF}, since the oracle is a human or a group of humans, reducing the number of queries is crucial to limit the labeling cost.
Therefore, the reward learning part requires techniques similar to those proposed in active learning, which we recall next.

\subsection{Active Learning}

In active learning~\citep{settles2012active}, the task is to strategically select data points for labeling to minimize the amount of labeled data required to achieve a desired level of performance, which is particularly valuable in scenarios like \gls{RLHF} where labeling is costly.
Unlike batch learning, where labeled data is predetermined, active learning empowers the learner to actively select the most informative unlabeled instances for labeling, maximizing the learning process with limited labeled data.
We will only briefly introduce the active learning task here and then discuss the strategies for creating informative queries considered thus far in \cref{sec:labelcollection}.

For \gls{RLHF} with pairwise comparisons, this setting can be formally described as follows.
Suppose there is a set of $N$ pairs of trajectories $\{(\tau_1^i, \tau_2^i) \mid i=1, \ldots, N\}$, where each pair $(\tau_1^i, \tau_2^i)$ can be interpreted as an unlabeled instance.
To efficiently learn a reward function to explain observed pairwise comparisons, an agent can select a set of unlabeled pairs (possibly a singleton) to query an oracle to obtain their labels.

At a high level, the main idea in active learning is to query data points to quickly reduce the epistemic (i.e., reducible) uncertainty about the predictions of the learned model, although other aspects can be important, such as the representativeness of the queried data points (see \citet{wang2024comprehensivea} for a survey).
Two main representations are considered to describe this epistemic uncertainty: either using an ensemble of models or using a Bayesian representation.
In both cases, a first basic approach selects queries using uncertainty-based criteria in order to focus on instances with high prediction uncertainty as measured by, e.g., variance or entropy computed over predictions.
In contrast to the first approach, where the selection criteria are instance-based, a second approach considers criteria that may depend on all instances.
Possible options are, for instance, expected model change, expected error reduction, or density-weighted uncertainty-based criteria.
Here, the weights in the expectation or density allow us to take into account the distributional information about the instances and, therefore, to focus on the higher-density regions.

\section{Feedback}\label{sec:feedback}

\begin{figure}[H]
	\centering
	\definecolor{bgcolor}{HTML}{3c8e4b}
\begin{tikzpicture}[%
	every text node part/.style={align=center},%
	>=stealth'%
]
	\node[inner sep=0,outer sep=0,text=white] (agent-title) {Agent};
	\node[rectangle,minimum width=3.50cm,below=0.3cm of agent-title,fill=white,text=black] (policy) {\small Policy};
	\node[rectangle,minimum width=3.50cm,below=0.7cm of policy,fill=white,text=black] (rewardmodel) {\small Reward Model};
	\begin{scope}[on background layer]
		\node[rectangle,rounded corners,fit={(agent-title) (policy) (rewardmodel)}, inner ysep=0.2cm,fill=bgcolor] (agent) {};
	\end{scope}

	\node[inner sep=0,outer sep=0,text=white,right=3.5cm of agent-title] (environment-title) {Environment};
	\node[rectangle,minimum width=2.2cm,below=0.3cm of environment-title,fill=white,text=black] (dynamics) {\small Dynamics};
	\begin{scope}[on background layer]
		\node[rectangle,rounded corners,fit={(environment-title) (dynamics)}, inner ysep=0.2cm,fill=bgcolor] (environment) {};
	\end{scope}
	\draw[thick,->] (agent.east |- environment.180) to node[above=-0.05cm,pos=0.04,anchor=south west] {\footnotesize Action \( a_t \)} (environment.180);

	\coordinate (midway) at ($(environment.west |- policy) + (-1.45,0)$);
	\draw[thick,-|] (environment.west |- policy) -- (midway) node[below=-0.05cm,pos=0.46] {\footnotesize State \( s_{t+1} \)}; %
	\draw[thick,->] (midway) -- (policy -| agent.east) node[below=-0.05cm,pos=0.53] {\phantom{S}\footnotesize \( s_t \)};

	\draw[thick,->] (rewardmodel.70) to node[right=-0.05cm] {\color{white}\footnotesize Reward \( \hat r_{t + 1} \)} (policy.south -| rewardmodel.70);
	\draw[thick,->] (policy.south -| rewardmodel.110) to node[left=-0.05cm] {\color{white}\footnotesize Action \( a_t \)} (rewardmodel.110);

	\node[rectangle,minimum height=0.7cm,minimum width=2cm,rounded corners,below=1.15cm of environment.south,anchor=south,fill=bgcolor,text=white,draw=red!90!black,very thick] (evaluator) {Labeler};

	\draw[thick,->] (rewardmodel.4) to[->] node[above=-0.05cm,midway] {\footnotesize Query \( q_i \)} (evaluator.west |- rewardmodel.4);
	\draw[very thick,->,draw=red!90!black] (evaluator.west |- rewardmodel.-4) to[->] node[below=-0.05cm,midway] {\footnotesize Label {\( l_i \)}} (rewardmodel.-4);
\end{tikzpicture}
	\caption{RLHF diagram highlighting components discussed in this section.}
\end{figure}

Feedback mechanisms are fundamental to the success of any \gls{RL} system.
In the standard setting as described in \cref{sec:preliminaries-rl}, \gls{RL} agents expect feedback in the form of scalar immediate rewards.
These rewards are most commonly determined by a hand-engineered reward function, which can be used to evaluate any state-action combination.
As discussed in \cref{sec:introduction-whyhumanfeedback}, it is desirable to allow humans to refine behavior interactively through feedback instead of requiring them to pre-specify a reward function.

While a human could, in principle, assign rewards to each of the agent's actions directly, thereby taking the role of the reward function, this is usually impractical for multiple reasons.
The main challenge is the human effort required to provide rewards on a sufficiently regular basis, i.e., at least once per episode.
In addition to that, directly integrating human rewards into the \gls{RL} loop would require these rewards immediately, which would impede the learning pace while waiting for human feedback.
Finally, the standard \gls{RL} setting expects numeric state-action rewards, which are challenging to provide in a consistent manner.

In contrast to directly rewarding each of the agent's actions, \gls{RLHF} as discussed in this survey (see \cref{sec:scope-of-the-survey}) harnesses indirect and asynchronous feedback methods.
Such methods avoid the challenges of immediate numeric rewards and are also better aligned with human interaction patterns, resulting in improved learning progress and human user experience.

A feedback type is a kind of interaction in which a human conveys some information about their preferences.
Examples include pairwise comparisons and direct critiques.
This section is concerned with the many ways in which human feedback can be expressed and used.
Several previous works have already studied and sorted feedback types by listing the most common ones~\citep{jeon2020rewardrational,yuan2024unirlhf} and discussing their attributes and dimensions~\citep{%
	metz2023rlhfblender,%
	lindner2022humans%
}.
The attributes and classes described in this section build upon this prior work and can be considered a synthesis and extension of it.

The remainder of this section will start by discussing relevant attributes of feedback types that can be used to classify them (\cref{sec:feedback-attributes}).
We will then discuss common classes and examples of interactive feedback types (\cref{sec:feedback-classes}) as well as some non-interactive types that can serve as initializations (\cref{sec:feedback-initializations}).

\subsection{Attributes of Feedback Types}\label{sec:feedback-attributes}

Feedback types may differ in many dimensions, some of which relate to the way feedback is given (arity, involvement), others to the form of the query instance it is given on (granularity, abstraction), and yet others to features of the human interaction (intent, explicitness).
The attributes we discuss in this section are based on the framework proposed by \citet{metz2023rlhfblender}.
We have adjusted and added terminology where it aids clarity, generalized the distinction between relative and absolute feedback to `arity', and added the categories of co-generative involvement and literal intent.
Furthermore, we systematically analyze a set of exemplary feedback types in the next section, expanding on the initial examination of a smaller set of abstract classes by \citet{metz2023rlhfblender}.
In the following, we will introduce each of the six attributes in more detail.
\begin{description}
	\item[Arity]
		This attribute describes whether a single instance is evaluated in isolation (\emph{unary}) or relative to other instances (\emph{binary}, \emph{\(n\)-ary}).
		Unary feedback, e.g., an absolute score for an observed behavior, is often convenient for detailed and descriptive feedback but lacks any grounding and therefore puts a great burden on the human to provide consistent feedback.
		Non-unary feedback always has an implicit grounding but requires the instances to be comparable.
		While \( n \)-ary feedback, such as a ranking over multiple observed behaviors, can provide more information than binary feedback, it also puts a higher cognitive burden on the labeler.
	\item[Involvement]
		The labeler may either passively \emph{observe} an instance, actively \emph{generate} it, or coactively participate in its generation (\emph{co-generation}).
		Passive involvement, exemplified by an evaluator that observes, but does not intervene in a robot's behavior, poses the smallest challenge to the labelers since it does not require the ability to demonstrate the task.
		It can also easily be directed at the most informative examples with active learning techniques.
		Unfortunately, passive feedback often cannot match the information density of active feedback such as a human steering a robot to generate a demonstration.
		It is, therefore, common to combine both types to first initialize the reward model from (possibly very suboptimal) active feedback and then refine it from passive feedback.
		Between these two extremes is co-generative feedback, in which a human can share control with the agent.
		This can be less demanding than active feedback and makes it possible to direct the human's attention to the most informative samples, but it is still more taxing than purely passive involvement.
	\item[Granularity]
		Feedback may also differ on the granularity of the instances being evaluated.
		This ranges from feedback over whole \emph{episodes} over partial \emph{segments} to feedback on individual \emph{steps} (i.e., states, actions, or state-action pairs).
		A more coarse-grained granularity has the advantage of giving humans more context and getting feedback for larger sections of behavior but also poses credit assignment problems.
		Finer-grained feedback is much easier to learn from and simplifies credit assignment, but is often impractical or tedious for humans to provide.
		It is also possible to strike a compromise, as demonstrated by \citet{guan2021widening} who propose to use queries on step granularity, but batch queries within an episode to provide additional context and make it easier to give feedback.
		Note that we only classify a type of feedback as ``episode'' granularity if it \emph{requires} entire episodes.
		If it is compatible with partial segments as well, we classify it as ``segment'' even if the discussed source paper uses entire episodes.
	\item[Abstraction]
		This describes whether feedback is given directly on raw \emph{instances}, e.g., behavior recordings (see granularity), or on abstract \emph{features} of the instances, such as an autonomous car's speed and the forces experienced by the passengers.
		While feature-level information can be easier to learn from, extracting useful features is challenging.
		In some contexts, it may also be harder for a human to make abstract judgments rather than more intuitive instance-level judgments.
		Note that this always refers to the level of abstraction that the user sees, which may differ from the features used as inputs for the reward model.
		Types of feedback that depend on active generation (see involvement), such as improvements, generally work on a raw instance level.
	\item[Explicitness]
		Humans may communicate \emph{explicitly} for the purposes of feedback, e.g., by responding to a pairwise comparison query, or \emph{implicitly} as a side-effect of actions directed at other purposes such as interventions in the agent's behavior.
		While explicit information is easiest to learn from, implicit information is often much more readily available and can possibly communicate more detailed preferences.
	\item[Intent]
		The assumed human intent can be important for feedback processing.
		A human may be \emph{evaluative}, \emph{instructive}, or \emph{descriptive} in their explicit feedback, while they are generally \emph{literal} in their implicit feedback.
		Evaluative, instructive, and descriptive feedback is pedagogical in nature, aiming to teach a reward function, whereas literal feedback is a byproduct of a human actor's efforts to optimize the reward function directly.
		Descriptive feedback is often given on the level of the task, e.g., through a partial reward function.
		The other types of intent, in contrast, are generally given within the context of a particular instance of behavior.

		The distinction between literal and pedagogical feedback was introduced by \citet{milli2020literal}.
		They argue that humans with pedagogical intent (i.e., evaluative, instructive, or descriptive) may act differently compared to humans with literal intent.
		Even though they find that assuming the (wrong) literal intent can still lead to better reward inference, it still indicates that it can be important to know this intent to choose the right human model (see \cref{subsec:misspecification}).
\end{description}

\subsection{Common Classes}\label{sec:feedback-classes}

Even though the concrete types of feedback used in the literature are rarely exactly the same, they can generally be sorted into a set of common classes.
We will describe a selection of those classes, their defining attributes, and examples of concrete instances in the literature in the following.
\newcommand{\classhl}[1]{#1}
\begin{table}
	\caption{%
		An overview of the common classes and their defining attributes.
		The feedback classes are grouped into primary classes, which can be used on their own to learn a reward model, supplementary (sup.) classes, which can be used in conjunction with a primary class, and representation-focused (rep.) classes, which can be used to learn a representation prior to learning a reward model.
		When \maybeany{} is specified for an attribute, this indicates that it is not a defining feature of the class and may vary in different instantiations.
	}\label{tbl:feedback-classes-features}
	\begin{tabular}{lllllllll}
		\cmidrule[\heavyrulewidth](l{0.25em}){2-8}
		& \textbf{Class}              & \textbf{Granularity} & \textbf{Involvement}   & \textbf{Arity}       & \textbf{Abstr.} & \textbf{Intent}        & \textbf{Expl.} \\
		\cmidrule(l{0.25em}){2-8}
		\ldelim\{{7}{4mm}[\parbox{4mm}{\rotatebox[origin=c]{90}{Primary}}]{} %
		&\classhl{Critique}           & \maybeany{} & Observed      & Unary       & \maybeany{} & Evaluative    & Explicit \\
		&\classhl{Comparisons}        & \maybeany{} & Observed      & 2+          & \maybeany{} & Evaluative    & Explicit \\
		&\classhl{Inter-Temporal}     & Segment     & Observed      & Unary       & \maybeany{} & Evaluative    & Explicit \\
		&\classhl{Proxy Rewards}      & Episode     & Observed      & \maybeany{} & Feature     & Descriptive   & Explicit \\
		&\classhl{Social Behavior}    & Segment     & Observed      & Unary       & Instance    & Literal       & Implicit \\
		&\classhl{Improvements}       & Episode     & Co-generative & Unary       & Instance    & \maybeany{}   & \maybeany{} \\
		&\classhl{Natural Language} & \maybeany{} & Observed   & Unary       & \maybeany{} & Descriptive   & Explicit \\
		\addlinespace[1.5ex]
		\ldelim\{{2}{4mm}[\parbox{4mm}{\rotatebox[origin=c]{90}{Sup.}}]{} %
		&\classhl{E-Stops}            & Episode     & Observed      & Unary       & Instance    & Literal       & Implicit \\
		&\classhl{Importance}         & \maybeany{} & Observed      & \maybeany{} & \maybeany{} & Descriptive   & Explicit \\
		\addlinespace[1.5ex]
		\ldelim\{{2}{4mm}[\parbox{4mm}{\rotatebox[origin=c]{90}{Rep.}}]{} %
		&\classhl{Feature Traces}     & Segment     & Active        & Unary       & Instance    & Descriptive   & Explicit \\
		&\classhl{Similarity Queries} & \maybeany{} & Observed      & Ternary     & \maybeany{} & Descriptive   & Explicit \\
		\cmidrule[\heavyrulewidth](l{0.25em}){2-8}
	\end{tabular}%
\end{table}
\Cref{tbl:feedback-classes-features} gives an overview of the classes and their attributes as described in \cref{sec:feedback-attributes}.

\subsubsection{Primary Feedback Classes}

This section introduces common feedback classes which can be used on their own to learn a reward model.
The classes are critique, comparisons, inter-temporal feedback, proxy rewards, social behavior, improvements, and natural language.

\paragraph{Critique}

Critique is arguably the most direct type of feedback.
In this setting, the human expresses their preference by directly critiquing an instance of agent behavior, often in the form of binary feedback.
Note that critique, as considered in this survey, is distinct from directly supplying a reward signal since it is given in an asynchronous and indirect manner (see \cref{sec:scope-of-the-survey}).
In the critique setting, a labeler may for example observe recordings of agent behavior and give their approval or disapproval.
The defining features of critique are that the human passively observes the behavior (\textbf{involvement}), gives feedback on a single instance (\textbf{arity}), and does so explicitly (\textbf{explicitness}) with an evaluative \textbf{intent}.
The feedback may be given for any \textbf{granularity} and on any level of \textbf{abstraction}.

There are many examples of critique feedback in the literature.
\Citet{xiao2020fresh} employ binary feedback on individual state and action pairs.
Although they learn a shaping reward that complements an environment reward signal, the same technique could be used without environment reward.
\Citet{huang2023ganbased} extend this to multi-label feedback, allowing the user to distinguish between a regular good or bad action and a terminal action that achieves the goal or fails to do so.
They map these classes to scalar values and then learn a reward model by regression.
\Citet{wang2020maximizing} present an approach to learning a reward model from noisy critiques in the form of human physiological signals (brain signals) using active querying.
In contrast to this action-level feedback, \citet{fu2018variational,singh2019endtoend} rely on binary outcome success labels.
\Citet{fu2018variational} introduce the basic approach, which \citet{singh2019endtoend} extend by moving to an off-policy setting and including online queries, thereby reducing the reliance on many positive examples by interactively correcting false positives.

In addition to learning the main reward function, critique can also be used for safety evaluation.
\citet{cosner2022safetyaware} train a secondary reward model focused on safety from binary action critiques.
This is in addition to the main reward model, which is trained from comparisons in their approach.
Note that this secondary safety model could, in principle, be trained with any of the feedback types discussed here, using methods identical to the ones used for reward learning.

\paragraph{Comparisons}

Binary comparisons and rankings are among the most common types of feedback.
The defining features of comparisons are that the human passively observes the behavior (\textbf{involvement}), gives relative feedback on multiple instances (\textbf{arity}), and does so explicitly (\textbf{explicitness}) with an evaluative \textbf{intent}.
It is most commonly given on a segment (\textbf{granularity}), but other granularities such as individual actions are also possible~\citep{cosner2022safetyaware}.
Similarly, comparisons are commonly requested on an instance level (\textbf{abstraction}), but this is not a requirement.

Comparisons were first used for direct policy learning~\citep{akrour2011preferencebased,cheng2011preferencebased}, but were later extended to the reward-learning setting~\citep{wirth2016modelfree,christiano2017deep}.
The most common setting~\citep{christiano2017deep} relies on pairwise comparisons of trajectory segments, but comparisons of individual states or actions were also considered in early \gls{PbRL} works~\citep{furnkranz2012preferencebased} and comparisons can even be extended to more abstract trajectory features \citep{pinsler2018sample}.

This basic setting has been extended and modified in various ways.
To reduce noise in the labels, it is common to extend the binary choice by giving the labelers the option to avoid the hard choice and instead indicate incomparability, uncertainty, or perceived similarity~\citep{holladay2016active}.
This option is commonly interpreted as ``equally preferable'', e.g., as if each trajectory had an equal probability of being preferred in the preference predictor.
It is also common, however, to additionally provide an ``incomparable'' option which results in simply omitting the query from the data set \citep{christiano2017deep,ibarz2018reward}.
In contrast to this, \citet{verma2023exploiting} explicitly state that they do not allow for the equally preferred option, arguing that these cases are rare enough not to matter very much.
Another line of research suggests that more precision in the expression of pairwise preferences, such as softening the hard binary choice to scalar feedback indicating the strength of a preference~\citep{wilde2021learning,touvron2023llama}, can be beneficial for preference learning.
Other extensions change the pairwise setting to choices among larger choice sets~\citep{ziegler2020finetuning} or even full rankings~\citep{myers2021learning,ouyang2022training}.
\Citet{jain2015learning} compare different kinds of re-ranking feedback (select first which is better than top, choose best from top-5, choose best from random set).
Their study suggests that the first two are roughly equivalent, while the latter is much less informative.
\Citet{askell2021general} evaluate binary comparisons as well as ranking, but find that binary comparisons (even if extracted from a ranking) perform better.
\Citet{ziegler2020finetuning} note that for language tasks, a larger choice set can amortize the time needed for a labeler to get acquainted with the context necessary to understand a query.
\Citet{basu2019active} propose to use hierarchical queries, i.e., a sequence of pairwise comparisons that build up on each other.

Pairwise comparisons are generally easier to supply than ratings, even when the feedback is given by a foundation model \citep{wang2024rlvlmf}.
While absolute feedback is generally dependent on a policy serving as an implicit baseline \citep{macglashan2017interactive}, the explicit baseline in pairwise comparisons can be more easily controlled.
Therefore, comparisons provide many benefits over absolute ratings, such as reduced bias, inconsistencies, and subjectivity~\citep{yannakakis2015ratings}.
On the flip-side, comparisons convey relatively little information per label.
\Citet{tien2023causal} study the weaknesses of pairwise-comparison-based reward learning.
Since the amount of information provided for each label is small, these models are prone to causal confusion, i.e., misattributing reward to noise and misidentifying the reward function.

\paragraph{Inter-Temporal Feedback}\label{sec:feedback-classes-intertemporal}

One limitation of trajectory comparisons is that they require a set of roughly comparable trajectories.
In many real-world environments, starting conditions or even the agent's current task may vary between episodes.
In these cases, it is hard for human labelers to compare trajectories from these episodes, limiting the usefulness of comparison feedback.
One way to remedy this limitation is to provide feedback within a single trajectory.
Instead of comparing a set of instances with each other as in regular comparative feedback, inter-temporal feedback conveys relative judgments over different states in time within a single instance, e.g., by comparing a robot's behavior at the start and the end of a trajectory.
The defining features of inter-temporal feedback are that it is given explicitly (\textbf{explicitness}) on a segment (\textbf{granularity}) while passively observing (\textbf{involvement}) a single instance (\textbf{arity}) of the agent's behavior with evaluative \textbf{intent}.
It is most commonly given on raw instances, but any level of \textbf{abstraction} is possible in principle.
There are two main ways to convey this feedback: \emph{Reward sketching} and \emph{inter-temporal preferences}.

Reward sketching, as introduced by \citet{cabi2020scaling}, involves users sketching a visual representation of the reward function over time.
This type of feedback, which can be given by sketching a graph with the mouse while watching a behavior recording, provides intuitive reward annotations for each time step.
\Citet{rahtz2022safe}
also adopted this approach, referring to it as ``one of the highest-bandwidth feedback mechanisms currently available''.

Inter-temporal preferences were introduced by \citet{abramson2022improving}.
In this setting, humans give feedback on multiple points of a trajectory, indicating whether an agent makes progress towards or regresses from a goal.
This is then interpreted as preferences relative to the other labeled and unlabelled points.
The authors note that one potential downside of this feedback type is that labelers may tend to give preferences on short-term actions that are easy to judge, failing to communicate long-horizon preferences.
\Citet{cui2018active} propose a similar type of feedback, in which humans segment a trajectory into good and bad parts.
This makes it possible to derive many state-action labels from a few segmentation points.

\paragraph{Proxy Rewards}

Proxy rewards are partial or inaccurate reward functions that convey information about the task the agent is supposed to complete but may not induce optimal behavior.
An example of this is a reward function that only rewards the agent for reaching the goal state but does not penalize negative side-effects.
This form of feedback does not generally refer to any particular behavior instance but instead gives global direction for the entire task.
However, in line with our selection criteria (\cref{sec:scope-of-the-survey}), we only consider proxy reward feedback that is interactive and online within the context of one or multiple observations to fill holes in the initial description.
The defining features of proxy reward feedback is that the labeler passively observes the agent's behavior (\textbf{involvement}) and gives feedback explicitly (\textbf{explicitness}) on a feature-level (\textbf{abstraction}) with descriptive intent (\textbf{intent}).
Proxy reward feedback may be given with respect to a single or multiple instances (\textbf{arity}), although it generally refers to multiple instances.
It is most commonly given on an episode (\textbf{granularity}), but other granularities are possible in principle.

The work by \citet{he2021assisted} exemplifies this form of feedback through an iterative process where designers provide proxy reward functions on training environments.
The method maintains a belief distribution over reward functions consistent with behaviors that achieve high proxy reward on training environments, then actively proposes new environments that maximize information gain about the true reward function, helping designers refine their reward specifications.
Alternatively, \citet{mindermann2018active} suggest querying about the reward function, extending the inverse reward design framework of \citet{hadfield-menell2017inverse}, which treats proxy reward functions as observations to infer the designer's true intent.
The active approach by \citet{mindermann2018active} allows users to choose from a set of understandable, linear proxy rewards or to specify which features are more critical in the linear reward structure.
In a related setting, \citet{guan2023relative} lets the user specify \emph{changes} to a current symbolic reward function in the form of changes to target attributes (e.g., walking speed).

\paragraph{Social Behavior}

Humans give rich implicit social feedback in the form of facial reactions and gestures when interacting with agents.
The defining attributes of this type of feedback are that it is given implicitly (\textbf{explicitness}) on passively observed (\textbf{involvement}) segments (\textbf{granularity}) with respect to a single instance (\textbf{arity}) and literal \textbf{intent}.

\Citet{cui2021empathic} propose a framework to learn reward functions from such social behavior.
They suggest a two-phase training setup.
In the first phase, they ground the implicit feedback by use of incentives, i.e., they incentivize humans to have a known objective.
After learning a mapping from feedback to reward, they use regular \gls{RL} techniques to learn a policy.
Note that the learned reward function can be seen as conditional on human implicit feedback and, therefore, they require a human in the loop throughout training.

\paragraph{Improvements}

Improvements are a form of feedback in which the human improves on the agent's behavior, either by intervening as the agent acts or by providing a corrected behavior after the agent acts.
To improve an episode, it is usually necessary to observe the entire episode (\textbf{granularity}) at the instance level (\textbf{abstraction}).
In this type of feedback, the human both observes and demonstrates behavior, resulting in co-generative \textbf{involvement}.
Improvements generally relate to a single reference trajectory being improved (unary \textbf{arity}), although an improvement could also be interpreted as a binary comparison between the improved and the non-improved trajectory.
Improvements are most commonly provided explicitly with instructive \textbf{intent}.

We distinguish between post-facto improvements, calling them \emph{corrections}, and improvements made while the agent is acting, calling them \emph{interventions}.
The key difference is that the uncorrected trajectory is available in the case of corrections, while it can only be estimated in the case of interventions.

Interventions can be considered to be an instance of the shared autonomy setting since the agent and the user share autonomy to reach a common goal.
There are two main ways to leverage interventions:
One is to learn from the occurrence of an intervention itself that the prior behavior was suboptimal, as done, e.g., by \citet{luo2024rlif} by directly inferring negative rewards from interventions without learning a reward model, and the other is to learn from the corrected trajectory.
The latter setting is studied by \citet{abramson2022improving}, who ask humans to intercede on agent failure.
They use this to collect targeted demonstrations where the agent is the weakest and to identify challenging situations for their evaluations.
The gathered data is then used for behavior cloning and reward model training.
The fact that a correction occurred is not directly used as feedback.
In addition to the above distinction, interventions can come in two main formats:
Either by overtaking control from the agent to correct its behavior or by physically correcting the agent's movements.
\Citet{losey2022physical} proposes to learn a reward model from such physical corrections, a method that was later extended \citep{losey2018including} to incorporate uncertainty for active learning and risk-sensitive deployment.
\Citet{li2021learning} further extend this setting to learn from a sequence of correlated physical corrections without needing to wait until the trajectory is completed.

The correction case is closely related to the setting of coactive learning~\citep{shivaswamy2015coactive}, in which a learning system repeatedly proposes a solution which a user may correct to reach a common goal.
\Citet{jain2015learning} treat corrections as a demonstration while \citet{jeon2020rewardrational} propose (but do not evaluate) an alternative interpretation of inferring implicit preferences from comparisons by assuming the corrected trajectory is preferred over the original one.
Corrections are also commonly used in the \gls{LLM} fine-tuning setting to incorporate AI feedback:
By prompting a language model with a set of principles, it can critique and revise a generated response according to those principles.
This can then be used to generate demonstrations \citep{bai2022constitutional} or targeted pairwise preferences \citep{castricato2024suppressing} and to internalize these principles with further training on this data.

\paragraph{Natural Language}

Natural language is a versatile form of feedback that can be used to convey a wide range of information.
It cannot only be used to express preferences, but also to suggest concrete changes.
Natural language feedback may be given on any \textbf{granularity}, at any level of \textbf{abstraction}.
Its defining features are that it is given explicitly (\textbf{explicitness}) in the context of a single (\textbf{arity}) observed behavior or policy (\textbf{involvement}) with descriptive \textbf{intent}.

While much work in language for \gls{RL} focuses on the problem of learning a policy that follows natural-language instructions in a non-interactive setting \citep{hermann2017grounded,nair2021learning}, we focus on works where language is used as feedback, not for task definition.
Note, however, that \gls{RLHF} can be a useful tool for learning a language-conditioned policy, regardless of whether or not the feedback itself is in the form of natural language.
As an example of this, \citet{abramson2022improving} use \gls{RLHF} in an interactive agents setting, where one agent (responsible for solving problems) is controlled by the learned policy and another agent (responsible for setting tasks) is controlled by the human.
\gls{RLHF} for \glspl{LLM} can also be viewed as language-conditioned \gls{RL}, although
\citet{shen2024trickledown} find that reward models in this context often fail to distinguish nuances in task descriptions.

Language may be interpreted as a form of feedback in the reward-rational choice setting \citep{jeon2020rewardrational}, where an utterance implies a set of trajectories compatible with the utterance (grounding) and the human can choose an utterance that maximizes the probability of a desired action (i.e., the human is assumed to be specific which leads to pragmatic reasoning).
This can be used to infer a reward which may have caused the human to provide the utterances they did provide.
A similar approach is followed, e.g., by \citep{sumers2022linguistic}.

An alternative way of incorporating language feedback is by extracting more structured forms of feedback from the language.
An example is to use sentiment analysis to extract information about the reward function from language feedback, as proposed by \citet{sumers2021learning}.

Language feedback can also be interpreted by an \gls{LLM} directly, without learning a reward model.
This is demonstrated in the concurrent works by \citet{ma2024eureka} and \citet{xie2024text2reward} who propose to learn a symbolic reward model in the form of a language-model written piece of (Python) code.
They use natural language feedback to improve the reward model based on observations of the agent's behavior induced by the previous version of the reward model.

In addition to the way language feedback is used, the way it is elicited can also have an impact on the learning process.
\Citet{sumers2022how} study the relative merits of natural language instructions and descriptions of the desired outcome.
They find that instructions tend to work better for low-autonomy settings, while descriptions are more effective in high-autonomy settings.

\subsubsection{Supplementary Classes}\label{sec:feedback-commonclasses-supplementary}
This section introduces two feedback classes, e-stops and importance, that can be used in conjunction with a primary class to learn a reward model.
These classes are not sufficient to learn a reward model on their own but can supplement a primary feedback type.

\paragraph{E-Stops}

Emergency stops (e-stops)~\citep{ghosal2023effect} are an active type of feedback.
In this type of feedback, the human may intervene with the agent's behavior by stopping it, i.e., they may choose to stop the agent's current trajectory at any point, for example to prevent the agent from causing harm.
This is closely related to interventions but, in contrast to those, e-stops do not suggest an alternative action.
The defining features of e-stops are that the human passively observes the agent's behavior (\textbf{involvement}), gives absolute feedback on a single instance (\textbf{arity}) on the instance level (\textbf{abstraction}), and does so implicitly (\textbf{explicitness}) as a side-effect of regular interaction.
The \textbf{intent} is literal due to the implicit nature.
For the purposes of intervention, the human usually observes the full episode (\textbf{granularity}).
Due to the small amount of infrequent information they provide, e-stops should only be considered as a supplementary feedback type.

E-stops are primarily intended to prevent bad behavior and only implicitly convey information about the correct behavior.
This interaction and the arising incentives have been formalized in the form of the ``off-switch game'' by \citet{hadfield-menell2017offswitch}.
\Citet{jeon2020rewardrational} propose to interpret this as a form of reward-rational feedback, where the `off' choice maps to the trajectory with the robot remaining still after the off switch has been triggered (see \cref{subsec:boltzmann}).
\Citet{kahn2021land} demonstrate that a robot can learn to navigate using such feedback.

\paragraph{Importance}

Another form of supplementary feedback may come in the form of importance labels, communicating which parts of the observation are important for the objective, e.g., by marking salient regions in an image representing the agent's final state or important words in the language model's response.
Its defining features are that the importance information itself does not contribute towards generating behavior samples (observed \textbf{involvement}), is of descriptive \textbf{intent}, and is given explicitly (\textbf{explicitness}).
\textbf{Granularity}, \textbf{arity}, and \textbf{abstraction} may vary depending on the primary feedback type.
Since importance feedback needs a base task with respect to which the importance is defined, it cannot be used on its own but is rather a supplementary type of feedback.

One way to convey this information is by labeling salient parts of a visual input.
This is explored by \citet{guan2021widening}, who augment pairwise comparisons with manually annotated visual saliency maps, informing the algorithm which parts of the visual input contributed to the decision.
They leverage these annotations for data augmentation by assuming that random perturbations to irrelevant (non-salient) regions do not impact the human preferences.
\Citet{basu2018learning} take an even more direct approach by combining comparative feedback with direct feature queries, i.e., asking the user which feature is important for inferring the reward.

\subsubsection{Representation-Specific Classes}\label{sec:feedback-commonclasses-representationfocused}

While the previous classes of feedback types are all aimed at directly learning a reward function, there are also classes of feedback types that do not directly learn a reward function but rather help to learn a better representation.

\paragraph{Feature Traces}\label{sec:feedback-commonclasses-featuretraces}

While many approaches either rely on hard-coded features or learn a model entirely end-to-end, it is also possible to actively elicit new features from human feedback.
Feature traces were proposed by \citet{bobu2022inducing} as an approach to actively learn new relevant features.
This type of feedback relies on a human operator to demonstrate a behavior in which a certain feature of interest, such as the distance to a sensitive object, monotonically increases or decreases, e.g., by moving a robot arm closer to the object.
They make it possible to extend the set of features once the current set can no longer adequately explain the human feedback supplied through another type of feedback.
The defining characteristics of feature traces are that they are of descriptive (\textbf{intent}) and explicitly (\textbf{explicitness}) given in an active manner (\textbf{involvement}) for a single (\textbf{arity}) segment (\textbf{granularity}) on an instance-level \textbf{abstraction}.

Feature traces are strongly related to inter-temporal preferences (\cref{sec:feedback-classes-intertemporal}) since both types rely on changes in feature or reward values in the course of a single trajectory.
\Citet{bobu2022inducing} propose to learn from feature traces by leveraging a Bradley-Terry model to learn the feature values, similar to other approaches that use such a model to learn reward values.
Similar to importance feedback, feature traces rely on another type of feedback to actually make use of the learned features and is, therefore, a purely supplementary form of feedback.
For instance, \citet{bobu2022inducing} use intervention feedback to train a reward model on the set of features derived using feature traces.

\paragraph{Similarity Queries}\label{sec:feedback-commonclasses-similarityqueries}

Similarity queries are a feedback type aimed at learning a representation conforming to a notion of similarity and difference in the trajectory space.
That aim is closely aligned with that of feature queries, though the actual queries are more similar to comparisons.
The queries consist of triples of trajectories, with one anchor and two alternatives, for which the human has to decide which pair is more similar.
Responses to similarity queries are given on observed behavior (\textbf{involvement}) with ternary \textbf{arity}, descriptive \textbf{intent}, and explicit feedback (\textbf{explicitness}), while the \textbf{granularity} and \textbf{abstraction} may vary.
This type of feedback was first introduced by \citet{bobu2023sirl}, who used it to learn representations for reward learning.

\subsection{Initializations}\label{sec:feedback-initializations}

Some modes of communicating reward functions are not interactive nor online and, therefore, do not directly fit within the scope of this survey (\cref{sec:scope-of-the-survey}).
However, since these are often used to initialize a reward function for later refinement with some of the previously discussed interactive feedback types, they are still worth mentioning.

Initializations are most commonly given by examples of successful task completions, either in the form of
terminal or goal states~\citep{xie2018fewshot},
expert demonstrations~\citep{ibarz2018reward,fu2018learning,palan2019learning,lee2021pebble,biyik2022learning,abramson2022improving,huang2023ganbased},
demonstrations with success labels~\citep{du2023visionlanguage}, or
ranked demonstrations~\citep{brown2019extrapolating}.
It is even possible to infer some human preferences by the state of the environment alone, e.g., assuming the parts of the initial environment that are under the user's control are largely set up in accordance with the user's preferences \citep{shah2019preferences,lindner2021learning}.
Since offline initialization is not the main focus of our survey, we do not cover these works in detail.
We refer the interested reader to literature on inverse \gls{RL}~\citep{arora2021survey} for further details on learning from demonstrations in particular.

\subsection{Choice of Feedback Type}

The best feedback type is not always clear and may depend on the task, the user, or the agent.
It may also change over time as the agent learns more about the task.
\Cref{tbl:feedback-classes-features} may serve as a starting point to select a set of feedback types which may be applicable based on the possible user interaction and expertise, e.g., by the desired granularity, level of abstraction or involvement.
This choice may be informed, among other features, by the task complexity or time horizon:
\Citet{jain2015learning} compare purely passive (re-ranking based) with active (slight trajectory improvements) feedback and conclude that the former is usually preferred by humans when the action space is large while the latter is preferred for complex tasks.
Similarly, \citet{sumers2023show} empirically demonstrate that language is a more effective teaching modality than demonstrations for complex concepts.
The relevance of the time horizon is highlighted in the work by \citet{sumers2022linguistic}, which suggests that instructive feedback is more efficient for short time horizons and descriptive feedback for longer horizons.

In addition to these static choices, \cref{sec:adaptive-feedback-choice} will discuss how to choose a feedback type adaptively.
It is also possible to let the user choose the feedback type, as demonstrated by \citet{jain2015learning} who let the user choose between re-ranking and improvement.
\Citet{jeon2020rewardrational} propose to use this choice of feedback type itself as a source of information (``meta-choice'').

\subsection{Combination of Feedback Types}\label{sec:combination-of-feedback-types}

In addition to using any of the previously described feedback types in isolation, combining multiple feedback types is both possible and often advantageous.
There are three main ways to combine feedback types:
(a) a two-phase setup, consisting of initialization and refinement,
(b) integrating a primary feedback type with a supplementary one; and
(c) merging multiple primary feedback types.

The two-phase setup can be used to either initialize the reward model from offline data or to learn a representation that improves later reward learning.
A common approach involves using one feedback type, typically demonstrations (see \cref{sec:feedback-initializations}), to initialize the reward model, subsequently fine-tuning this model with another type, such as comparisons.
This method is exemplified by \citet{ibarz2018reward}, who combined demonstrations and pairwise comparisons.
For a more detailed discussion on feedback types suitable for initialization, we refer to \cref{sec:feedback-initializations}.
Alternatively, a representation-focused type of feedback might be employed initially (\cref{sec:feedback-commonclasses-representationfocused}) to cultivate a superior representation, followed by the application of a primary feedback type for reward model learning.

Combining a primary feedback type with a supplementary one can be beneficial to make the most of the available user interactions.
Supplementary feedback, while not sufficient for reward model training by itself, can often be collected cheaply or as a side-effect of other interactions, making it a valuable addition to the primary feedback type.
We refer to \cref{sec:feedback-commonclasses-supplementary} for a discussion on supplementary feedback types.

Finally, combining multiple primary feedback types can be beneficial to capitalize on the strengths of each.
For instance, \cite{koppol2020iterative} combine informative and demanding queries with less-informative and less-demanding ones to achieve a balance between cognitive load and informativeness.
Similarly, to enhance expressivity, \citet{mehta2023unified} combines demonstrations, pairwise comparisons, and corrections, allowing users to select their preferred type of feedback.

\section{Label Collection}\label{sec:labelcollection}

\begin{figure}[H]
	\centering
	\definecolor{bgcolor}{HTML}{3c8e4b}
\begin{tikzpicture}[%
	every text node part/.style={align=center},%
	>=stealth'%
]
	\node[inner sep=0,outer sep=0,text=white] (agent-title) {Agent};
	\node[rectangle,minimum width=3.50cm,below=0.3cm of agent-title,fill=white,text=black] (policy) {\small Policy};
	\node[rectangle,minimum width=3.50cm,below=0.7cm of policy,fill=white,text=black] (rewardmodel) {\small Reward Model};
	\begin{scope}[on background layer]
		\node[rectangle,rounded corners,fit={(agent-title) (policy) (rewardmodel)}, inner ysep=0.2cm,fill=bgcolor] (agent) {};
	\end{scope}

	\node[inner sep=0,outer sep=0,text=white,right=3.5cm of agent-title] (environment-title) {Environment};
	\node[rectangle,minimum width=2.2cm,below=0.3cm of environment-title,fill=white,text=black] (dynamics) {\small Dynamics};
	\begin{scope}[on background layer]
		\node[rectangle,rounded corners,fit={(environment-title) (dynamics)}, inner ysep=0.2cm,fill=bgcolor] (environment) {};
	\end{scope}
	\draw[thick,->] (agent.east |- environment.180) to node[above=-0.05cm,pos=0.04,anchor=south west] {\footnotesize Action \( a_t \)} (environment.180);

	\coordinate (midway) at ($(environment.west |- policy) + (-1.45,0)$);
	\draw[thick,-|] (environment.west |- policy) -- (midway) node[below=-0.05cm,pos=0.46] {\footnotesize State \( s_{t+1} \)}; %
	\draw[thick,->] (midway) -- (policy -| agent.east) node[below=-0.05cm,pos=0.53] {\phantom{S}\footnotesize \( s_t \)};

	\draw[thick,->] (rewardmodel.70) to node[right=-0.05cm] {\color{white}\footnotesize Reward \( \hat r_{t + 1} \)} (policy.south -| rewardmodel.70);
	\draw[thick,->] (policy.south -| rewardmodel.110) to node[left=-0.05cm] {\color{white}\footnotesize Action \( a_t \)} (rewardmodel.110);

	\node[rectangle,minimum height=0.7cm,minimum width=2cm,rounded corners,below=1.15cm of environment.south,anchor=south,fill=bgcolor,text=white,draw=red!90!black,very thick] (evaluator) {Labeler};

	\draw[very thick,->,draw=red!90!black] (rewardmodel.4) to[->] node[above=-0.05cm,midway] {\footnotesize Query \( q_i \)} (evaluator.west |- rewardmodel.4);
	\draw[thick,->] (evaluator.west |- rewardmodel.-4) to[->] node[below=-0.05cm,midway] {\footnotesize Label {\( l_i \)}} (rewardmodel.-4);
\end{tikzpicture}
	\caption{RLHF diagram highlighting components discussed in this section.}
\end{figure}
\begin{table}
\centering
	\caption{%
		An overview of query selection strategies used in \gls{RLHF} approaches.
	}\label{tbl:acquisition-overview}
 \small
 \begin{tabular}{@{}ll@{\hskip4pt}c@{\hskip10pt}c@{\hskip10pt}c@{\hskip10pt}c@{\hskip10pt}c@{}}
	\toprule
    References & \multicolumn{6}{c}{Factors} \\
    \cmidrule(l){2-7}
  & Uncertainty & On-policy & Query & Trajectory & Query & Query \\
  & & Data & Simplicity & Quality & Diversity & Cost \\
		\cmidrule(r){1-1}\cmidrule(l){2-7}
		\citet{daniel2014active} & Probability of improvement & \sxmark{} & \sxmark{} & \sxmark{} & \sxmark{} & \sxmark{} \\
        & Expected improvement & \sxmark{} & \sxmark{} & \sxmark{} & \sxmark{} & \sxmark{} \\
        & Upper confidence bound & \sxmark{} & \sxmark{} & \sxmark{} & \sxmark{} & \sxmark{} \\
		\citet{christiano2017deep} & Ensemble variance & \sxmark{} & \sxmark{} & \sxmark{} & \sxmark{} & \sxmark{} \\
		\citet{sadigh2017active} & Volume removal & \sxmark{} & \sxmark{} & \sxmark{} & \sxmark{} & \sxmark{} \\
        \citet{biyik2024batch} & Volume removal & \sxmark{} & \sxmark{} & \sxmark{} & \scmark{} & \scmark{} \\
        \citet{wilde2018learning} & Feasible space reduction & \sxmark{} & \sxmark{} & \scmark{} & \sxmark{} & \sxmark{} \\
		  \citet{ibarz2018reward} & Random & \sxmark{} & \sxmark{} & \sxmark{} & \sxmark{} & \sxmark{} \\
        \citet{mindermann2018active} & Information gain & \sxmark{} & \sxmark{} & \sxmark{} & \sxmark{} & \sxmark{} \\
        \citet{cui2018active} & Information gain & \scmark{} & \sxmark{} & \scmark{} & \sxmark{} & \sxmark{} \\
        \citet{racca2019teacheraware} & Entropy & \sxmark{} & \scmark{} & \sxmark{} & \sxmark{} & \sxmark{} \\
        \citet{biyik2019asking} & Mutual information & \sxmark{} & \scmark{} & \sxmark{} & \sxmark{} & \sxmark{} \\
		\citet{biyik2020active} & Information gain & \sxmark{} & \sxmark{} & \sxmark{} & \sxmark{} & \sxmark{} \\
        \citet{reddy2020learning} & Ensemble-averaged KL- & \sxmark{} & \sxmark{} & \scmark{} & \scmark{} & \sxmark{} \\
         & divergence to mean output & \sxmark{} & \sxmark{} & \sxmark{} & \sxmark{} & \sxmark{} \\
         \citet{novoseller2020dueling}  & Dueling posterior sampling & \sxmark{} & \sxmark{} & \scmark{} & \sxmark{} & \sxmark{} \\
         \citet{wilde2020active,wilde2021learning} & Maximum regret & \sxmark{} & \sxmark{} & \sxmark{} & \sxmark{} & \sxmark{} \\
        \citet{lee2021pebble} & Ensemble-averaged entropy & \sxmark{} & \sxmark{} & \sxmark{} & \sxmark{} & \sxmark{} \\
		\citet{lindner2021information} & Information gain & \scmark{} & \sxmark{} & \sxmark{} & \sxmark{} & \sxmark{} \\
		\citet{katz2021preferencebased} & Posterior sampling & \sxmark{} & \sxmark{} & \scmark{} & \sxmark{} & \sxmark{} \\
        \citet{myers2023active} & Expected value of information & \sxmark{} & \sxmark{} & \sxmark{} & \sxmark{} & \sxmark{} \\
        \citet{biyik2024batch} & Mutual information
        & \sxmark{} & \sxmark{} & \sxmark{} & \scmark{}& \sxmark{} \\
        \citet{dwaracherla2024efficient} & Double Thompson sampling
        & \sxmark{} & \sxmark{} & \scmark{} & \sxmark{} & \sxmark{} \\

        \citet{hu2024querypolicy} & Random
        & \scmark{} & \sxmark{} & \sxmark{} & \sxmark{} & \sxmark{} \\

		\bottomrule
	\end{tabular}
\end{table}

In this section, we explain how preference data can be collected for training a reward model, independent of the specific type of feedback used.
We start by overviewing the active learning problem posed by query selection, e.g., how queries can be generated for a given query type and how even the type of feedback itself can be selected.
Following this, we discuss issues arising in such human-computer interaction.

\subsection{Active Learning}

We first discuss query generation and selection for a given feedback type, then the extension of these techniques to the choice of feedback type.

\subsubsection{Query Generation and Selection}\label{sec:labelcollection-activelearning}

One core problem that needs to be tackled in \gls{RLHF} is that of learning about the human's preferences.
This problem shares some similarities with the active learning setting since the agent can actively query a human teacher about those preferences.
However, in contrast to standard active learning, which usually assumes a supervised learning setting, in \gls{RLHF}, the agent needs to solve this problem in the context of \gls{RL}.
This means that it can both influence the distribution of the data (i.e., the transitions) and decide which data should be labeled.

As the \gls{RL} agent is trained and its learned policy changes, the trajectories it generates will naturally evolve.
Most work directly uses the trajectories obtained during \gls{RL} training for preference learning.
Using such trajectories, candidate queries are often generated randomly.
However, for pairwise comparisons, a more efficient approach is to generate queries by exploiting preference transitivity \citep{hwang2023sequential}.
Alternatively, the agent can also generate trajectories specifically to be used for querying (not necessarily for \gls{RL} training), possibly with a learned transition model to limit the sampling cost (e.g., \citet{reddy2020learning,liu2023efficient}).
This kind of active generation of queries can possibly lead to more informative ones.

In order to efficiently learn a suitable reward model, the agent must generate and select queries (Line~\ref{alg:generateQueries} from \cref{alg:generic}) so that it can quickly learn a good strategy using those queries.
This selection is performed via a criterion, usually called \emph{acquisition function}\footnote{We follow the terminology used in Bayesian active learning.}, which allows the queries to be compared.
Although most existing work in \gls{RLHF} uses an acquisition function that provides some measure of \emph{uncertainty} (e.g., about the learned rewards), which is arguably one of the most important factors in active learning, an efficient acquisition function (see \Cref{tbl:acquisition-overview} for works dedicated to improving query selection) may need to include various additional other aspects, such as:
\emph{on-policy data}, \emph{query simplicity}, \emph{trajectory quality}, \emph{query diversity}, or \emph{query cost}, which will be discussed one by one in the following.
As a side note, interestingly, as highlighted by \citet{habibian2022here}, the queries asked by an \gls{RL} agent may also reveal its current reward learning stage.

\begin{description}
\item[Uncertainty]
This factor usually corresponds to epistemic uncertainty \citep{hullermeier2021aleatoric}, which represents how uncertain the agent is about the ground-truth reward function.
Epistemic uncertainty can be contrasted to aleatoric uncertainty, which describes inherent stochasticity in the system.
While the latter cannot be fully eliminated, the former can be reduced with additional queries.
Uncertainty is usually one of the most important aspects to consider when deciding which query to ask.
It is usually represented either as a probability distribution (i.e., belief) in Bayesian approaches or using an ensemble of reward networks to approximate this belief.
However, other representations are also possible, e.g., using the recently-proposed epistemic neural network \citep{osband2023epistemic}.

With a belief representation, various classic  acquisition functions have been considered in the Bayesian framework, such as the \emph{probability of improvement}, the \emph{expected improvement}, or the \emph{upper confidence bound}.
For instance, \citet{daniel2014active}  compare those three criteria in a robotic domain and observe that the latter yield the best performance, but at a high cost in terms of number of queries, while the first asks the fewest number of queries, but with a slightly lower asymptotic performance.
Other alternative criteria have been considered, such as \emph{volume removal}~\citep{sadigh2017active,basu2018learning,basu2019active} or \emph{information gain}~\citep{mindermann2018active,biyik2019asking,biyik2020active}.
The \emph{volume removal} criterion uses the minimum volume of the hypothesis set removed by an answer to a query as the acquisition function and has been shown to be effective in practice.
However, \citet{biyik2019asking} show that volume removal has uninformative global optima and argue that the practical effectiveness is due to non-convexity leading to local optima that are informative.
They also show that optimizing for information gain has the same computational complexity while avoiding this flaw.
One drawback of these Bayesian approaches is that they require maintaining a distribution over reward functions (e.g., using Gaussian processes~\citep{daniel2014active} or simpler probability distributions, such as Gaussian distribution, but with a linear reward model) and, therefore, may not be suitable for more complex domains due to the computational complexity or the strong assumption about the reward model.

When using an ensemble instead of a direct belief representation, these criteria for epistemic uncertainty reduction correspond to measures of disagreement within the ensemble.
Previous criteria could possibly be applied, but one popular candidate is the \emph{variance} of the ensemble outputs~\citep{lee2021pebble,metcalf2023sampleefficient,gleave2022uncertainty} or equivalently its standard deviation \citep{eberhard2022actively}.

The \emph{average entropy} of the ensemble outputs (e.g., when assuming a Bradley-Terry model for the human answers) has also been used~\citep{lee2021pebble,lee2021bpref,park2022surf}.
However, note that it does not quantify epistemic uncertainty but rather the aleatoric uncertainty in the human's answers, as provided by the response model of the human.
Therefore, this criterion may not be suitable in the \gls{RLHF} setting since it amounts to focusing on the queries for which an answer is expected to be the most random (according to the Bradley-Terry model).
By definition of this model, the segments in the pairwise comparisons are the most similar in terms of returns and are, therefore, the hardest to answer for the human.

In contrast to those acquisition functions that lead to deterministic query selection, sampling-based approaches have also been studied, from pure random selection \citep{ibarz2018reward} to Thompson sampling \citep{katz2021preferencebased} and its variants \citep{novoseller2020dueling,dwaracherla2024efficient}.
Recently, in the context of fine-tuning LLMs with pairwise comparison queries,  \citet{dwaracherla2024efficient} shows that using the relatively novel epistemic neural network \citep{osband2023epistemic}, double Thompson sampling \citep{wu2016double}, which naturally favors better elements to be compared, performs well experimentally.

In addition to epistemic uncertainty, one may also take the outcomes into account to select queries, that is consider utilities (i.e., returns or expected returns in \gls{RL}).
In a Bayesian setting, this leads to acquisition functions such as
\emph{expected value of information}~\citep{myers2023active} or \emph{information gain over return differences}~\citep{lindner2021information}, while in a non-Bayesian setting, the notion of \emph{regret}, which measures the difference of performance between a policy optimal for the ground-truth reward function and a policy optimal for a learned reward function, can be used~\citep{wilde2020active}.

Finally, it should be mentioned that naturally approaches with theoretical guarantees usually also use uncertainty as the main criterion for label collection (e.g., \citet{novoseller2023diprl}, see \cref{sec:theory} for more details).

\item[On-Policy Data]
Only focusing on uncertainty is likely insufficient or inefficient in \gls{RL} because
the previous methods may focus on choosing queries to identify the reward function as precisely as possible uniformly on the whole state-action space.
However, it may be important to favor more on-policy trajectories to guarantee the relevance of the generated queries for the current policy, assuming that the behavior policy is stochastic (i.e., the current stochastic policy in SAC or the current deterministic policy with some noise, such as in DDPG).
Indeed, improving reward learning in state-action regions that may never be visited with the current policy would lead to wasteful queries~\citep{lindner2021information}.
One simple approach to ensure that the data is more on-policy is by simply sampling from the current policy~\citep{cui2018active} or favoring more recently-generated trajectories~\citep{hu2024querypolicy}.

\item[Query Simplicity]
Selecting queries only based on their informativeness may lead to queries that are hard for a human to answer, which is, for example, the case for the average entropy.
The ease of answering a query is important to alleviate the cognitive load of the human oracle.
Some work specifically takes this aspect into account, for instance, by considering the similarity of consecutive queries~\citep{racca2019teacheraware} or the information gain.
For this latter criterion, \citet{biyik2019asking} show that in contrast to volume removal, it naturally leads to queries that are easier to answer for a human because information gain can be increased when the uncertainty in the human answer is lower.

\item[Trajectory Quality]
Most approaches directly use the trajectories generated during \gls{RL} training.
Especially early in training, these can be very bad with respect to the ground-truth reward function.
In addition to that, they can be irrelevant or even contradictory for a given task~\citep{katz2021preferencebased}.
Building queries on such trajectories may lead to unnatural queries for a human to respond to, such as comparing a very bad trajectory with an irrelevant one.
\citet{katz2021preferencebased} measure trajectory quality by optimizing over sampled reward functions.
Similarly, \citet{cui2018active} generate trajectories using optimal policies for reward functions sampled from the current Bayesian belief.

\item[Query Diversity]
When asking many queries (in batch, in sequence), the diversity of the queries becomes especially crucial to avoid asking redundant queries.
Most work \citep{christiano2017deep,lee2021pebble,verma2024hindsight} follows a very myopic approach:
Queries are often selected from a usually randomly-generated set of potential queries, and sequences of queries are not really coordinated.
While some work exists that specifically tackles the selection of a batch of diverse queries \citep{biyik2018batch,biyik2024batch}, the latter is rarely considered due to its computational intractability. %
Indeed, planning ahead a sequence of queries would amount to solving a sequential decision-making problem under uncertainty over a combinatorial action space (i.e., the set of possible queries).
For diverse batch querying, previous work considered using clustering methods such as k-medoids \citep{biyik2018batch} or more recently determinantal point processes, which define probability distributions that promote diversity \citep{biyik2024batch}.
In addition to approaches focused on selecting diverse queries, another promising research direction involves generating inherently diverse and informative behaviors to select from.
This can be achieved with techniques such as entropy-regularized \gls{RL} \citep{ziebart2008maximum,haarnoja2017reinforcement} or generative methods based on flow networks \citep{bengio2021flow}.
The latter is particularly relevant for RLHF for generative models such as \glspl{LLM} \citep{hu2024amortizing}.
To our knowledge, this direction has not yet been explored in the context of \gls{RLHF}.

\item[Query Cost]
The cost of generating queries may also be an important factor if the interaction of the human is live since it may not be practical to let the human wait before showing any queries~\citep{biyik2024batch}.
In that case, it may be more important to quickly show some relatively good queries instead of computing the most informative ones.
Although this factor may not translate directly into an acquisition function, it may influence the choice of the acquisition function and its implementation in a given problem.
\end{description}

Since various different acquisition functions have been considered, some effort~\citep{lee2021pebble,lee2021bpref} has been made to compare them.
Generally speaking, uncertainty-based criteria (e.g., variance or average entropy) seem to often perform better empirically compared to random selection, a query diversity-based criterion alone or combined with an uncertainty-based criterion.
Surprisingly, random selection has been shown to perform competitively in some cases~\citep{christiano2017deep,ibarz2018reward}.
Thus, a better understanding of which acquisition function should be preferred in which situation or domain is still an open question.

In addition, combinations of different criteria have naturally also been evaluated.
For instance, \citet{reddy2020learning} use four acquisition functions (high uncertainty, high novelty, high reward, low reward) in parallel.
This approach has also been validated in a 3D environment~\citep{rahtz2022safe}.
A more sophisticated approach consists of considering a portfolio of acquisition functions and learning to select them using a multi-armed bandit approach~\citep{hoffman2011portfolio}.

Various extensions to the basic setting have also been investigated.
In the context of multiple human labelers, the issue of selecting reliable teachers to query arises~\citep{daniels-koch2022expertise}.
Assuming all teachers have the same preferences, this can be modeled for pairwise comparisons by incorporating a rationality coefficient \( \beta \) into a Bradley-Terry model and estimating this factor:
\begin{align}\label{eq:bt model beta}
	\max_\psi \prod_{i=1}^N \frac{1}{1 + \exp(\beta(\R_\psi(\traj^i_2) - \R_\psi(\traj^i_1)))}
	\,\text,
\end{align}
where a higher $\beta$ corresponds to a more reliable human (see \cref{subsec:rationality_coefficient}).
The setting in which this assumption does not hold, i.e., the labeler's reward functions differ~(a setting already considered in inverse \gls{RL}~\citep{choi2012nonparametric}), has also been studied recently~\citep{siththaranjan2024understanding,xue2024reinforcement,dong2024aligndiff,myers2021learning,bakker2022finetuning}.
Interestingly, a noisy oracle may sometimes provide more information than a completely reliable oracle.
For instance, in \cref{eq:bt model beta}, the probability of erroneous answers given by the noisy oracle is related to how much a segment is preferred to the other one, and \citet{chan2021human} show that such Boltzmann-rational behavior can be more informative for reward inference than perfectly rational behavior.
In contrast, only a binary preorder over segments can be inferred from the answers of a reliable and deterministic oracle, which may not be enough to recover the true reward function.

Another notable recent work is that by \citet{ellis2024generalized} who raise the issue of identifiability of the ground-truth reward function: many reward functions result in the same optimal behaviors.
The authors propose a framework that enables the generation of acquisition function for various definitions of reward similarity, such as the one discussed in \cref{sec:rewardmodel-utilitylearning-evaluating}.

\subsubsection{Adaptive Choice of Feedback Type}\label{sec:adaptive-feedback-choice}

In addition to selecting queries within a given feedback type, it is also possible to actively select the feedback type itself~\citep{fitzgerald2022inquire}.
The best choice of feedback type can depend on many factors, such as human rationality as well as task-dependent factors, some of which may change during the labeling process.
\Citet{ghosal2023effect} formalize this setting as one in which we try to select a feedback design (or feedback type) \( x \) out of the space of possible designs \( \mathcal X \) such that the expected information gain over the distribution of reward functions is maximized for the next human response.
Concretely, the goal is to choose a feedback design by means of
\[
	x = \argmax_{x \in \mathcal X} \mathbb E_{c_h \sim P(c_h \mid x)} \big[ D_{\mathit{KL}} \big( \mathbb P(\theta \mid c_h, x) \mid \mathbb P(\theta) \big)\big]
	\,\text,
\]
where \( c_h \) is the human response to a query defined by \( x \) and \( \mathbb P(\theta) \) is the prior distribution over reward functions.

\Citet{ghosal2023effect} find that the most informative feedback type depends on the (type-dependent) rationality of the human labeler (see \cref{sec:labeling-psychologyaware}).
More precisely, it is shown that the most informative feedback depends on the rationality factor, e.g., while demonstrations are more informative than comparisons when the human is highly rational, comparisons should be preferred in less-rational settings.
Given that this rationality might change due to factors such as fatigue or an individual labeler's capabilities, this suggests that adaptively adjusting the feedback type during the labeling process may be worthwhile.
Further study of this relationship is a promising area for future work.

\subsection{Challenges of Human Labeling}\label{sec:challenges-of-human-labeling}

This section explores the label collection process, which follows after query selection.
This task intersects with several related disciplines, especially within the social sciences, as it encompasses the design of interactions to facilitate informative query responses.
A prominent field in this area is psychometrics~\citep{furr2021psychometrics}, which focuses on measuring psychological attributes, including preferences.
Similarly, survey research~\citep{fowler2013survey} is dedicated to developing techniques for gathering information from individuals via surveys.
Human-computer interaction plays a significant role as well, investigating the design of user interfaces tailored for preference elicitation~\citep{pommeranz2012designing}.
Moreover, preference label collection is also necessary for discrete choice experiments within health economics~\citep{ryan2008using}, where it is used for the assessment of service values.

\subsubsection{Psychology-Aware Preference Elicitation}\label{sec:labeling-psychologyaware}

Understanding human psychology is essential for effective preference elicitation in \gls{RLHF} systems.
Human decision-making is complex, often diverging from traditional rational choice models due to cognitive, social, and emotional factors.
This complexity is exemplified by phenomena like fatigue, which can affect the reliability of choices based on the order of queries.
This section overviews these phenomena, exploring how constructive preferences, biases, framing effects, and social interactions shape the observed choices.
Recognizing and addressing these psychological underpinnings is key to developing more accurate and reliable systems.
In this section, we will discuss various psychological phenomena, such as cognitive biases and response biases, and related effects (fallacies, biases, heuristics, psychological phenomena impacting decision-making processes), which may falsify labels by adding systematic bias or noise.

Preference learning methods typically assume the existence of inherent, stable preferences that can be elicited through querying.
Contrary to this assumption, psychological research, such as the work by \citet{lichtenstein2006construction}, indicates that preferences are often constructed during the elicitation process and may vary with the method of elicitation or over time.
This suggests that the feedback type not only affects elicitation's effectiveness but also shapes preferences.
Systematic biases, noise, and other psychological factors may influence observed choices, challenging the traditional models of human choice used to infer latent utilities (see \cref{sec:choice-model}).
The elicitation method, query presentation, and context thus play a critical role in shaping measured preferences, compounded by cognitive biases and irrationalities.

The influence of psychological phenomena on preference learning has been well-documented in the literature, especially within the context of explicit preference elicitation for recommender systems.
For instance, \citet{tran2021humanized} provide a thorough discussion of the relationship between psychology and recommender systems.
Similarly, \citet{atas2021psychologyaware} review how preference construction is influenced by cognitive biases, personality traits, and emotional states in recommender systems, discussing effects like serial position, framing, anchoring, choice overload, and preference visibility.
In a more specialized discussion, \citet{mandl2011consumer} focus on cognitive biases in the context of consumer decision-making and its interaction with recommender systems.
Finally, \citet{kaufmann2023challenges} link these psychological aspects to \gls{RLHF}, discussing the common practice of using synthetic instead of real human feedback for algorithm evaluation and highlighting the limitations of that approach.
They further discuss challenges posed by real human feedback, many of which are related to the concepts discussed in the following paragraphs, as well as the opportunities provided by integrating psychological insights into \gls{RLHF} systems.

Constructive preferences are closely related to \emph{framing effects}, which refer to changes in elicited preferences based on how tasks or alternatives are described, even when these descriptions are essentially equivalent.
For example, presenting a choice as a loss versus a gain can lead to different decisions despite identical outcomes.
Moreover, \emph{serial position effects}, commonly known as primacy and recency effects, also play a significant role.
These effects describe the tendency for the beginning and end of an experience to influence subjective experience disproportionately.
This phenomenon becomes particularly relevant in scenarios like video choices, where the initial or concluding segments might disproportionately affect preferences.
\Citet{atas2021psychologyaware} discuss both of these effects in the context of recommender systems.

\emph{Ordering effects} pose another challenge in preference elicitation, where the sequence of queries can affect responses.
\Citet{day2012ordering} outline several factors contributing to these effects: institutional learning, changing preferences, and varying levels of cognitive effort.
Institutional learning involves gaining familiarity with the task and feedback type, which can enhance labelers' expertise and, consequently, the accuracy of their responses.
However, due to the constructive nature of preferences, these preferences may evolve during the elicitation process, leading to changing preferences.
This evolution might also be influenced by \emph{anchoring effects}, where previously seen instances bias responses.
Furthermore, cognitive effort levels can fluctuate due to factors like fatigue or boredom.
This is closely related to \emph{choice overload}, a form of fatigue from excessive choices, as discussed by \citet{atas2021psychologyaware} and bounded rationality, as explored by \citet{chen2013human}.
In such scenarios, labelers might opt out of making a choice when overwhelmed by options.
Bounded rationality refers to the limitations in human decision-making capabilities, particularly when processing large amounts of information, which aligns with the concept of choice overload.
To address these challenges, studies like \citet{biyik2019asking} and \citet{zhang2022timeefficient} propose methods to reduce cognitive effort in responding to queries.
\Citet{biyik2019asking} focus on posing queries that are straightforward for humans to answer, while \citet{zhang2022timeefficient} enhance the human evaluation process by presenting queries in a user-friendly format.

Multiple labelers often collaborate on the same task in preference elicitation, as studied, e.g., by \citet{barnett2023active} and \citet{daniels-koch2022expertise}.
This collaboration may lead to another set of biases if they have the opportunity to exchange information.
The exchange may either be direct or indirect through observing the system's predictions, which are based on the other labeler's feedback.
Such interactions can affect their preferences through several mechanisms, as identified by \citet{atas2021psychologyaware}: anchoring effects, transfer of emotional states, and conflict avoidance.
\emph{Anchoring effects}, for instance, occur when a labeler's choices are influenced by the knowledge of others' preferences or system predictions, a phenomenon also discussed under the term \emph{preference visibility}.
This bias can lead labelers to align their preferences with the anchors they are exposed to, which is a significant consideration in recommender systems.
Understanding these biases is crucial for designing \gls{RLHF} systems that mitigate the influence of labeler interactions on preference construction.

The effects previously discussed stem from systemic biases in preference expression.
In addition to these biases, choices may also be affected by noise.
This is commonly discussed under the term stochastic rationality, where an agent's behavior is rational with respect to an unobserved random state.
The reward-rational implicit choice framework, as introduced by \citet{jeon2020rewardrational}, addresses this by integrating a rationality factor \( \beta \) into the human choice model (see \cref{eq:bt model beta}).
This factor's impact has been further examined by \citet{ghosal2023effect} through synthetic experiments and user studies, demonstrating that accurately estimating this type-dependent rationality coefficient can enhance learning performance and guide feedback type selection (see \cref{sec:adaptive-feedback-choice}).
However, a practical method for estimating this factor remains a challenge.
While \citet{ghosal2023effect} use calibration feedback with a known latent utility function for estimation, such an approach is not feasible for most tasks.
In a related study, \citet{daniels-koch2022expertise} investigate a scenario with multiple teachers, focusing on the agent's ability to select the most knowledgeable or rational teacher.
Therefore, developing more advanced methods to estimate this factor, along with understanding its variability due to factors like fatigue or other ordering effects, presents a vital area for future research in preference elicitation.

Finally, the quality of human feedback is biased towards factors that are easy to judge.
\Citet{hosking2024human} demonstrates that in the case of \gls{LLM} fine-tuning, humans tend to favor assertiveness over factuality, since the latter is hard to judge without external assistance or resources.
A similar phenomenon was observed in the control setting by \citet{amodei2017learning}, where the agent learned a behavior that looked good only from the camera angle that the human labelers had access to.

Incorporating psychological insights into the preference-learning components of \gls{RLHF} systems is essential for optimizing their efficacy.
A key area of focus should be research aimed at mitigating biases and harnessing cognitive aspects of preference formation.
For instance, designing user interfaces that minimize framing effects and developing algorithms that account for ordering and serial positioning are crucial steps.
In this realm, \citet{metz2023rlhfblender} and \citet{yuan2024unirlhf} each propose a configurable user interface for studying various feedback types and their combinations.
Additionally, the study by \citet{krening2018interaction} on the impact of feedback type, such as binary critiques versus action advice, on task performance and labeler satisfaction highlights the significant role of feedback type in preference elicitation.
Furthermore, the work of \citet{pommeranz2012designing} in user-interaction design underlines the importance of having an expressive feedback type to increase user engagement.

The integration of these research findings into \gls{RLHF} systems points to a clear need for a more multidisciplinary approach.
Drawing insights from related fields like behavioral economics and psychology can provide valuable methodologies and perspectives.
Addressing irrational choice patterns and enhancing the quality of human feedback remain critical challenges.
As we continue to develop and refine these systems, the focus should be on creating robust frameworks that align learning processes with human behavior, effectively managing the inherent complexity and variability of human feedback.

\subsubsection{Importance of Researcher-Labeler Agreement}

High-quality labels are important for the final policy in an \gls{RLHF} process.
Early work on fine-tuning language models using \gls{RLHF} noticed a mismatch between the researcher's goals and the (paid) labeler's actual labels (researcher-labeler disagreement).
\Citet{ziegler2020finetuning} note that researchers agreed with each other about \( 60\% \) of the time (on 4-way comparisons, where random choice would result in \( 25\% \) agreement), while agreeing with labelers only \( 38\% \) or \( 46\% \) of the time (depending on the task).
\Citet{stiennon2020learning} attempt to reduce these disagreements by maintaining a hands-on relationship with the labelers, thereby ensuring high researcher-labeler agreement.
Concretely, they provide on-boarding with detailed instructions, keep an open channel of communication between researchers and labelers, and give feedback to the labelers.
They evaluate the researcher-labeler agreement and reach an agreement rate of \( 77\% \pm 2\% \).

Perfect labels are often impossible due to the inherently subjective nature of the task.
Returning to the example given by \citet{stiennon2020learning}, different researchers agreed with each other in only \( 73\% \pm 4\% \) of the cases.
\Citet{ouyang2022training} also report the agreement rates on a different task (instruction fine-tuning instead of summarization) and find that labelers agree with each other in \( 72.6 \pm 1.5\% \) of the time, after a screening procedure that, amongst others, selects labelers that agree with researcher labels.
Preferences can additionally be inconsistent between feedback types, as demonstrated by the findings of \citet{bansal2024peering}, which show
that preferences inferred from ratings and rankings significantly disagree for both human and AI annotators. %

The importance of quality does not trump the importance of quantity, however.
Indeed, \citet{stiennon2020learning} note that excluding low-confidence samples from the data set generally did not help with reward model training.
This indicates that even though quality is important, a larger quantity is still generally better.

The scale of the labeled data set required for effective training and refinement varies widely, impacting the quality of the resulting models.
In practice, the required amount of data depends on many factors including domain, task complexity, and level of pretraining.
For example, while \gls{LLM} fine-tuning generally often requires tens of thousands of samples due to the complexity of the task, at the same time it can leverage the extensive prior knowledge of the pretrained \gls{LLM} to reduce sample complexity compared to other tasks of similar complexity.

Studies have shown a broad range in data set sizes, from tens of labels in smaller studies~\citep{jain2015learning} to hundreds in more complex scenarios~\citep{christiano2017deep}.
Larger-scale applications may require thousands~\citep{guan2021widening,ouyang2022training} or even millions of labels~\citep{abramson2022improving}, each bringing its own challenges in ensuring label accuracy and consistency.

This variability in data set size underscores the need for rigorous label quality control measures across different scales.
In smaller data sets, each label carries more weight, making accuracy and precision critical.
Conversely, in larger data sets, the challenge lies in maintaining consistency and mitigating systematic biases that might emerge from the sheer volume of data.
The labeling setting varies in the surveyed works, from author-provided feedback~\citep{kim2023preference}, over small in-person studies~\citep{katz2021preferencebased}, to larger remote studies~\citep{kim2023preference}.
Each setting provides unique challenges to ensure high-quality labels.

Various works have suggested measures to improve label quality.
\Citet{hagendorff2022methodological} discuss the possible failure modes of the labeling task in more detail, for example, discussing systematic biases and conflicting motivation, and propose concrete changes to the training and evaluation methodology to alleviate these.
\Citet{glaese2022improving} suggest providing labelers with multiple natural language rules and collecting preference labels for each rule individually to improve label quality.
This is related to \citet{bai2022constitutional}, who propose to generate feedback automatically based on such a set of rules and a language model.

\section{Reward Model Training}\label{sec:reward_learning}

\begin{figure}[H]
	\centering
	\definecolor{bgcolor}{HTML}{3c8e4b}
\begin{tikzpicture}[%
	every text node part/.style={align=center},%
	>=stealth'%
]
	\node[inner sep=0,outer sep=0,text=white] (agent-title) {Agent};
	\node[rectangle,minimum width=3.50cm,below=0.3cm of agent-title,fill=white,text=black] (policy) {\small Policy};
	\node[rectangle,minimum width=3.50cm,below=0.7cm of policy,fill=white,text=black,draw=red!90!black,very thick] (rewardmodel) {\small Reward Model};
	\begin{scope}[on background layer]
		\node[rectangle,rounded corners,fit={(agent-title) (policy) (rewardmodel)}, inner ysep=0.2cm,fill=bgcolor] (agent) {};
	\end{scope}

	\node[inner sep=0,outer sep=0,text=white,right=3.5cm of agent-title] (environment-title) {Environment};
	\node[rectangle,minimum width=2.2cm,below=0.3cm of environment-title,fill=white,text=black] (dynamics) {\small Dynamics};
	\begin{scope}[on background layer]
		\node[rectangle,rounded corners,fit={(environment-title) (dynamics)}, inner ysep=0.2cm,fill=bgcolor] (environment) {};
	\end{scope}
	\draw[thick,->] (agent.east |- environment.180) to node[above=-0.05cm,pos=0.04,anchor=south west] {\footnotesize Action \( a_t \)} (environment.180);

	\coordinate (midway) at ($(environment.west |- policy) + (-1.45,0)$);
	\draw[thick,-|] (environment.west |- policy) -- (midway) node[below=-0.05cm,pos=0.46] {\footnotesize State \( s_{t+1} \)}; %
	\draw[thick,->] (midway) -- (policy -| agent.east) node[below=-0.05cm,pos=0.53] {\phantom{S}\footnotesize \( s_t \)};

	\draw[very thick,->,draw=red!90!black] (rewardmodel.70) to node[right=-0.05cm] {\color{white}\footnotesize Reward \( \hat r_{t + 1} \)} (policy.south -| rewardmodel.70);
	\draw[thick,->] (policy.south -| rewardmodel.110) to node[left=-0.05cm] {\color{white}\footnotesize Action \( a_t \)} (rewardmodel.110);

	\node[rectangle,minimum height=0.7cm,minimum width=2cm,rounded corners,below=1.15cm of environment.south,anchor=south,fill=bgcolor,text=white] (evaluator) {Labeler};

	\draw[thick,->] (rewardmodel.4) to[->] node[above=-0.05cm,midway] {\footnotesize Query \( q_i \)} (evaluator.west |- rewardmodel.4);
	\draw[thick,->] (evaluator.west |- rewardmodel.-4) to[->] node[below=-0.05cm,midway] {\footnotesize Label {\( l_i \)}} (rewardmodel.-4);
\end{tikzpicture}
	\caption{RLHF diagram highlighting components discussed in this section.}
\end{figure}

In this section, we delve deeper into the process of reward model learning, which we briefly touched on in \cref{subsec:reward_learning_basic}.
We will discuss various aspects associated with this topic, namely the various human feedback models, utility (i.e., reward) learning per se, different reward model inputs, and how to increase feedback efficiency. %

\subsection{Human Feedback Model}\label{sec:choice-model}

The basic premise underlying the majority of approaches in \gls{RLHF} is that human feedback is directly related to the reward function to be learned.
To this end, the human feedback must first be captured in a sound mathematical framework that establishes the connection to the reward function.
On a high level, one can break down almost all feedback types in \cref{sec:feedback-classes} to a choice scenario:
The human chooses one specific feedback option (label) from an available pool of possible feedback options (choice sets), which may be infinite\footnote{This point of view goes back to the work of \citet{jeon2020rewardrational}.}.
The query determines the explicit contents of the choice set.
For example, if the query is to compare two trajectories, then the choice set consists of all possible outcomes for these two trajectories.

Assuming that human choices are not always optimal, one obtains a fruitful mathematical framework when focusing on the probability
\begin{align}\label{def:general_human_choice}
	\mathbb{P}\left( \mbox{$c$ is chosen} \, | \, \mathcal{C} \right)
	\,\text,
\end{align}
where $\mathcal{C}$ is the set of possible choices and $c \in \mathcal{C}$ one explicit choice.
For the \gls{RLHF} scenario, where the agent asks queries $q_i$ and the human gives labels $l_i$ as feedback (see \cref{sec:RLHF_setting_intro}), the choice set is specified by a function of the query.
Formally, $\mathcal{C} = m(q)$ for some mapping $m$ that maps a query $q$ to the set of all possible candidate labels extractable from $q$ for the specific feedback type.
For example, if the query is to rank a finite number of trajectories, then the choice set is the set of all possible rankings that can occur for the trajectories involved.

With this view, we can therefore place~\eqref{def:general_human_choice} in the \gls{RLHF} context and write
\begin{align}\label{def:general_human_choice_RLHF}
	\mathbb{P}\left( \mbox{label $l$ is provided} \, | \, m(q) \right)
\end{align}
for the probability that a human labeler returns a label $l$ from all possible candidate labels that can be extracted from a given query $q$.
We explain next how this probability can be modeled and discuss various related modeling questions (e.g., human rationality, multiple humans, or Markov assumption).
One could also recover the noiseless scenario if the latter probability distribution is degenerated for all possible candidate label sets.

\subsubsection{Boltzmann Distribution}\label{subsec:boltzmann}

Human choice models as in~\eqref{def:general_human_choice} have been studied for a long time in various scientific fields such as psychology~\citep{thurstone1927law}, economics~\citep{train2009discrete}, or behavioral science~\citep{cattelan2012models}.
Accordingly, there are many different choice models to resort to for~\eqref{def:general_human_choice_RLHF}, which, in some cases, are the same models, just under different names.
A popular class of such human choice models assumes every choice option $c$ to be equipped with a (latent) utility $u_c$, which the human perceives in a perturbed way.
This perturbation is modeled by means of perturbation random variables $\epsilon_c$ that perturb the utility in an additive way, so that~\eqref{def:general_human_choice} becomes
\begin{align}\label{def:RUM_human_choice}
	\mathbb{P}\left( \mbox{$c$ is chosen} \, | \, \mathcal{C} \right) = \mathbb{P}\left( c = \argmax_{c \in \mathcal C} \ u_c + \epsilon_c \right)
	\,\text.
\end{align}
The translation for the \gls{RLHF} setting for~\eqref{def:general_human_choice_RLHF} is then accordingly
\begin{align}\label{def:RUM_human_choice_RLHF}
	\mathbb{P}\left( \mbox{label $l$ is provided} \, | \, m(q) \right) = \mathbb{P}\left( l = \argmax_{l \in m(q)} \ u_l + \epsilon_l \right)
	\,\text,
\end{align}
and we shall now stick to the \gls{RLHF} translation from now on.
These probabilities depend on the specific distributional assumptions that are made on the perturbation variables that only for specific cases lead to a closed-form of the right-hand sides of the latter equations.
When stipulating a standard Gumbel distribution for the perturbations, one always obtains a closed form that is proportional to the exponential utility of the provided label:
\begin{align}\label{def:Boltzmann_human_choice}
	\mathbb{P}\left( \mbox{label $l$ is provided} \, | \, m(q) \right)    \propto \exp( u_l )
	\,\text.
\end{align}
This is known as the \emph{Boltzmann distribution} that also appears in a perhaps slightly modified version in various different subfields of \gls{ML} and statistics.
When restricting to discrete (choice) sets for $m(q)$, this distribution is also known as the multinomial logit model~\citep{train2009discrete} or Gibbs measure~\citep{georgii2011gibbs}, and as the Bradley-Terry model~\citep{bradley1952rank} when the choice sets consist of pairs.
All of these also have a link to the Plackett-Luce model~\citep{luce1959individual,plackett1975analysis}, which is a probability distribution on the space of total orders or rankings (see~\citet{alvo2014statistical} for details).

This model is often used for various reasons.
A particularly compelling reason is the closed analytic form, which in turn makes it possible to obtain a closed form for the gradient with respect to the utilities.
Another reason is that this model satisfies Luce's axiom of choice~\citep{plackett1975analysis}, which requires the probability of choosing an option from a pool of choice options not being affected by the presence or absence of other options in the pool.
In this way, coherent decision-making is ensured, which, however, might be challenged as humans likely do not make fully rational decisions (see \cref{sec:labeling-psychologyaware}).
\Citet{jeon2020rewardrational} show that the usage of the Boltzmann distribution is justified by the principle of maximum entropy.
More precisely, they show that it is the maximum entropy distribution over choices for a so-called satisficing human decision maker, i.e., one who is making a choice with an optimal reward in expectation up to some slack $\epsilon>0$.

To build the bridge between reward learning and the modeling of human feedback, the Boltzmann distribution can be used by assuming that the utilities can be represented as a function of the reward function, usually featuring the return of a trajectory.
More specifically, one assumes a \emph{grounding function} $G$ that maps choice options (or labels) to the set of distributions over trajectories and sets the utility of a label $l$ as
\begin{align}\label{def:utility_RLHF}
	u_l := \mathbb{E}_{\tau \sim G(l)}[R(\tau)]
	\,\text.
\end{align}
Note that $u_l$ depends on the return $R$, so that we also may use $ u_l(R)$ to emphasize this dependency.
For the common case of pairwise trajectory comparisons, where for two trajectories $\tau_1, \tau_2$   we obtain for the possible labels  $l \in \{ \tau_1 \succ \tau_2 , \tau_1 \prec \tau_2 \} $ the respective utility by using the projection onto the preferred trajectory as the grounding function $G$.
Accordingly, the utility of the label represents essentially the utility of the preferred trajectory of that label, i.e., $\tau_1$ or $\tau_2$ in this case.
As another example consider the case of e-stops feedback (see \cref{sec:feedback-commonclasses-supplementary}).
Here, the possible labels $l$ provided by the user are $\texttt{STOP}_t$ and $\texttt{CONT}$ encoding the stopping at time $t$ or the continuation of a trajectory.
For the grounding function, one can define
\[
G(l) =
\begin{cases}
    \tau, & l = \texttt{CONT}\text, \\
    \tau_{0:t}\tau_t\dots\tau_t, & l = \texttt{STOP}_t\text,
\end{cases}
\]
where $\tau_{0:t}$ is the trajectory of $\tau$ trimmed to the stopping time $t$, and $\tau_t$ is the action-state pair at time $t$.
Table 1 in \citet{jeon2020rewardrational} provides an overview of the different grounding functions that lead to a specific feedback type.
It is worth noting that one can also easily find a grounding function for the feedback type of a (partial) order over trajectories as considered, for instance, by \citet{myers2021learning}.
Moreover, one can generalize this modeling approach by using (partial) segments instead of trajectories.

Although this general human feedback model has been much in use and shown to be useful for the sake of human alignment, it is not without its critics (see \citet{lindner2022humans} or Section 3.2.1 in \citet{casper2023open}).
This has led to different adaptions of the general model based on the Boltzmann distribution that will be discussed in the following.
Moreover, we will also concisely review other human feedback models that have been in use besides the Boltzmann distribution, discuss relevant work on the consequences or robustness of human feedback model misspecification, and highlight contributions on varying the standard assumptions on the nature of the human feedback.

\subsubsection{Human-Specific Rationality Coefficient}\label{subsec:rationality_coefficient}

The Boltzmann distribution in~\eqref{def:Boltzmann_human_choice} can be extended by a rationality coefficient $\beta \in [0, \infty)$ that reflects the precision of the human labeler: %
\begin{align}\label{def:Boltzmann_human_choice_rational}
	\begin{split}
		&\mathbb{P}\left( \mbox{label $l$ is provided} \, | \, m(q) \right) = \mathbb{P}\left( l = \argmax_{l \in m(q)} \ \beta \, u_l + \epsilon_l \right) \propto \exp( \beta \cdot u_l )
		\,\text.
	\end{split}
\end{align}
The higher $\beta$, the more~\eqref{def:Boltzmann_human_choice_rational} resembles a pointmass distribution modeling a highly rational human labeler that is always able to identify the option with highest utility.
The lower $\beta$, the more it resembles a uniform distribution modeling a highly irrational human labeler acting purely at random.
Without this extension, the commonly considered Boltzmann distribution (or Bradley-Terry model in the common case of pairwise comparisons) in~\eqref{def:Boltzmann_human_choice} assumes a rationality coefficient of \( 1 \).
\Citet{ghosal2023effect} show in their experiments that the estimation of this coefficient can indeed positively influence reward learning.
However, estimation requires a calibration reward function, as the rationality coefficient is otherwise not identifiable~\citep{bengs2020preselection}.
Similar findings are shown by \citet{daniels-koch2022expertise}, who model the rationality coefficient as a query-dependent function that might differ for the human labelers (see \cref{subsec:diverse_preferences}).

Another alternative to the rationality coefficient for representing irrational humans is achieved by introducing a query-independent error probability~\citep{christiano2017deep}.
To be more precise, it is assumed that the human labeler only adheres to the Boltzmann distribution in~\eqref{def:Boltzmann_human_choice} in 90\% of cases and otherwise makes completely random decisions.
This formulation is similar to Huber's contaminated model~\citep{mu2023huber}.

\subsubsection{Alternative Utility Notions}

\Citet{knox2024models}
show that the Boltzmann model does not generally lead to an identifiable reward function using~\eqref{def:utility_RLHF} by presenting three concrete scenarios for which identification is not possible.
The root cause of the non-identifiability is the usage of a trajectory's return as the utility in~\eqref{def:utility_RLHF}.
They therefore suggest using a trajectory's regret as an alternative, which provably leads to identifiable rewards.

A trajectory's regret is the negated sum of the optimal policy's advantage over each state-action pair in the trajectory.
Empirically, it has been shown that this modification improves the alignment of the learned strategy with human preferences.
The downside of this alternative is that regret depends on the unknown optimal policy.
Recently, it has also been suggested to consider $Q$-values of a human policy as the utilities~\citep{myers2023active}, while \citet{holladay2016active} used differences of cost functions that depend on the available choice set and the human's uncertainty.

\subsubsection{Human Feedback Models Beyond Boltzmann}\label{subsec:beyond_boltzmann}

While the human feedback model based on the Boltzmann distribution is the most widely used model nowadays, other models have also been considered in the literature.
In particular, for the probability in~\eqref{def:general_human_choice} other models such as the Thurstone model\footnote{The probit model, as used in \citet{biyik2020active}, corresponds to Thurstone's Case V.}~\citep{wilson2012bayesian,kupcsik2018learning,biyik2020active}, the ridge-noise model~\citep{schoenauer2014programming}, the binary model~\citep{sugiyama2012preferencelearning} or mixed forms thereof~\citep{wirth2016modelfree} have been considered.
Of these models, only the Thurstone model~\citep{thurstone1927law} has a similar interpretation as the Boltzmann distribution based on perturbed utilities, only differing in the distribution of the perturbance random variables.

\paragraph{Link functions} Another possibility, which is particularly popular in theoretical work on \gls{RLHF} (see  \cref{sec:theory}), is the use of other functions on the right-hand sides of \cref{def:Boltzmann_human_choice} than the exponential function.
The concept is primarily used for pairwise comparisons of trajectories.
It essentially states that the probability of the result of a pairwise comparison between two trajectories is the difference of their utility values under a so-called \emph{link function}.
More specifically, let $q = \{\tau_1,\tau_2\}$ be the query to compare the trajectories $\tau_1$ and $\tau_2$, then, assuming a link function $\Phi: \mathbb{R} \to [0,1]$, one models the probability in~\eqref{def:general_human_choice_RLHF} for $l$ representing a preference for $\tau_1$ as
\begin{align}
	\mathbb{P}\left( \mbox{label $l$ is provided} \, | \, m(q) \right)
	&= \mathbb{P}\left(  \tau_1 \succ \tau_2 \, | \, m(\{ \tau_1,\tau_2 \}) \right)
	= \Phi(u_{\tau_1} - u_{\tau_2})
	\,\text.
\end{align}
For $l$ representing a preference for $\tau_2$, one proceeds similarly.
The minimal assumptions on the link functions are that
\begin{itemize}
	\item [(i)] it is (strictly) monotonically increasing to take into account that trajectories with higher utilities will also have a higher chance to be picked;
	\item[(ii)] $\Phi(x) = 1 -\Phi(-x)$ to ensure that $\mathbb{P}\left(  \tau_1 \succ \tau_2 \, | \, m(\{ \tau_1,\tau_2 \}) \right) = 1 - \mathbb{P}\left(  \tau_1 \prec \tau_2 \, | \, m(\{ \tau_1,\tau_2 \}) \right)$.
\end{itemize}
Note that the second property implies $\Phi(0) = 1/2$ so that trajectories with the same utility also have the same chance of being selected.
Any cumulative distribution function of a symmetric continuous random variable fulfills these two conditions.
The two most common link functions that both fulfill the conditions are the linear link function given by
\[ \Phi(x) = \max{}\{ 0, \min{}\{ 1, 1/2 \cdot(1 + x) \} \} \]
and the logistic link function given by
\[
	\Phi(x) =  \frac{1}{1 + \exp(-x)}
	\,\text.
\]
Both are cumulative distribution functions:
The linear link function is the cumulative distribution function of a continuous uniform distribution on $[0,1]$.
In contrast, the logistic link function is the cumulative distribution function of a logistic distribution with location parameter 0 and scale parameter 1.
Moreover, both are intensively studied in theoretical approaches (see \cref{subsec:policy_learning}), and the latter leads to~\eqref{def:Boltzmann_human_choice} (when restricted on pairwise comparisons) and is a special case of the softmax function.

\paragraph{Two-Staged Choice Model}

\Citet{bobu2020less} propose the Limiting Errors due to Similar Selections (LESS) model that is inspired by the attribute rule model suggested by \citet{gul2014random}.
It assumes a feature map for trajectories and a (similarity) function mapping trajectory features and trajectories to integers and uses a two-stage process for modeling the human feedback (or choice):
First, choosing a trajectory feature according to the Boltzmann distribution and then a trajectory with the (logarithmic) similarity functions as the utilities within the Boltzmann distribution.
Their experiments show that this model can capture human feedback more appropriately than the standard Boltzmann distribution.

\paragraph{Generative Model}
\Citet{abramson2022improving} evaluate the usage of a generative model for learning from human preferences.
More specifically, instead of assuming some underlying utility as in the Bradley-Terry model, they attempt to train a model to generate the human feedback (inter-temporal preferences in this case, see \cref{sec:feedback-classes-intertemporal}) and directly interpret this feedback as reward.
However, they found that this empirically does not perform as well as the inter-temporal Bradley-Terry model.

\subsubsection{Misspecification}\label{subsec:misspecification}

The human feedback model may be misspecified in various ways.
\Citet{milli2020literal} investigate the problem of misspecifying the nature of human feedback that can be either literal or pedagogical.
The former means that the human gives targeted feedback for solving the actual \gls{RL} problem, while the latter means that the human gives targeted feedback that is deemed helpful for the learner.
They show theoretically and empirically that the case of a learner assuming a pedagogical feedback with an actual literal human always performs worse than the reversed case, i.e., a learner assuming a literal feedback with an actual pedagogical human.

A related question is studied by \citet{freedman2021choice}, namely, what if the learner makes incorrect assumptions about the choices from which the human selects its feedback?
They consider different types of such choice set misspecification and show that depending on the type of misspecification, the performances might vary drastically, even leading to no losses at all in some specific cases.

In the field of inverse \gls{RL}, the general question of the robustness of reward learning in terms of a misspecified human feedback model is theoretically investigated by \citet{skalse2023misspecification}.
It turns out that the optimality model is not robust with respect to any misspecification, the Boltzmann model is robust for quite a range of specific misspecifications, and the degree of robustness of the maximal causal entropy model lies between the latter two.
Even though these results are primarily derived for inverse \gls{RL}, they also have similar immediate implications for \gls{RLHF}.

\subsubsection{Diverse Preferences}\label{subsec:diverse_preferences}

One potential issue with the \gls{RLHF} framework is that it does not specify whose preferences to align to.
It is common to request feedback from multiple labelers in a crowd-sourcing setting, in which case the different labelers may disagree.
There are two main ways to deal with this challenge:
Either trying to learn each labeler's preference separately,
or trying to learn a model of the group's mean preference.

\Citet{bakker2022finetuning} investigate the first option by proposing to learn multiple reward functions, which can then be aggregated in arbitrary manners and even be used to find consensus among people with different preferences.
The second is more commonly used, however:
\Citet{xue2024reinforcement} learn a single reward function from multiple humans who may give diverse and inconsistent feedback, aiming to stabilize learning in spite of these inconsistencies using regularization, a consistency constraint, and ensembling.
Similarly, \cite{chhan2024crowdprefrl} try to estimate the correct preference for pairwise trajectories directly by combining the users' expressed preference labels instead of assuming individual reward functions.
As a middle-ground between the single reward function and multiple ones, \citet{myers2021learning} propose to learn a multimodal reward function that captures multiple people's preferences and use a mixture model of Plackett-Luce models to represent the feedback more accurately.

With a stronger focus on the active retrieval of human feedback, \citet{freedman2023active}
model the problem of selecting a suitable human labeler as a variant of the multi-armed bandit problem~\citep{lattimore2020bandit}, which they call hidden-utility bandit.
In this variant, the agent has in each decision round the choice between two options:
(i) drawing a bandit arm, then receiving a hidden arm-dependent utility, and finally observing an item, or
(ii) querying a human to observe a preference between two sampled items and incurring a human-specific query cost.
The feedback mechanism of all human teachers is modeled via a same Boltzmann distribution, differing only in their known individual rationality coefficients.
The same modeling of human feedback is also considered by \citet{barnett2023active}, who, however, use a Bayesian approach to determine which person should be queried in order to obtain the most informative feedback in expectation.
\citet{daniels-koch2022expertise} %
investigate the rationality coefficient already considered in the previously mentioned work and model it as a query-dependent function that might differ for the human labelers.

When dealing with diverse preferences, aggregating them via the mean may not always be suitable.
Instead, after learning the individual preferences, one needs to address the important question of preference aggregation, which is studied in social choice theory \citep{brandt2016handbook}.
Thus, a recent rapidly-developing research trend has started to investigate the tools developed in social choice in AI alignment \citep{conitzer2024position,sorensen2024position,mishra2023ai} in general and in RLHF \citep{dai2024mapping,zhong2024provable,park2024rlhf,chakraborty2024maxminrlhf,swamy2024minimaximalist,lambert2023history} in particular.

\subsubsection{Relaxation of the Markov Assumption}

Most works assume that the human feedback is given based on a latent Markovian reward model, i.e., the return of a trajectory $\tau$ decomposes into a sum of independent rewards over state-action pairs (see~\eqref{def:return}).
\Citet{early2022nonmarkovian} relax this assumption by dropping the need for the Markov property, such that the instantaneous reward might depend on hidden states.
Similarly, \citet{kim2023preference} avoid the Markov assumption by using a transformer as the preference model.
A similar effect may be achieved by learning a state representation with a recurrent network in which the rewards are Markov, similar to the approach taken by \citet{hafner2023mastering}, but we are not aware of any work exploring this.
\Citet{abramson2022improving} work in a non-Markovian setting as well by using memory-augmented networks for both the policy and the reward model.

\subsection{Utility Learning}

After choosing a human model to relate feedback to utilities, we can use the observed feedback to recover the latent utilities.
This utility learning can be reduced to a standard supervised learning problem and, therefore, is commonly solved with the techniques of empirical risk minimization or Bayesian approaches, both of which will be discussed in the following.

\subsubsection{Empirical Risk Minimization}

The most prevalent variant for learning the reward function, already presented in \cref{subsec:reward_learning_basic}, is a special case of empirical risk minimization.
The general approach of empirical risk minimization for reward function learning, assuming an underlying human feedback model with utilities  as in~\eqref{def:utility_RLHF}  is to find the minimizer of
\begin{align}\label{eq:ERM}
	\mathcal{L}(R;\mathcal{D}) = \sum_{i=1}^N \ell\left(u_{\cdot}(R),l_i,m(q_i)\right)
	\,\text,
\end{align}
where $\mathcal{D} = {\{(l_i,q_i)\}}_{i=1}^N$ is the given data set of observed label and query pairs,
$\ell: \mathbb{R} \times \mathcal{Q} \times \mathcal{C}$ is a suitable loss function with $\mathcal{Q}$ being the set of all possible label sets,
and $u_{\cdot}(R)$ denoting the utility (depending on the return $R$) of the possible labels for the given label-query pair $l_i,m(q_i)$.
As an illustration, consider the common case of pairwise trajectory comparisons where queries are pairs of trajectories $q_i=\{\tau_1^{i},\tau_2^{i}\}$, and labels $l_i \in \{ \tau_1^{i} \succ \tau_2^{i} , \tau_1^{i} \prec \tau_2^{i} \} = m(q_i) $ are the human's preference over the two trajectories.
For a given query \( q_i \), we then obtain~\eqref{eq:ml} as a special case of~\eqref{eq:ERM} by using the loss function:
\begin{align*}
	\ell(u_{\cdot}(R),l_i,m(q_i))
	&= - \log \left( \frac{1}{1 + \exp( u_{m(q_i) \setminus l_i}(R) -  u_{l_i}(R))} \right) \\
	&= - \log \left( \frac{1}{1 + \exp( \mathbb{E}_{\tau \sim G(m(q_i) \setminus l_i)}[R(\tau)] - \mathbb{E}_{\tau \sim G(l_i)}[R(\tau)]   )  } \right)
	\,\text,
\end{align*}
where in the case of pairwise comparison, $u_{\cdot}(R) = (u_{l_i}(R), u_{m(q_i) \setminus l_i}(R))$ and the grounding function $G$ is the projection onto the preferred trajectory.
This is the negative log-likelihood for the Boltzmann distribution for the observational pair $(l_i,m(q_i))$.

For the entire learning process, a model class $\mathcal{R}$ is assumed for the reward function $R$.
This model class is usually a parameterized class of functions, such as, for example, the class of linear reward functions~\citep{katz2021preferencebased}
\[  \mathcal{R} = \{ R_{\bm{\psi}}(s,a) = \bm{\psi}^\top \bm\phi(s,a) \, | \, (s,a) \in \mathcal{S} \times \mathcal{A}, \bm{\psi} \in \mathbb{R}^d \},  \]
where $\bm\phi: \mathcal{S} \times \mathcal{A} \to \mathbb R^d$ is some state-action feature mapping.
This entails that good features are known in advance such that rewards can be expressed as a linear combination of those features.
Using such linear models may lead to reward model misspecification.
Studying this setting, \citet{bobu2020quantifying} propose to adapt the hyperparameter $\beta$ in \eqref{eq:bt model beta} to account for this issue.

Since the assumption of a linear reward model may be too strong in practice, most recent work is based on non-linear models, especially using differentiable models, but other cases have been investigated as well.
In the latter case, for instance, decision trees have been considered to learn an interpretable reward model (see \cref{subec:interpretable}).
In the former case, simple \gls{MLP} has naturally been considered, but more recent deep learning architectures are more commonly used in the recent literature.
Thus, especially with partially observable domains, the reward network may be composed of a state-action encoder followed by fully connected layers.
For instance, \citet{abramson2022improving} combine ResNet blocks for image processing, a learnable embedding table, a multi-modal transformer, LSTMs, and \glspl{MLP}.
Besides, \citet{kim2023preference} use a Transformer-based architecture~\citep{vaswani2017attention}, motivated by the observation that rewards are often non-Markovian.

In addition to the usual empirical risk in~\eqref{eq:ERM}, it is also typical, as in supervised \gls{ML}, to add a regularization function to prevent overfitting:
\begin{align}
	\mathcal{L}(R_{\bm\psi};\mathcal{D}) = \sum_{i=1}^N \ell(u_{l_i}(R_{\bm{\psi}}),l_i, m(q_i)) + \lambda_r(\bm{\psi})
	\,\text,
\end{align}
where $\lambda_r: \Psi \to \mathbb{R}_+$ is a regularization function defined on the parameter space $\Psi$ of the underlying reward model class.
For instance, \citet{christiano2017deep} simply use L2 regularization and also consider dropout in some domains.
Recently, \citet{verma2024hindsight} propose to define a more complex regularization term, which consists in biasing the learned rewards to be proportional to an approximate state importance provided by a trained Transformer-based forward model.

The main supervised loss to train the reward model can also be augmented with additional losses corresponding to auxiliary tasks to avoid overfitting and improve generalizability.
For instance, \citet{abramson2022improving} use a behavior cloning loss and add a policy head to the reward network, thereby preventing that the reward model drifts too far from its initialization from the policy.
\citet{metcalf2023sampleefficient} design a reward model using state-action representations trained to be temporally consistent via self-supervised learning.
On a related note, the scalar preference optimization problem has been extended to a multidimensional one by \citet{zhong2024panacea} to represent diverse human preferences and \citet{marta2023aligning} for query efficiency.

\subsubsection{Bayesian Approach}

As is generally the case in supervised \gls{ML}, there is also the variant of using Bayesian modeling for learning a target object instead of the (empirical) minimization of a loss function.
To this end, one starts with a prior distribution $\rho$  over the parameter space of the reward function that is updated in light of the data set $\mathcal{D}$ by means of Bayes theorem:
\begin{align*}
	\mathbb P(\bm\psi \, | \, \mathcal{D}) \propto L_{\bm\psi}(\mathcal{D}) \cdot \rho(\bm\psi)
	\,\text,
\end{align*}
where $L_{\bm\psi}(\mathcal{D}) = \prod_{i=1}^N \mathbb{P}\left( \mbox{$l_i$ is provided} \, | \, m(q_i) , {\bm\psi} \right)$
is the likelihood of the data under the assumed human feedback model with reward function $R_{\bm\psi}$.
Such an approach is used for pairwise trajectory comparisons, for instance, by \citet{schoenauer2014programming} for the noisy-ridge model or by \citet{sadigh2017active} for the Boltzmann distribution as the human feedback model.
In inverse \gls{RL}, such Bayesian approaches have been considered as well (see Section 4.3 in \citet{arora2021survey}).

Instead of assuming that the reward functions are parameterized, one can use the reward functions directly as a parameter class and use a prior distribution over them.
This could, for example, be a Gaussian process as initially considered by \citet{kupcsik2018learning} for pairwise trajectory comparisons and adapted in later works~\citep{biyik2020active,cosner2022safetyaware}.
Here, again, it is worth mentioning that such considerations have also been made in inverse \gls{RL} before (see Section 4.3 in \citet{arora2021survey}).

\subsubsection{Partial Identifiability}
A crucial question when it comes to learning the reward function is whether the reward function can be identified at all.
If two reward functions induce exactly the same human feedback model, the reward function is called partially identifiable or ambiguous.
\Citet{skalse2023invariance} study this topic for the Boltzmann distribution as the underlying human feedback model when demonstrations (inverse \gls{RL}) or pairwise trajectory preferences are given as feedback.
For demonstrations, this question has also been examined in other works~\citep{ng2000algorithms,dvijotham2010inverse,kim2021reward,cao2021identifiability}.
On a related note, \citet{ellis2024generalized} tackle this identifiability issue by considering suitable acquisition functions (see \cref{sec:labelcollection-activelearning}).

\subsubsection{Interpretability}\label{subec:interpretable}

The field of explainable artificial intelligence (XAI) has emerged in recent years to improve the transparency and explainability of models or even to enable them in the first place.
Roughly speaking, the aim is to resort to more interpretable methods or provide explanations for both experts and non-experts, shedding light on why a certain input in a (black box) model leads to a certain result.
Explanations can take different forms, as can the ways to ensure the transparency of models, and for a detailed overview, we refer to \citet{barredoarrieta2020explainable}.
It is worth noting that the field has grown so extensively over the years that even dedicated overviews for the field of interpretable and explainable \gls{RL} are by now available~\citep{puiutta2020explainable,qing2023survey,milani2023explainable,glanois2024survey}.

For the branch of \gls{RLHF}, the existing works are quite sparse and mostly limited to using tree models as transparent and explainable models for learning the reward function~\citep{bewley2022interpretable,bewley2022reward,kalra2022interpretable,bewley2024learning,kalra2024can}.
Another way to realize explainability within \gls{RLHF} suggested by \citet{zhang2023learning} is to learn simultaneously the reward function and the importance of states using a weight network.
Assuming that for (long) trajectories, only a few states are important for the preference outcome, their framework can be used to select samples for explainability purposes.
Moreover, a perturbation analysis is suggested to evaluate explanations in a quantitative manner using the learned state importance weights.

\subsubsection{Online Improvements}\label{sec:online-improvements}

\Citet{christiano2017deep} demonstrate that it is important to improve the reward model online, a finding that has been confirmed by subsequent works such as the one by
\citet{gao2023scaling}, which empirically demonstrates that overoptimization of a reward model trained offline leads to performance degradation.
Without online improvements, issues of overoptimization of an imperfect reward model may occur.
\Citet{abramson2022improving} give an example of this:
They attempt to fine-tune an agent initialized with behavioral cloning with an engineered reward function and find that it fails to generalize and actually worsens the performance.
They also compare \gls{RLHF} with a reward model trained offline with iterative improvement and find that iterative improvement leads to better performance, even sometimes exceeding human performance.

This is related to issues posed by the approximate nature of the reward model in general, discussed in further detail in \cref{par:approximate_rewards}, but improving reward model accuracy, in general, is not sufficient:
\Citet{mckinney2022fragility} further show the interdependence of the reward model and the policy, demonstrating that reward models trained online together with a policy may not be effective when a completely new policy is trained.

Solutions to the problems of overoptimization and interdependence can take different forms:
One is to update the reward model online with sufficient frequency using notably more on-policy queries (see \Cref{sec:labelcollection-activelearning}),
another is to improve the reward model, e.g., by leveraging ensembles \citep{eisenstein2024helping,coste2024reward}, combining multiple reward models with weight averaging\citep{rame2024warm}, or by modifying the training procedure to place additional emphasis on challenging examples \citep{zheng2024improving},
and a third, discussed in \cref{par:approximate_rewards}, is to add constraints to the policy training.

\subsubsection{Learning from Multiple Feedback Types}\label{sec:learning-from-multiple-feedback-types}

As discussed in \cref{sec:combination-of-feedback-types}, it is often desirable to combine several feedback types.
This requires extensions of the learning process to incorporate different sources of feedback.
Learning from multiple feedback types can be achieved by pre-processing the feedback, assuming common latent factors, or by using the feedback types for distinct purposes.

The first approach is demonstrated by \citep{novoseller2023diprl}, who infer preferences from demonstrations, allowing them to treat both types of feedback equally in the learning pipeline.
In the style of the second approach, \citet{jeon2020rewardrational} proposes the unified framework of reward-rational choice, which allows for interpreting many forms of human feedback as Boltzmann-rational choices and, through this common framework, enables combination and adaptive selection of feedback types.
Finally, different types of feedback can be used for entirely different purposes, such as one for objective learning and another for safety-constraints~\citep{cosner2022safetyaware} or for representation learning (see \cref{sec:feedback-commonclasses-representationfocused}).

Since multiple sources of reward information may conflict, it is important to consider how to combine them.
\Citet{krasheninnikov2021combining} study several possible strategies of combining several reward functions in this setting, relating it to multi-task inverse \gls{RL}.
Note that this challenge of conflicting sources of reward relates tightly to challenges posed when receiving diverse preferences from different labelers, as discussed in \cref{subsec:diverse_preferences}.

\subsubsection{Offline Reward Learning}

There is a recent trend towards offline \gls{RLHF}, where both the reward model and the policy are trained offline.
The offline setting is also frequently considered in \gls{RLHF} theory (\cref{sec:theory}).
Early approaches in this area \citep{kim2023preference,shin2023benchmarks} first generate queries from an offline dataset of behaviors, gather human responses, train a reward model from the resulting preference data, and then leverage offline \gls{RL} algorithms to derive a policy.
We do not cover these works in detail, since this survey primarily focuses on the interactive and online setting (see \cref{sec:scope-of-the-survey}).
Nonetheless, the offline setting is particularly useful for evaluating novel approaches, e.g., for active query selection, using offline datasets.
We refer the interested readers to \cref{sec:datasets} for a discussion of available datasets.

Note that offline reward learning does not necessarily need to be combined with offline policy learning, as the reward model can be used to train a policy online as well.
This can be problematic in practice, as the distribution of experiences will generally shift during online policy learning, which can lead to poor performance of the reward model.
Nonetheless, this setup is commonly used in the LLM finetuning setting, since distributional shift is less of a concern due to the large and diverse training data (often collected off-policy) and the bandit-like nature of the task (compare \cref{sec:rlhf-for-llms-bandit}).
In this setting, it is also common to directly integrate preference and policy learning through \gls{DPO} techniques further discussed in \cref{subsec:direct-methods}.

\subsection{Evaluating Learned Reward Functions}\label{sec:rewardmodel-utilitylearning-evaluating}

A central question when it comes to learning the reward function is how to evaluate the learned reward function and how reward functions can be compared with each other.
For this purpose, different approaches are available, e.g.:

\begin{description}
	\item[Rollout Method]
	In inverse \gls{RL}, a common method for evaluation is the rollout method~\citep{ng1999policy,fu2018learning,ibarz2018reward,brown2019extrapolating}.
	In this approach, one first learns an optimal policy for the learned reward function and then estimates the value of this policy for online trajectory rollouts using the known ground-truth reward function.
	This approach can be transferred to \gls{RLHF} as well.
	In many cases, however, especially in safety-critical areas such as medicine or autonomous driving, such online rollouts cannot be executed.

	\item[Off-policy Evaluations]
	When online rollouts are not possible, so-called off-policy evaluations, which estimate the value of the optimal policy based on an available data set, may be considered.
	For coping with biases or large variances due to policy mismatch, approaches using importance sampling~\citep{precup2000eligibility}, regression- or classification-based methods~\citep{paduraru2013offpolicy,le2019batch,irpan2019offpolicy}, or combinations of these~\citep{jiang2016doubly} have been proposed.
	Off-policy evaluation is particularly well-suited for evaluating \glspl{LLM} reward models. \Citet{lambert2024rewardbench} propose a particularly large-scale dataset for this purpose, focusing on examples that are difficult to judge. They then use accuracy-based metrics across multiple categories to evaluate reward models.
	The problem with these approaches, however, is that the traces of the explicit sources of error through policy learning or reward learning are blurred, and that some require access to the ground-truth rewards.

	\item[Distance Functions]
	Yet another alternative, which has been advanced in the seminal paper by \citet{gleave2022quantifying}, is using a suitable distance function for reward functions.
	Suitable here means that two reward functions, which differ only by certain transformations such as potential shaping~\citep{ng2000algorithms} or positive scaling, should have zero distance if these transformations do not change the policy ranking with regard to the expected return.
	For this purpose, \citet{gleave2022quantifying} present a pseudometric, called Equivalent-Policy Invariant Comparison (EPIC) distance, that is determined in three steps: First, mapping two reward functions to a so-called \emph{canonicalization form} that is invariant to transformations of the latter kind.
	Second, normalizing these canonicalization forms by means of a specific weighted $L_2$-norm whose weights are determined by a distribution over the transitions.
	Finally, the EPIC distance is the weighted $L_2$-norm distance of the normalized canonicalization forms.

	Even if some attractive properties, above all a Lipschitz continuity in terms of the EPIC distance of two reward functions for the difference of the value functions of the induced optimal policies is shown, this distance has its shortcomings.
	One shortcoming is that the canonicalization form used by EPIC distance does not encode sufficient knowledge about the underlying transition dynamics, which might lead to unreliable distances when evaluating reward functions on physically non-realizable transitions.
	To this end, \citet{wulfe2022dynamicsaware} propose the Dynamics-Aware Reward Distance (DARD), which uses a slightly different form of canonicalization but restricts the evaluation of the reward functions to transitions that are approximately physically feasible.

	Recently, EPIC-like distances~\citep{jenner2022general} and STAndardised Reward Comparison (STARC) metrics~\citep{skalse2024starc}, which are entire classes of pseudometrics on the space of all reward functions were proposed that generalize the three-step approach underlying the EPIC distance (and DARD) by parameterizing each of the steps.
	Specifically, the canonicalization function in the first step, the normalization in the second, and the metric in the third step are kept variable.
	If these three functional parameters fulfill certain requirements, then the resulting distance has some appealing properties, e.g., being a pseudometric that is zero if and only if the two reward functions induce the same ordering of policies or imply upper and lower bounds on value function differences.
    In particular, these metrics retain the flexibility of DARD (in terms of specifying transition dynamics), while at the same time preserving the theoretical justification of EPIC.

	\item[Visual and Human Inspection]
	For an evaluation by visual inspection, \citet{jenner2021preprocessing} propose a method for preprocessing reward functions by transforming them into simpler but equivalent reward functions for better interpretability.
	Related to this and the rollout method, the quality of the reward function learned can also be evaluated by a human (or expert) by examining the behavior of the agent on the target task.
	This can even extended to the \gls{RLAIF} setting, by using another trained model as the evaluator \citep{li2024generative}.
 Besides, in the context of LLMs, datasets have been proposed and specifically designed to evaluate the (in-)consistency of learned reward models with respect to semantic changes of prompts \citep{shen2024trickledown}. %
\end{description}

\subsection{Reward Model Inputs}\label{sec:reward-model-inputs}

Besides the feedback type, another factor is the modality of the reward model input data.
This usually consists of the agent's observations and actions.
Observations can range from true state to high-dimensional inputs (e.g., images), while
actions can range from discrete finite actions to continuous actions.

For instance, many typical \gls{RL} benchmarks are in the continuous control domain (e.g., MuJoCo simulated robotics tasks) with true state representations and simple discrete actions.
In such problems, \citet{christiano2017deep} train reward models from these inputs.

When no compact state representation is available, raw images are often used in control tasks, which makes the learning of rewards more challenging since the setting becomes partially observable and the reward function is generally not Markov with respect to the observations.
In such cases, the conventional trick of approximating a true state with a sequence of frames is often employed.
This approach is used, for instance, by \citet{christiano2017deep} to train reward models on the Atari benchmark suite.
When taking only observations as inputs, one can resort to recurrent models \citep{abramson2022improving} or Transformer-based models \citep{kim2023preference}.

More recently, many applications of \gls{RLHF} are in the \gls{NLP} domain.
In these settings, the policy takes natural language as both input and output while the reward model takes it as input (see, e.g., the work by \citet{ouyang2022training}).
Naturally, more complex scenarios (e.g., with both language and vision inputs~\citep{abramson2022improving}) have also been studied.

\subsection{Increasing Feedback Efficiency}

Maximizing feedback efficiency is vital in \gls{RLHF} due to the high cost of human feedback.
This section delves into methods that enhance learning from limited human feedback.
We discuss methods that leverage prior offline data, methods that use (partially unlabeled) data more efficiently, and methods that aim to gather more informative data.

\subsubsection{Using Prior Data}\label{sec:prior-data}

There are often large amounts of prior data available at little or no additional cost.
While this data generally was generated for other tasks, many basic human preferences are the same for various tasks and can often even be extracted from completely unsupervised data such as text corpora.
By leveraging this prior data, we can greatly reduce the amount of feedback necessary to learn the current task's objective.
We explore various methods, including meta- and transfer learning, leveraging foundation models, reward model initialization, preference model pretraining, and supervised representation learning.

\paragraph{Meta- and Transfer Learning}

Meta- and transfer learning techniques in reward model training exploit the commonalities across different objectives, facilitating quick adaptation to new tasks.
\Citet{ren2022efficient} develop a broadly applicable meta-reward model, pre-trained on a diverse set of tasks to capture a wide range of preference patterns, enabling efficient adaptation to new tasks with fewer examples.
\Citet{xie2018fewshot} use a similar meta-learning approach to build a goal classifier across multiple visuomotor tasks.
Closely related to these meta-learning approaches, \citet{hejna2022fewshot} integrate few-shot learning principles, optimizing their approach for scenarios where only a few examples are available for adapting to new tasks.
In the domain of transfer learning, \citet{liu2023zeroshot}
explore zero-shot transfer of preferences, a method that enables adapting preferences without additional data from the new task.
In a different vein, but closely related to meta- and transfer learning, \citet{mendez2018lifelong} tackle the lifelong inverse \gls{RL} problem, focusing on inferring reward functions for multiple tasks over time, which involves knowledge transfer between tasks.
Collectively, these studies underscore the potential of meta- and transfer learning in enhancing the efficiency and applicability of reward models in \gls{RLHF}.

\paragraph{Leveraging Foundation Models}

Foundation models, i.e., large models trained on large amounts of often unlabeled data, can acquire significant knowledge about basic human preferences.
A language model trained to predict the next token in a text corpus, for example, may learn to complete the sentence `Frank was mad that his vacuum robot broke the vase', thereby learning that humans prefer non-destructive behavior.
These learned preferences can then be leveraged in \gls{RLHF} approaches.
For instance, \Citet{kwon2023reward} propose to use \gls{LLM} as a source of rewards.
\Citet{du2023visionlanguage} is another example, where a success detector is trained using a pre-trained vision-language model (Flamingo).
Their approach uses a dataset of trajectories with binary success labels, employing a non-interactive training method.

\paragraph{Reward Model Initialization}\label{par:reward-model-initialization}

It is often beneficial to initialize the reward model with parameters from a model trained on a related task.
This strategy is particularly common in language model fine-tuning, where self-supervised pretraining is a common practice.
In such scenarios, it becomes logical to use these pre-trained models for initializing not just the policy but also the reward model.
This methodology is adopted by \citet{askell2021general} and \citet{ouyang2022training}.
Specifically, \citet{ouyang2022training} use a pretrained language model for the reward model, opting for a smaller model relative to the policy to mitigate unstable learning.
Notably, while they apply supervised fine-tuning to the policy before the \gls{RLHF} phase, the reward model is initialized directly from the language model without any preliminary fine-tuning.
This approach's applicability extends beyond language models to other areas.
A notable example is \citet{abramson2022improving}, who, in the control domain, begin by training a policy using contrastive self-supervised learning and behavioral cloning.
They then add an \gls{MLP} head to the policy for the prediction of cumulative rewards.

\paragraph{Reward Model Pretraining}\label{par:reward-model-pretraining}

Reward model pretraining~\citep{askell2021general,bai2022training} leverages prior offline data to pretrain the preference model before training it on policy samples.
\Citet{askell2021general} note that in the case of language models, noisy preference data can be readily obtained from sources such as rated Reddit comments, preferred Stack Overflow answers, and reverted Wikipedia edits.
They leverage this as a pretraining step to increase data efficiency.
This is in addition to regular language model pretraining, as discussed in the previous paragraph.
A similar approach could be applied to control in case prior data and some means of inferring preferences, such as human corrections, are available.
Even if no inferred preferences are available, \citet{verma2023data} show that it can be beneficial to pre-train the preference model to predict close to constant reward on an initial set of trajectories.
This avoids excessive fitting of the policy to random initialization differences in the reward function.

\paragraph{Supervised Representation Learning}

A compact representation that captures all relevant information for human preferences while minimizing noise can greatly enhance preference learning efficiency.
It may also generalize better than a representation learned end-to-end as part of the preference learning task, which may contain spurious correlations.
\Citet{bobu2022inducing} address this by proposing the learning of features through explicit human feedback using feature traces.
Feature traces (see \cref{sec:feedback-commonclasses-featuretraces}) involve human labelers explicitly teaching relevant features one by one by demonstrating behavior in which the feature monotonically increases or decreases.
This method directly aligns the learned representation with human-identified features, enhancing preference learning efficiency but requiring detailed human input.
However, feature traces require labelers to be able to identify and articulate relevant features, which can be challenging.
\Citet{bobu2023sirl} offer an alternative approach with their Similarity-based Implicit Representation Learning (SIRL) method.
SIRL learns representations from similarity queries (see \cref{sec:feedback-commonclasses-similarityqueries}), where human labelers provide feedback on whether behaviors are similar or different concerning the features that matter to them.
This method captures a broader range of human notions of similarity without needing explicit feature knowledge, thus reducing the cognitive load on human labelers.
In summary, while both approaches emphasize human feedback's centrality in representation learning, they differ in their methods of gathering this feedback.
The feature traces used by \citet{bobu2022inducing} require specific feature knowledge, whereas SIRL used by \citet{bobu2023sirl} uses more intuitive similarity assessments, potentially offering a more user-friendly way to capture human preferences.

These diverse methods of using prior data demonstrate the potential for enhancing data efficiency in \gls{RLHF}, enabling more effective learning from limited human feedback.

\subsubsection{Using Data More Efficiently}\label{sec:using-data-more-efficiently}

Beyond the application of prior data, several techniques can enhance the efficiency of data utilization in training processes.
This section will discuss a range of such methods, including self-supervised and semi-supervised training, as well as the integration of inductive biases and data augmentation strategies.
These approaches are designed to make the most of the available human interactions and improve the final performance of \gls{RLHF} models.

\paragraph{Self-Supervised Auxiliary Tasks}

Self-supervised training enhances data efficiency in reward model training by using unannotated data to capture information about the task.
This technique extends beyond the scope of pretraining methods, as discussed in the prior section, to incorporating concurrent auxiliary tasks to maximize the utility of available data.
A prevalent technique, as applied by \citet{abramson2022improving}, \citet{brown2020safe}, and \citet{metcalf2023sampleefficient}, involves adding self-supervised losses to enhance representation learning for rewards.
\Citet{abramson2022improving} implement a contrastive task where the reward network differentiates between observations that are consistent between multiple modalities and those that are not, blending this with preference learning loss and behavioral cloning.
\Citet{brown2020safe} add multiple auxiliary tasks such as inverse and forward dynamics modeling, temporal distance prediction, and variational autoencoder training.
Similarly, \citet{metcalf2023sampleefficient} use the self-predictive representations technique~\citep{schwarzer2021dataefficient} to learn state representations that encode environmental dynamics, enabling a linear model to anticipate successor states, thereby forming an efficient basis for preference learning and significantly boosting sample efficiency.
However, auxiliary losses for better representation learning are not the only approach to leverage self-supervised training.
An alternate approach by \citet{verma2024hindsight} involves identifying important states using attention weights from a world model transformer and state importance estimates based on a preference predicting transformer.
These estimates can aid credit assignment for observed preferences, further optimizing the training process.

\paragraph{Semi-Supervised Training}

Semi-supervised training, blending labeled and unlabeled data, can leverage the unlabeled data to glean information about the task and the environment.
This is most commonly done by generating pseudo-labels for the unlabeled data, either by leveraging model predictions or by making assumptions.
The first approach is used by \citet{cao2021weak}, using supervised learning methods to predict human preferences, and \citet{zhan2021humanguided}, employing a GAN-based approach to learn and mimic human preference patterns.
Similarly, \citet{park2022surf} expand their data set with high-confidence unlabeled samples based on the preference predictor’s evaluations.
The second strategy, making assumptions to augment data, is showcased by \citet{zhou2020learning}.
They generate preference data by assuming that (i) human-written examples are better than model-written examples, (ii)  human-written and model-written examples are indistinguishable amongst themselves, and (iii) generations of later model iterations are better than those of earlier ones.
This is closely related to \gls{RLAIF}, where a pretrained language model is used to gather preference feedback, effectively leveraging the knowledge about human preferences learned by the pretrained language model through self-supervised learning.

\paragraph{Data Augmentation}

Data augmentation focuses on creating additional examples from existing labeled data.
Temporal augmentation is particularly effective in \gls{RLHF}, involving trajectory data.
This is exemplified by \citet{brown2019extrapolating} and \citet{park2022surf} who base their augmentation on the premise that preferences for complete trajectories can be extrapolated to cropped segments, allowing the generation of multiple derivative pairs from a single labeled trajectory pair.
\Citet{park2022surf} additionally explore state modifications, such as random re-scaling and Gaussian noise addition, finding temporal cropping to be the most effective, with noise sometimes negatively impacting performance.
In a similar vein, \citet{verma2023state} focus on augmenting trajectories by concentrating on changing elements in observations and perturbing the other parts, based on the premise that movement indicates importance in image-based observations.
Complementing these methods, \citet{abramson2022improving} employ augmentation by randomly altering instructions and language responses, thus creating artificial examples of non-preferred behavior.
These diverse data augmentation methods collectively enhance the training data set, contributing to the increased robustness and efficacy of \gls{RLHF} models.

Relatedly, Meta-Reward-Net \citep{liu2022metarewardnet} optimizes not only for the preference prediction accuracy of the learned reward function but also of the learned \( Q \) function in an actor-critic \gls{RL} algorithm.
This is beneficial since it avoids the phenomenon of confirmation bias, where one learned model (in this case the \( Q \) function) overfits to targets predicted by another model (the reward model).
It is not strictly a data augmentation technique, but closely related in practice.

\subsubsection{Gathering Better Data}\label{sec:gathering-better-data}

In addition to leveraging unlabeled data and using labels more efficiently, sample efficiency can be further increased by collecting more informative samples in the first place.
This can either be achieved by selecting more informative samples from the experience buffer or by generating more informative experiences.
While selecting informative samples from the experience buffer is addressed under active learning (see \cref{sec:labelcollection-activelearning}), this section focuses on generating more informative experiences.

While we are not aware of many works in this area, one possible approach involves steering the agent's exploration towards regions of the state space where human feedback would be most beneficial.
\Citet{liang2022reward} implement this by employing intrinsic motivation, driven by the estimated uncertainty of the reward model, to guide the agent's exploration.
This highlights the potential of not just using data more efficiently but also generating data in a more targeted manner.

\section{Policy Learning}\label{sec:policy-learning}

\begin{figure}[H]
	\centering
	\definecolor{bgcolor}{HTML}{3c8e4b}
\begin{tikzpicture}[%
	every text node part/.style={align=center},%
	>=stealth'%
]
	\node[inner sep=0,outer sep=0,text=white] (agent-title) {Agent};
	\node[rectangle,minimum width=3.50cm,below=0.3cm of agent-title,fill=white,text=black,draw=red!90!black,very thick] (policy) {\small Policy};
	\node[rectangle,minimum width=3.50cm,below=0.7cm of policy,fill=white,text=black] (rewardmodel) {\small Reward Model};
	\begin{scope}[on background layer]
		\node[rectangle,rounded corners,fit={(agent-title) (policy) (rewardmodel)}, inner ysep=0.2cm,fill=bgcolor] (agent) {};
	\end{scope}

	\node[inner sep=0,outer sep=0,text=white,right=3.5cm of agent-title] (environment-title) {Environment};
	\node[rectangle,minimum width=2.2cm,below=0.3cm of environment-title,fill=white,text=black] (dynamics) {\small Dynamics};
	\begin{scope}[on background layer]
		\node[rectangle,rounded corners,fit={(environment-title) (dynamics)}, inner ysep=0.2cm,fill=bgcolor] (environment) {};
	\end{scope}
	\draw[very thick,->,draw=red!90!black] (agent.east |- environment.180) to node[above=-0.05cm,pos=0.04,anchor=south west] {\footnotesize Action \( a_t \)} (environment.180);

	\coordinate (midway) at ($(environment.west |- policy) + (-1.45,0)$);
	\draw[thick,-|] (environment.west |- policy) -- (midway) node[below=-0.05cm,pos=0.46] {\footnotesize State \( s_{t+1} \)}; %
	\draw[thick,->] (midway) -- (policy -| agent.east) node[below=-0.05cm,pos=0.53] {\phantom{S}\footnotesize \( s_t \)};

	\draw[thick,->] (rewardmodel.70) to node[right=-0.05cm] {\color{white}\footnotesize Reward \( \hat r_{t + 1} \)} (policy.south -| rewardmodel.70);
	\draw[thick,->] (policy.south -| rewardmodel.110) to node[left=-0.05cm] {\color{white}\footnotesize Action \( a_t \)} (rewardmodel.110);

	\node[rectangle,minimum height=0.7cm,minimum width=2cm,rounded corners,below=1.15cm of environment.south,anchor=south,fill=bgcolor,text=white] (evaluator) {Labeler};

	\draw[thick,->] (rewardmodel.4) to[->] node[above=-0.05cm,midway] {\footnotesize Query \( q_i \)} (evaluator.west |- rewardmodel.4);
	\draw[thick,->] (evaluator.west |- rewardmodel.-4) to[->] node[below=-0.05cm,midway] {\footnotesize Label {\( l_i \)}} (rewardmodel.-4);
\end{tikzpicture}
	\caption{RLHF diagram highlighting components discussed in this section.}
\end{figure}

After learning a reward model, or, more commonly, interleaved with reward model learning, the next step is to train a policy that maximizes the expected accumulated reward.
This section will discuss algorithms for policy learning, which can be categorized into two main techniques:
adaptation of conventional \gls{RL} algorithms and \gls{DPO}.

\subsection{Adaptation of RL Algorithms}

Using the learned reward model, any standard \gls{RL} algorithm (e.g., DQN, A3C, PPO, SAC) could potentially be applied to train a policy.
However, in the setting of \gls{RLHF}, this direct application may suffer from two issues:
The non-stationary nature of the learned reward function in \gls{RLHF} and its inaccuracy, especially in the earlier stages of training.
We will discuss these issues and possible adaptations of \gls{RL} algorithms to address them in the following.

\paragraph{Non-Stationary Rewards}

\Gls{RL} algorithms are designed to learn a policy that maximizes the expected accumulated reward in an \gls{MDP} framework, which assumes a stationary reward function.
The \gls{RLHF} setting violates this assumption by periodically updating the reward model, leading to a non-stationary reward function.

Various works have empirically demonstrated that conventional \gls{RL} algorithms can be applied nonetheless, with little to no modification.
\Citet{christiano2017deep} argue that policy-gradient methods are better suited for non-stationary reward functions compared to value-based methods.
They and various follow-up works successfully apply policy-gradient methods without any modifications in this setting.
This approach has been picked up for language-model fine-tuning as well \citep{ouyang2022training}.

Later works have shown that value-based methods (possibly in an actor-critic scheme) can also be effective in \gls{RLHF} \citep{ibarz2018reward,lee2021pebble,park2022surf,liu2022metarewardnet,xue2023prefrec}.
One trick to make value-based methods work is to use the reward model to relabel the experiences in the replay buffer whenever it is updated \citep{ibarz2018reward,lee2021pebble}.
Similar to conventional \gls{RL}, the use of such a replay buffer can greatly decrease the amount of environment interactions necessary for successful learning.
As demonstrated by \citet{gulcehre2023reinforced}, the sample efficiency can be increased even further by using offline-\gls{RL} techniques in a growing-batch \gls{RL} setting, an offline-RL technique that iteratively increases the size of the dataset by policy rollouts while still being more sample-efficient than online \gls{RL}.

In addition to the basic \gls{RL} approaches, there are also some policy learning approaches tailored specifically for \gls{RLHF}.
\Citet{wu2023pairwise} propose a policy gradient algorithm, called Pairwise Proximal Policy Optimization (P3O), as an alternative to PPO, which avoids estimating the value function and at the same time is provably invariant with respect to equivalent rewards (unlike PPO).
In a similar vein, \citet{zhu2023finetuning}
replace the KL-regularization of PPO with a squared error term of the logarithmic probabilities, resulting in a seemingly more stable \gls{RL} learner.

\paragraph{Overoptimization of Approximate Rewards}\label{par:approximate_rewards} %

Since the learned reward model, which is only an approximation of the true reward function, is used to train a policy, overoptimization \citep{gao2023scaling} or reward hacking \citep{skalse2022defining} can happen.
\Cref{sec:online-improvements} discusses the interdependence of the reward model and the policy in more detail as well as possible improvements from the reward model side, while here we focus on how to adapt policy training to cope with possibly inaccurate rewards in general.

One approach is to regularize the policy so as not to diverge too much from human-given demonstrations using KL-divergence regularization \citep{jaques2017sequence}.
This is particularly common for language-model fine-tuning~\citep{ouyang2022training,abramson2022improving}, but \citet{abramson2022improving} explores this for control as well.
They found that this was important for some cases, in particular for deciding when to output language, but not for all.
Going beyond KL-regularization, \citet{moskovitz2024confronting} investigate several techniques of constrained \gls{RL} to only maximize rewards up to a threshold while avoiding excessive deviation from a pre-trained policy.

\subsection{Framing RLHF for Generative Models as a Bandit Problem}\label{sec:rlhf-for-llms-bandit}

So far, we have assumed that we ultimately want to solve a reinforcement learning problem represented by an MDP.
However, especially with regard to the application of \gls{RLHF} in the area of \glspl{LLM}, there is now another simplified way of looking at the problem.
Namely, as an instantiation of a (contextual) preference-based bandits problem  \citep{bengs2021preferencebased}, which can of course be modeled by the more general case of a Markov decision process (MDP).
In both cases, the focus is on the concept of tokens or rather sequences of tokens.
However, in the MDP point of view, the state space $\St$  consists of all previous tokens and the prompt (represented as a sequence of tokens), while the action space  $\Ac$ consists of all potential next tokens.
A terminal state is often indicated here by the special token $\texttt{<eos>}$ and trajectories are filled with this token until the maximum length $H$ is reached.
Moreover, the transition function $\T$ is degenerate (or deterministic), with a value of one only for the state that is the concatenation of the current state and the taken action.
A (latent) reward is only received at the end of the trajectory giving rise to a sparse feedback scenario.

The (contextual) preference-based bandits view, on the other hand, naturally considers no state space and no transition function, but an action space consisting of all possible responses to a prompt (both represented as a sequence of tokens).
Here, the prompt specifies the context for which at least two actions are executed and for which a qualitative comparison is observed as feedback.
In bandit literature, this is also referred to as a ``(multi-)duel'', coining the term \emph{dueling bandits}. %
Thus, this point of view takes a trajectory-wise perspective, while the MDP point of view takes a token-wise perspective.

Note that the bandit formulation considers an entire episode (response in the \gls{LLM} context) as an action with a single associated reward, resulting in sparse feedback from an MDP viewpoint.
This is in contrast to the standard \gls{RLHF} formulation as it is often used in control settings, where it is assumed that the reward of a trajectory is composed of the sum of the rewards of individual steps, which allows the optimizer to distribute rewards densely as best fits the data.
On an intuitive level, this leads to state-action pairs that often occur in preferred trajectories to be highly rewarded, without necessarily putting all reward on the terminal actions.
In practice, this can lead to nicely-shaped reward functions \citep{christiano2017deep}, which cannot directly be achieved in the bandit setting.
However, \citet{chan2024dense} show how to take advantage of the predominantly used transformer architecture for the reward model in order to obtain a denser reward, even when assuming the bandit setting:
More specifically, since the transformer architecture maintains attention weights in the last layer for each token,  these can be used to attribute the overall reward signal to individual tokens.

The main appeal of the bandit formulation is that since the environment's dynamics are deterministic, exploration is simplified (although the space of trajectories to be explored is still huge, i.e., exponential with respect to the trajectory length).
This setting even allows for policy improvements without any training using best-of-$n$-sampling (also called rejection sampling), a form of search that chooses the highest-rated response from multiple samples \citep{nakano2022webgpt,menick2022teaching}.
Most notably, however, this formulation notably enables direct policy learning approaches, such as \gls{DPO} \citep{rafailov2023direct} or $\Psi$PO \citep{azar2024general}, discussed in the following section.

\subsection{Direct Policy Optimization}\label{subsec:direct-methods}

The two-phase approach involving utility learning and policy optimization is not the only viable path to learning a policy from human feedback.
While we have previously discussed the case in which we learn a reward function from observed preferences by assuming a human feedback model, an emerging branch of the literature circumvents the reward-learning step and uses preferences directly to address the actual \gls{RL} problem.
One important representative algorithm in this direction is DPO~\citep{rafailov2023direct}.
The key trick in this method is to reparameterize the reward model as a function of its optimal policy, which allows the likelihood of the observed feedback (still using the Bradley-Terry model) to be rewritten as a function of the policy parameter.
Thus, a policy can be directly trained from pairwise comparisons by minimizing the following negative log likelihood computed on a dataset $\mathcal D$ of triplets $(x, y_w, y_l)$ with $x$ being a context (e.g., prompt to LLM) and $y_w, y_l$ being two outputs (e.g., generated answers) such that $y_w \succ y_l$:
\begin{equation}
    \mathcal{L}_{DPO}(\pi_\theta, \pi_{\mathrm{ref}}) =
    -\mathbb{E}_{(x, y_w, y_l) \in \mathcal D}
    \left[
    \log \sigma\left(\beta \log \frac{\pi_\theta(y_w \mid x)}{\pi_{\mathrm{ref}}(y_w \mid x)} -
    \beta \log \frac{\pi_\theta(y_l \mid x)}{\pi_{\mathrm{ref}}(y_l \mid x)}\right)
    \right]
\end{equation}
where $\sigma$ is the logistic function and $\beta$ is a hyperparameter controlling the deviation with respect to a reference policy $\pi_{\mathrm{ref}}$ (e.g., pretrained LLM).
This loss amounts to increasing the probability of generating better outputs while decreasing that of generating worse ones with respect to the reference policy.

Many improvements, variations, or alternatives to DPO have been proposed.
Since those methods are not the main focus of our survey, we refer the interested readers to their respective papers for more details, e.g.,

SLiC-HF~\citep{zhao2023slichf}, OPPO~\citep{kang2023reward}, DPPO~\citep{an2023direct}, PRO~\citep{song2024preference}, RSO~\citep{liu2024statistical},
SRPO~\citep{choi2025selfimproving}, or by formulating policy search as a zeroth-order optimization \citep{tang2024zerothorder}.
\Citet{azar2024general} introduce an objective called $\Psi$-preference optimization ($\Psi$PO) that unifies the objective functions in \gls{DPO} and \gls{RLHF}.
More specifically, for a specific instantiation of $\Psi$, the objective in $\Psi$PO recovers \gls{DPO} and SLiC-HF.
In addition, \gls{DPO} has been further generalized to include diverse divergence constraints \citep{wang2024reverse}.
Besides, \citet{hejna2024contrastive} propose contrastive preference learning based on a regret preference model instead of the usual one in \gls{RLHF}.
It is also possible to learn a \( Q \) function from human preferences directly, which implies a policy without the need for separate policy- and reward-model training \citep{myers2023active}.

It is worth noting that approaches for directly learning the policy from preferences have been considered in the past as well~\citep{wilson2012bayesian,furnkranz2012preferencebased,wirth2013policy,wirth2013epmc,busa-fekete2014preferencebased}.
In Sections 3.2.1 and 3.2.2 in the survey by \citet{wirth2017survey}, these methods are explained in more detail.

Another recent trend in fine-tuning models with human feedback is to even manage it without the usage of RL.
An alternative is based on supervised reward learning with new types of loss functions \citep{lee2023aligning,yuan2023rrhf} or a specific learning process   \citep{dong2023raft,korbak2023pretraining}.
There are also RL-free approaches that do not use a reward model to train a policy to execute natural-language instructions using a transformer
architecture \citep{brohan2023rt1,yu2023scaling}.
On a related note, \Citet{liu2024chain} suggest a way how to convert human feedback to natural language sentences for the task of fine-tuning language models.

\section{Theory}\label{sec:theory}

The field of \gls{RLHF} has recently made some progress in terms of theoretical results, which we will discuss in this section.
First, we consider the contributions where the goal is to learn a provably (near) optimal policy both in an online and offline fashion or even in a way that falls in between.
Then, we discuss and highlight recent contributions related to different theoretical aspects of \gls{RLHF}, such as its relation to the standard reward-based \gls{RL}.
\Cref{tab:overview_optimal_policy_learning_online,tab:overview_optimal_policy_learning_offline} provide a concise overview of the results for the online or offline policy learning setting.
Here, $\mathcal{N}_\mathcal{F}( \epsilon, d)$ denotes the $\epsilon$-covering number of a set $\mathcal{F}$ under some metric $d$\footnote{The $\epsilon$-covering number is the minimum integer $N$ such that there exists a subset $\mathcal{F}' \subset \mathcal{F}$ with $|\mathcal{F}'| = N$, and for any $f \in \mathcal{F}$, there exists some $f' \in \mathcal{F}'$ satisfying $d(f, f') \le \epsilon$.}.
It is worth mentioning that (almost) all works have two standard assumptions, namely that the reward function is bounded and that the ground-truth reward, human feedback model, or transition dynamics are elements of the considered model space, respectively.

\subsection{Policy Learning}\label{subsec:policy_learning}

In the literature focusing on theoretical results, there is a distinction (similar to the distinction made in standard \gls{RL}) between an offline and online setting.
In the former, learning is based on a given fixed data set, usually previously collected through an interaction with the environment.
In contrast, in the online environment, one interacts directly with the environment to learn from real-time feedback and continuously updates one's strategies based on the feedback received, allowing the agent to learn and adapt as it engages with the environment.
Accordingly, an important component of the online variant is the sampling procedure, i.e., how the labels are selected.
This is usually accomplished using an acquisition function that is based on uncertainty (see \cref{sec:labelcollection-activelearning}).

\begin{table}
	\newcommand{\tblrowsepa}{\arrayrulecolor{gray}\cmidrule(l{0.1em}r{0.1em}){1-5}\arrayrulecolor{black}}
	\footnotesize
	\caption{%
		An overview of approaches, their assumptions, goals, and properties for online policy learning.
		$\widetilde{\mathcal{O}}$ is used to hide $\log$-factors.
		$T$ is the number of iterations of the respective algorithm.
	}\label{tab:overview_optimal_policy_learning_online}
	\centering
	\hbadness=10000 %
	\hfuzz=\maxdimen{} %
	\begin{tabular}%
						{@{}p{2.25cm}p{3cm}p{3cm}p{2.5cm}p{4.25cm}@{}}
		\toprule
		\textbf{Algorithm} \newline \textbf{(Reference)}
		& \textbf{Algorithmic \newline approach}
		& \textbf{Assumptions}
		& \textbf{Target(s) and goal(s) of learner} %
		& \textbf{Theoretical guarantee(s)} \\ %
		\midrule
		Dueling Posterior Sampling (DPS) \newline \citep{novoseller2020dueling}
		& Leveraging Posterior Sampling from dueling bandits
		& Linear link function, tabular MDP
		& Bayes regret minimization w.r.t.\ optimal policy based on trajectory comparison feedback
		& Asymptotic regret rate: \newline
		  {\tiny$\mathcal{O} \left(|\mathcal{S}|\sqrt{|\mathcal{A}| T \log(|\mathcal{A}|)} \right) $ }
		\\
		\tblrowsepa{}
		Logistic Preference Reinforcement Learning (LPbRL)
		\newline \citep{saha2023dueling}
		& Leveraging MaxInP from contextual dueling bandits
		& Logistic link function, tabular MDP, linear rewards \& $d$-dimensional feature embedding of trajectories
		& Expected regret minimization w.r.t.\ optimal policy based on trajectory comparison feedback
		& Transition dynamics: \newline
		1.\ Known \newline {\tiny $\widetilde{\mathcal{O}} \left( |\mathcal{S}| H d \sqrt{ T \log(T)}  \right) $ } \newline
		2.\ Unknown \newline {\tiny $\widetilde{\mathcal{O}}((\sqrt{d} + H^2 + |\mathcal{S}|)\sqrt{dT} %
		\newline \hphantom{\widetilde{\mathcal{O}}(} +  \sqrt{|\mathcal{S}||\mathcal{A}|TH} ) $} %
		\\
		\tblrowsepa{}
		Preference-based Optimistic Planning (PbOP)
		\newline \citep{chen2022humanintheloop}
		& Leveraging MaxInP; general function approximation
		&
		General model classes $\mathcal{F}_{\mathbb{T}}$ for feedback and $\mathcal{F}_{\mathbb{P}}$ for transition dynamics with finite $l_2$-norm $\rho$-Eluder dimension $d_{\mathbb{T}}^{(2)}(\rho)$ and $d_{\mathbb{P}}^{(2)}(\rho)$.
		& High probability regret minimization w.r.t.\ optimal policy based on trajectory comparison feedback
		& {\tiny $ \widetilde{\mathcal{O}} \left( \sqrt{d_{\mathbb{P}}(\tfrac{1}{T}) H T \log\left( \mathcal{N}_{\mathcal{F}_{\mathbb{P}}}\left( \tfrac{1}{T},d \right)\right)} \right) \newline + \widetilde{\mathcal{O}} \left(\sqrt{d_{\mathbb{T}}(\tfrac{1}{T}) T \log\left(\mathcal{N}_{\mathcal{F}_{\mathbb{T}}} \left( \tfrac{1}{T}, d \right)  \right)} \right) $ \newline
			$d$ being the $\ell$-infinity norm ${\| \cdot \|}_{\infty}$}
		\\
		\tblrowsepa{}
		Preference-based Policy Search (PPS)
		\newline \citep{xu2020preferencebased}
		& Dynamic programming, policy search, $(\epsilon,\delta)$-PAC black-box dueling bandit algorithm and simulator
		& Uniform dependence of policy preference probabilities on value function differences, tabular MDP, $(\epsilon,\delta)$-PAC dueling bandit algorithm with $\Psi(K,\varepsilon, \delta)\varepsilon^{-\alpha}$ sample complexity for $K$ arms
		& $(\epsilon,\delta)$-PAC for optimal policy based on trajectory comparison feedback
		& Simulator step bound \newline
		{\tiny $\mathcal{O} \left(\frac{H^{\alpha+1} |\mathcal{S}| \Psi(|\mathcal{A}|, \varepsilon/H, \delta/|\mathcal{S}|)}{\varepsilon^{\alpha}}\right)$ } \newline
		Sample complexity bound \newline
		{\tiny $ \mathcal{O} \left(\frac{H^{\alpha} |\mathcal{S}| \Psi(|\mathcal{A}|, \varepsilon/H, \delta/|\mathcal{S}|)}{\varepsilon^{\alpha}}\right)$ }
		\\
		\tblrowsepa{}
		Preference-based Exploration \& Policy Search (PEPS)
		\newline \citep{xu2020preferencebased}
		& Similar to PPS, instead of simulator using an auxiliary synthetic reward function
		& Same as PPS and stochastic triangle inequality of trajectory comparisons preferences
		& $(\epsilon,\delta)$-PAC for optimal policy based on trajectory comparison feedback
		& Step complexity bound \newline
		{\tiny $ \widetilde{\mathcal{O}} \left(\frac{H^{\alpha+1}|\mathcal{S}|^2 \Psi(|\mathcal{A}|, \varepsilon/H, \delta/|\mathcal{S}|)}{\varepsilon^{\alpha+1}} \right) $ } \newline
		Comparison complexity bound \newline
		{\tiny $\mathcal{O} \left(\frac{H^{\alpha} |\mathcal{S}| \Psi(|\mathcal{A}|, \varepsilon/H, \delta/|\mathcal{S}|)}{\varepsilon^{\alpha}}\right)$ } \\
		\\
		\tblrowsepa{}
		UCBVI-Planning
		\newline \citep{kong2022provably}
		& Optimistic least-squares value iteration, maximum information gain, value iteration based on pessimistic expected value function estimation
		& Binary rewards for state-action pairs based on human response model $f \in \mathcal{F}_H$ with bounded noise $\Delta>0$, compliant and tabular/linear MDP with dimension $d$
		& $(\epsilon,\delta)$-PAC for optimal policy based on binary state-action reward feedback
		& Tabular MDP: \newline
		{\tiny ${\mathcal{O}} \Bigg( \frac{H^4 |\mathcal{S}| |\mathcal{A}| \log\left(\frac{H |\mathcal{S}| |\mathcal{A}|}{\epsilon\delta}\right) }{\epsilon^2} $ \newline %
			$ \hphantom{\mathcal{O} \Big( } +\frac{H^3 |\mathcal{S}|^2 |\mathcal{A}| \log\left(\frac{H |\mathcal{S}| |\mathcal{A}|}{\epsilon\delta}\right)}{\epsilon} \Bigg)$} %
			\newline
			Linear MDP: \newline
		{\tiny ${\mathcal{O}} \left( \frac{|\mathcal{A}|^2d^5  d_{\mathcal{F}_H} H^4\log\left(\frac{H|\mathcal{S}||\mathcal{A}|}{\epsilon\delta\Delta}\right)}{\epsilon^2}\right)$ } \newline
		\\
		\tblrowsepa{}
		Preference-based \& Randomized Least-Squares Value Iteration (PR-LSVI)
		\newline \citep{wu2024making}
		& Least-squares value iteration with perturbed state-action-wise reward model
		& General differentiable link function $\Phi$, linear MDP, linear rewards with $d$-dimensional feature embedding of trajectories
		& Expected regret minimization w.r.t.\ optimal policy and/or low trajectory comparison feedback complexity steered by  $\epsilon\in [0,1]$
		& Expected regret bound: \newline
		{\tiny 	$\widetilde{\mathcal{O}} \Big( \epsilon T d^{1/2} %
			+ \sqrt{T} \cdot d^3 H^{5/2} \gamma
			$ \newline
			$ \hphantom{\mathcal{O} \Big( } + d^{17/2} H^{11/2} \gamma^{3} \Big)$ } \newline %
		Comparison complexity bound: \newline
		{\tiny	$\widetilde{\mathcal{O}} \Big( d^4 {( \kappa + R_{\mathrm{max}})}^2 / \epsilon^2 \Big)$} %
		\newline
		$\kappa = \inf_{x \in [-R_{\mathrm{max}},R_{\mathrm{max}}]} \Phi'(x) $ %
		\\
        \tblrowsepa{}
		   Algorithm for Policy Alignment in Reinforcement Learning (A-PARL)
		\newline \citep{chakraborty2024parl}
		& Iterative bilevel optimization via gradient descent based on an estimated  policy gradient
		& Lipschitz assumptions on the objective function, the reward function, the parametric policy class, and convexity assumptions on the value function
		& Solving the bilevel optimization problem
		& Convergence rate: \newline
		{\tiny $O(1/T)$   }
		\\
		\bottomrule
	\end{tabular}
\end{table}

\paragraph{Online Learning}
The first work dealing with the question of theoretical guarantees for learning an optimal policy from trajectory comparison feedback (see \cref{sec:feedback-classes}) in an online manner is by \citet{novoseller2020dueling}.
It laid the foundation for a paradigm embraced by many subsequent research endeavors: Adapting learning algorithms from the dueling or preference-based bandit literature~\citep{yue2009interactively,sui2018advancements,bengs2021preferencebased} to the underlying situation with additional states.
The preference-based bandit problem can be viewed as a preference-based \gls{RL} problem with one state, so state transition dynamics must be considered accordingly for a fruitful adaptation.
It is worth mentioning that \citet{jain2015learning} used a quite similar idea before for feedback in the form of corrections (see \cref{sec:feedback-classes}) by resorting to the coactive learning setting~\citep{shivaswamy2012online}.
Assuming the existence of a ground-truth context-trajectory scoring function and that the user's feedback is informative, the Preference Perceptron algorithm by \citet{shivaswamy2012online} is used and analyzed in terms of its cumulative regret.

\Citet{novoseller2020dueling} suggest the Dueling Posterior Sampling (DPS), which is an adaptation of the self-sparring algorithm~\citep{sui2017multidueling}.
It takes a Bayesian perspective on the problem and defines a Dirichlet prior on the dynamics and a Gaussian prior on the rewards that are subsequently updated, while the trajectories to be compared by the human labeler are chosen based on their (posterior) probability of being optimal\footnote{The latter probability is assessed by posterior sampling; a commonly used technique in the bandit literature used by so-called Thompson Sampling strategies, see \citet{lattimore2020bandit} for more details.}.
Assuming a linear link function (see \cref{subsec:beyond_boltzmann}) as well as a tabular \gls{MDP}, it is shown that DPS is (i) consistent, i.e., converges in distribution to the optimal policy, and (ii) achieves an asymptotic expected regret bound (see \cref{tab:overview_optimal_policy_learning_online}).

\Citet{xu2020preferencebased} combine dynamic programming and policy search with a black-box preference-based bandit algorithm for each state to design routines that return an almost optimal (a.k.a.\ $\varepsilon$-optimal) policy with high probability\footnote{This is a so-called PAC learning setting~\citep{valiant1984theory} in which the goal of finding the/an optimal object is relaxed to finding a ``good enough'' object, usually specified by some distance measure on the object domain. }.
The first routine, called Preference-based Policy Search (PPS), requires access to a simulator, while the second routine, called Preference-based Exploration and Policy Search (PEPS), gets rid of this requirement by exploring the state space by means of an auxiliary synthetic reward function.
By assuming that the probability of one policy dominating another policy is bounded uniformly over all states from below by a multiplicative of their value function, they show generic upper bounds for both routines on the number of pairwise trajectory comparisons (see \cref{tab:overview_optimal_policy_learning_online}).
If these dominance probabilities have even more structural properties, such as fulfilling stochastic transitivity or stochastic triangle inequality (see \citet{haddenhorst2020generalized,bengs2021preferencebased}), then these upper bounds can be further refined.

A follow-up work by \citet{saha2023dueling} assumes a feature embedding of trajectories that gives rise to a feature embedding of policies and adapts the MaxInP algorithm~\citep{saha2021optimal} for contextual dueling bandits by essentially viewing the policy embeddings as the contexts.
More precisely, assuming a logistic link function (see \cref{subsec:beyond_boltzmann}), confidence sets for the expected scores of the policies are constructed based on the \gls{MLE}, and the two policies with the highest uncertainty in terms of maximal variance are used to sample a trajectory, respectively, to be compared.
In this way, the \gls{LPbRL} is derived and also extended to the case of unknown dynamics by taking the uncertainty regarding the dynamics into account when constructing the confidence sets.
For both cases, i.e., known or unknown dynamics, upper bounds on the regret of \gls{LPbRL} are shown (see \cref{tab:overview_optimal_policy_learning_online}).

In contrast to previous work that all considers tabular \glspl{MDP}, \citet{chen2022humanintheloop} consider the case of a general unknown human feedback model and unknown dynamics each from function classes with a finite Eluder dimension\footnote{Roughly speaking, the Eluder dimension of a function class refers to the number of worst-case errors one must make to identify an unknown function from that class.}~\citep{russo2013eluder}.
They propose and analyze the Preference-based Optimistic Planning (PbOP) algorithm, which essentially follows a similar design as \gls{LPbRL} but uses least-square estimates for the human feedback model and transitions dynamics along with confidence sets based on them.
Moreover, \citet{chen2022humanintheloop} derive lower bounds for the regret of any learning algorithm by reducing the once-per-episode-feedback \gls{RL} problem~\citep{chatterji2021theory} to the \gls{PbRL} problem.
Finally, they extend their analysis to the case of $K$-wise comparisons, where one obtains all $\binom{K}{2}$ pairwise comparisons for $K$ many queried trajectories.
In essence, the regret term coming from the human feedback model class improves by a factor of $\sqrt{K}$.

\Citet{wu2024making} consider a similar learning scenario as \citet{saha2023dueling} but with the additional objective to keep the number of queries of trajectory comparisons low, which is a combination of two competing objectives also studied in the bandit literature~\citep{degenne2019bridging}.
For this purpose, they suggest the Preference-based and Randomized Least-Squares Value Iteration (PR-LSVI) algorithm, which combines least-squares value iteration with a perturbed state-action-based reward model with Gaussian noise for regret minimization; a similar idea to CoLSTIM suggested for contextual dueling bandits~\citep{bengs2022stochastic}.
More specifically, in each time step, the policy maximizing the value function of the perturbed state-action-based reward model and the policy maximizing the latter in the previous time steps are ``played''.
By sampling trajectories of these two policies and computing their expected absolute reward difference (based on the perturbed state-action-based reward model) as a measure of uncertainty, preference feedback for these two trajectories is queried if the uncertainty exceeds a certain threshold.
Moreover, they also suggest a posterior sampling counterpart of this algorithm, the Preference-based Thompson Sampling (PbTS) algorithm, and analyze it in terms of Bayesian quantities.

Recently, \citet{chakraborty2024parl} proposed a bilevel optimization problem that generalizes the standard optimization problem for \gls{RLHF} with trajectory feedback and the negative log-likelihood as a loss function (i.e., \eqref{eq:ml}).
This problem, which they call PARL (Policy Alignment in Reinforcement Learning), is characterized by explicitly taking into account the dependence on the data-collecting process at one level for the optimal policy parameters at the other level.
For this problem, A-PARL is proposed, which is shown to have an $O(1/T)$ convergence rate under specific assumptions, where     $T$ is the number of iterations.

Finally, for the LLM training scenario, there are two perspectives on the problem (MDP vs.\ contextual dueling bandits)  as mentioned in \cref{sec:policy-learning}.
Accordingly, work from the field of contextual preference-based bandits \citep{dudik2015contextual,saha2021optimal,saha2022efficient,bengs2022stochastic,sekhari2023contextual}
can also be viewed as theoretical contributions for the \gls{RLHF} setting for the LLM application.
This view is advocated in particular by \citet{xiong2024iterative}, who view the LLM fine-tuning task as a reverse-KL regularized contextual bandit problem.
Typically the context is chosen externally, but \citet{mehta2023sample} consider the learning variant in which the learning agent chooses the context as well.
This variant is referred to as active contextual dueling bandits.

\begin{table}
	\newcommand{\tblrowsepb}{\arrayrulecolor{gray}\cmidrule(l{0.1em}r{0.1em}){1-5}\arrayrulecolor{black}}
	\small
	\caption{Overview of approaches, their assumptions, goals, and properties for offline policy learning with a data set of size $n$. $ \widetilde{\mathcal{O}}$ is used to hide $\log$-factors. $R_{\mathrm{max}}$ is a bound on the reward.}\label{tab:overview_optimal_policy_learning_offline} %
	\centering
	\hbadness=10000 %
	\hfuzz=\maxdimen{} %
	\begin{tabular}{@{}p{2.4cm}p{2.3cm}p{2.9cm}p{2.4cm}p{5.1cm}@{}}
		\toprule
		\textbf{Algorithm} \newline \textbf{(Reference)}& \textbf{Algorithmic \newline approach} & \textbf{Assumptions} & \textbf{Target(s) and goal(s) of learner} & \textbf{Theoretical guarantee(s)} \\ %
		\midrule
		Pessimistic MLE \newline \citep{zhu2023principled}
		& Greedy policy for pessimistic expected value function estimation
		& Logistic link function, linear reward function for a state-pair feature embedding with some regularity assumptions on weights, known transition dynamics
		& High probability bound for the performance gap based on trajectory-based (and action-based) feedback
		& $\hphantom{text}$ \newline $\mathcal{O} \left(  e^{2 H R_{\mathrm{max}}} \sqrt{ \frac{ d + \log(1/\delta) }{n}} \right)$ %
		\\
		\tblrowsepb{}
		oFfline ReinforcemEnt lEarning with HumAN feeDback (FREEHAND)
		\newline \citep{zhan2024offline}
		& Greedy policy for pessimistic expected value function estimation
		& General differentiable link function $\Phi$, general bounded reward function class $\mathcal{F}_r$  and general transition dynamic class $\mathcal{F}_{\mathbb{P}}$
		& High probability bound for the performance gap based on trajectory-based (and action-based) feedback
		& Transition dynamics: \newline
		1.\ Known \newline
		{\tiny $\mathcal{O} \left(  \sqrt{ \frac{ C_r^2 \kappa^2 \log( \mathcal{N}_{\mathcal{F}_r}(1/N,|\cdot|) /\delta) }{n}} \right)$ }\newline
		2.\ Unknown \newline
		{\tiny $\mathcal{O} \left(  \sqrt{ \frac{ C_r^2 \kappa^2 \log( \mathcal{N}_{\mathcal{F}_r}(1/N,|\cdot|) /\delta) }{n}} \right) + 	\mathcal{O} \left(  R_{\mathrm{max}} \sqrt{ \frac{ C_P^2 \kappa^2 \log( \mathcal{N}_{\mathcal{F}_{\mathbb{P}}}(1/N,|\cdot|) /\delta) }{n}} \right)$ } \newline %
		$\kappa = \inf_{x \in [-R_{\mathrm{max}},R_{\mathrm{max}}]} \Phi'(x) $ \newline %
		$C_r$, $C_P$  reward and transition concentrability coefficient
		\\
		\tblrowsepb{}
		Dynamic-Choice-Pessimistic-Policy-Optimization (DCPPO)
		\newline \citep{li2023reinforcement}
		& Value iteration based on pessimistic expected value function estimation
		& Dynamic discrete choice model, linear MDP, linear reward function for a state-pair feature embedding with some regularity assumptions on weights, known model class entailing the value and reward function of the dynamic discrete choice model
		& High probability bound for the performance gap based on action-based feedback
		& Linear model class: \newline
		{\tiny $\mathcal{O} \left(|\mathcal{A}|d^{3/2} H^2 e^H \sqrt{ \frac{ \log(dHn/\delta) }{n}} \right)$} \newline
		RKHS model class with different eigenvalue decay: \newline
		{\tiny   $ \mathcal{O} \left(\tilde{d} He^H|\mathcal{A}| \sqrt{\mu \log (nR_{\mathrm{max}}H / \delta)}\right) $ $ \mu \text {-finite spectrum, } $ \newline %
		$ \mathcal{O} \left(\tilde{d} He^H |\mathcal{A}|\sqrt{{(\log (n R_{\mathrm{max}}H) / \delta)}^{1+1 / \mu}}\right) $ $ \mu \text {-exponential decay, } $ \newline %
		$ \mathcal{O} \left( \tilde{d} He^H|\mathcal{A}| {(nR_{\mathrm{max}})}^{\kappa^*}  \sqrt{\log (nR_{\mathrm{max}} H / \delta)} \right) $ $ \mu \text {-polynomial decay,}$ \newline $\kappa^*=\frac{d+1}{2(\mu+d)}+\frac{1}{\mu(1-2 \tau)-1}$, \newline $\tilde{d}=$population effective $\cdot$ sampling effective dimension } %
		\\
		\tblrowsepb{}
		LCBVI-Tabular-Offline
		\newline \citep{kong2022provably}
		& Maximum information gain for reward querying, value iteration based on pessimistic expected value function estimation for policy learning
		& Binary rewards for state-action pairs based on human response model with bounded noise, compliant and tabular MDP
		& High probability bound for the performance gap based on binary state-action reward feedback
		& Linear model class: \newline
		{\tiny $\mathcal{O} \Big(  H\sqrt{ |\mathcal{S}| \log( |\mathcal{S}| |\mathcal{A}| H n/\delta)}$ \newline $ \hphantom{ \Big( } \cdot\mathbb{E}_{\pi^*}\left[\sum^H_{h=1}{(N_h(s_h,a_h)+1)}^{-1/2}\right] \Big)$ \newline %
		$N_h$ are numbers of visit time } %
		\\
		\bottomrule
	\end{tabular}
\end{table}

\paragraph{Offline Learning}

\Citet{zhu2023principled} study the performance of a greedy policy trained from a data set consisting of trajectory pairs along with the observed preference that is assumed to be generated by means of a Bradley-Terry model with linear rewards.
For this purpose, different results with respect to the \gls{MLE} of the Bradley-Terry model for different feedback scenarios are derived that are quite of independent interest.
In particular, they show concentration inequalities of the \gls{MLE} for trajectory-based comparison feedback and additionally its asymptotic normality for action-based comparison feedback that also holds for $K$-wise comparisons.
Based on these, it is shown that the greedy policy using the \gls{MLE} in the case of action-based feedback might fail while using a pessimistic \gls{MLE} leads to minimax-rates with respect to the performance gap\footnote{The expected difference between the optimal value function and the value function of the used policy.}.
The latter is also shown to be true in the case of trajectory-based feedback.
Technically, the pessimistic \gls{MLE} is realized by taking the policy that has the largest pessimistic expected value function, i.e., the lowest realization of the value function within a hyperparameter-dependent confidence region around the \gls{MLE}.
Further results of independent interest are the inferred theoretical guarantees for maximum entropy inverse \gls{RL}~\citep{ziebart2008maximum} and action-based inverse \gls{RL} algorithms~\citep{neu2009training}.

The simple model assumptions underlying~\citep{zhu2023principled} were then replaced by more sophisticated assumptions in some subsequent work.
The linear reward assumption has been replaced by more general reward function classes by \citet{zhan2024offline} and \citet{li2023reinforcement}.
In addition, \citet{zhan2024offline} also consider more general unknown human feedback models and construct the confidence regions for the pessimistic approach directly from the log-likelihood function.
The resulting approach, called FREEHAND, is analyzed in terms of its performance gap, for which some problem-dependent coefficients, the per-step, per-trajectory, and transition concentrability coefficient, are introduced.
On the basis of a lower bound, it is shown that the per-trajectory concentrability coefficient should naturally appear in the bound on the performance gap.
Moreover, the concentrability coefficient is shown to be upper bounded by the constant appearing in the special case of linear rewards considered by \citet{zhu2023principled}.
Finally, it is worth mentioning that both trajectory-based and action-based comparison feedback are considered.

In follow-up work, \citet{zhu2024iterative} found overfitting as well as overoptimization issues of the MLE in the Boltzmann model for pairwise comparison feedback.
This can arise in particular if the observations of labels are strongly unbalanced and thus the utilities can become infinite.
To overcome this problem, they propose the Iterative Data Smoothing (IDS) algorithm, which implicitly weights observed labels appropriately by their frequency and their current likelihood.
Note that these issues do not contradict the results shown by \citet{zhu2023principled} as these are based on the assumption of bounded utilities (or rewards).

Assuming a dynamic discrete choice model~\citep{rust1987optimal} underlying the given data set of observed trajectories (without explicitly observed preferences), \citet{li2023reinforcement} suggest the Dynamic-Choice-Pessimistic-Policy-Optimization (DCPPO) algorithm.
It first estimates the reward model using this assumption and then learns a policy in a (pessimistic) value iteration manner from the estimated reward model.
In the case of a linear \gls{MDP} and a known model class that entails both the value and the reward function of the dynamic discrete choice model, DCPPO is analyzed with respect to its performance gap.
This is done for the case of a linear function model class as well as a subset of a reproducing kernel Hilbert space (RKHS) as the model class.

Focusing on the estimation of the weight parameter in the Bradley-Terry model for the action-based feedback under label differential privacy conditions~\citep{dwork2008differential}, \citet{chowdhury2023differentially} analyze two estimation procedures, \gls{MLE} and \gls{SGD}, under similar assumptions as in \citet{zhu2023principled}.
In both cases, the cost of ensuring label differential privacy is a multiplicative factor.

Reward collapse, a term introduced by \citet{song2023reward}, describes the issue when rank-based training methods for \glspl{LLM} lead to the same reward distribution regardless of the prompts used in the final training steps.
The authors show that this occurs because the rank-based approach does not adequately account for prompt-related information.
To address this problem, the authors propose a family of utility functions as well as an optimization method that successfully creates prompt-dependent reward distributions, effectively mitigating the collapse of rewards during training.

\paragraph{Blending Online and Offline Learning}

\Citet{kong2022provably} study the problem of optimal policy learning from critique feedback (see \cref{sec:feedback-classes}), i.e., binary rewards for state-action pairs, with as few queries to the human as possible.
They assume an underlying ground-truth human feedback model
that leads to a positive evaluation for a state-action pair if it exceeds a specific threshold evaluated at that pair.
In addition, the learning process consists of two phases:
First, exploring the environment in an unsupervised manner, and then querying user feedback in an active reward learning phase to learn the human feedback model.
This learning process is again analyzed in two variants:
Either the exploration phase was performed externally, and a data set consisting of trajectories is provided (offline), or this data set is actively collected itself (online).
For both variants, an active learning algorithm is proposed that essentially selects query points (state-action pairs) that provide the most information gain given the points already designated to be queried.
For the online variant, an exploration strategy based on optimistic least-squares value iteration~\citep{jin2020provably} is also introduced for tabular or linear \glspl{MDP}.
In both variants, policy learning is carried out by a pessimistic value iteration with the empirical transitions and the estimated reward function, resulting in UCBVI-Planning (online) and LCBVI-Tabular-Offline (offline).
Under the assumption of bounded noise~\citep{massart2006risk} or low-noise assumption~\citep{korba2017learning,haddenhorst2021testification}, bounds on the performance gap of both algorithms are derived.

The question of the ideal experimental design for \gls{RLHF} is addressed by \citet{zhan2024rewardagnostic}, in particular, how to separate the process of data acquisition (e.g., trajectories to be evaluated) from the process of retrieving human feedback to avoid constantly involving humans in the training loop.
Assuming linear rewards, the Bradley-Terry model and either a transition oracle (e.g., available for tabular or low-rank \glspl{MDP}) or a linear \gls{MDP} they suggest the expeRimental dEsiGn for queryIng huMan prEference (REGIME) algorithm that first samples exploratory trajectories indented to be as informative as possible for learning the reward via \gls{MLE} and then applies a greedy policy based on the reward learned by the latter.
They explicitly show that REGIME requires less human feedback to be queried in order to output an $\epsilon$-optimal policy at the end than the approach by \citet{saha2023dueling}.

\subsection{Preference-Based vs.\ Reward-Based Learning}

There have been some theoretical analyses regarding the question of how far, or if at all, preference-based feedback in the form of trajectory comparisons is more suitable compared to numerical feedback.
\Citet{ji2023provable}
suggest a human rating model for this purpose in the numerical feedback case and analyze the LCB algorithm~\citep{jin2021pessimism} in order to compare it with the pessimistic \gls{MLE}~\citep{zhu2023principled}.
It is shown that under specific assumptions, LCB has a constant performance gap, while the preference-based pessimistic \gls{MLE} under similar assumptions has a similar bound as in \cref{tab:overview_optimal_policy_learning_offline}.

\Citet{wang2023rlhf} provide reduction-based algorithms that can directly use state-of-the-art results in reward-based \gls{RL} for \gls{RLHF} with utility-based and general state-action and trajectory-based comparison feedback.
They show, in general, how theoretical results of the underlying standard \gls{RL} algorithm can be translated to theoretical results for the resulting preference-based \gls{RL} algorithm.
For some special cases, such as \glspl{MDP} with finite Eluder dimension and utility-based preference feedback, the theoretical guarantees are explicitly derived using state-of-the-art \gls{RL} algorithms that are qualitatively similar to explicit preference-based \gls{RL} algorithms.

\subsection{Nash Learning from Human Feedback}

The majority of theoretical works use the modeling of (pairwise comparison) feedback by means of a link function (see \cref{subsec:beyond_boltzmann}).
Even if this often leads to simpler derivations, this modeling has the decisive disadvantage that it imposes transitivity of the human feedback that does not necessarily prevail in reality.
In other words, it is quite possible that preference cycles can occur.
For this reason, there is a new direction in theoretical work that dispenses with parametric modeling of the preference probability similar to \citet{chen2022humanintheloop} but uses it to formulate a new learning objective.
Specifically, the problem is considered from a game theory perspective, where two policies each propose a trajectory that should be highly preferred by the human user.
Thus, the goal is to find a policy that suggests trajectories that are preferred to the trajectories of any other policy, i.e.,  a Nash equilibrium or a von Neumann winner.

This learning variant was first considered by \citet{wang2023rlhf}, who showed that the problem can be reduced to finding restricted Nash equilibria in a multi-agent \gls{RL} problem (based on numerical rewards).
For special situations of the latter problem, wrapper algorithms are proposed that have been shown to find the von Neumann winner with high probability.
The learning problem was recently taken up and analyzed by \citet{munos2024nash} and \citet{ye2024theoretical} in a KL-regularization variant.
While the former considers the online-learning setting assuming a known preference model, the latter considers both online as well as offline learning settings and the preference model belonging to a finite function class.

\section{Applications and Benchmarks}\label{sec:application}

The field of \gls{RLHF} has advanced significantly in the last few years, with increasing interest driven by prominent applications.
First and foremost are applications to large language models, exemplified by ChatGPT~\citep{openai2022introducing}.
This section starts by providing a sample of such applications, showcasing how this technology is being used in fields as varied as robotics, language processing, image generation, and more.
We will also delve into libraries that provide foundational support for \gls{RLHF} research, enabling researchers and practitioners to experiment with and refine a range of approaches.
We then explore a spectrum of benchmarks that have been developed to standardize and simplify the evaluation of new approaches, offering insights into their performance in different settings.
Finally, and closely related to those benchmarks, we will discuss common evaluation practices.

\subsection{Applications}

\gls{RLHF} finds applications across various domains, showcasing its versatility in addressing complex and nuanced tasks.
The most prominent application is ChatGPT~\citep{openai2022introducing}, which is an example of an application in the domain of language models.
Beyond that, however, applications extend across diverse domains such as control tasks, generative models, and recommender systems.
This section provides an overview of notable works applying \gls{RLHF} in different areas.

\paragraph{Control and Interactive Environments}
There is a long history of using control environments as benchmark tasks for \gls{RL}.
In addition to the breadth of available environments, control applications are of particular interest because tasks are often hard to specify.
\Citet{christiano2017deep} demonstrated the effectiveness of \gls{RLHF} in games as well as simulated continuous control tasks, matching the performance of \gls{RL} agents trained on ground-truth rewards with a fraction of the feedback.
Extending to robotics, \citet{ding2023learning} trained a reward model for diverse tasks with a single robot, achieving human-like behavior.
\Citet{kupcsik2018learning} applied \gls{RLHF} for precise robot-to-human handovers.
Similarly, \citet{abramson2022improving} used \gls{RLHF} in the Playhouse simulator, a platform for sensorimotor task training, and \citet{milani2022solving} showcase an application in the context of the MineRL Basalt competition for Minecraft tasks.
Recently, \citet{dong2024aligndiff} use \gls{RLHF} to guide a diffusion-based planning model.

\paragraph{Generative Models in Language and Imaging}
Generative models, i.e., models that generate new data instead of just predicting labels, can be framed as an \gls{RL} setting in which a policy assembles the output through its actions.
In the context of language models, this means that the language model is interpreted as a policy with tokens as actions.
Using this reframing, we can use \gls{RLHF} approaches to fine-tune generative models to produce preferred outputs.
ChatGPT~\citep{openai2022introducing} and GPT-4~\citep{openai2023gpt4} are prime examples of language models fine-tuned using \gls{RLHF}.
These applications build on earlier work, such as by \citet{ouyang2022training}, \citet{ziegler2020finetuning} and \citet{glaese2022improving}.
This method extends to text summarization~\citep{gao2018april, gao2020preferencebased, stiennon2020learning}, dialogue summarization~\citep{chen2023humanintheloop}, and question answering~\citep{nakano2022webgpt,menick2022teaching}.
In image generation, \citet{lee2023aligning} and \citet{xu2023imagereward} demonstrate the use of reward modeling for text-to-image tasks, while \citet{pinto2023tuning} and \citet{kazemi2020preferencebased} explore \gls{RLHF} applications in broader computer vision tasks.
Interestingly, in the context of LLMs, reward learning has also been expressed as density estimation \citep{dumoulin2024density} instead of the supervised approach described in \cref{sec:reward_learning}.

\paragraph{Recommender Systems}
In the context of recommender systems, \citet{xue2023prefrec} have shown the potential of \gls{RLHF} in optimizing for long-term engagement.
Although it is, in principle, possible to algorithmically evaluate policies in this domain, these rewards are sparse.
To combat this, \citet{xue2023prefrec} use \gls{RLHF} to distill sparse, global feedback into a dense reward model.

These diverse applications underscore the adaptability of \gls{RLHF} and its growing importance in various technological domains, paving the way for innovative solutions and enhanced human-computer interactions.

\subsection{Supporting Libraries}

Several libraries have emerged that aim to provide a toolset for implementing and experimenting with \gls{RLHF} and reward learning algorithms, contributing to the ease and efficiency of research and development.
One notable example is the \href{https://github.com/HumanCompatibleAI/imitation}{\texttt{imitation}} library~\citep{gleave2022imitation}.
It encompasses a collection of imitation and reward learning algorithms, including those introduced in the seminal work by \citet{christiano2017deep}.
In the offline realm, \href{https://github.com/pickxiguapi/Clean-Offline-RLHF}{\texttt{Clean-Offline-RLHF}}~\citep{yuan2024unirlhf} provides implementations for offline \gls{RL} algorithms with human feedback.
Two other libraries, \href{https://github.com/Stanford-ILIAD/}{\texttt{APReL}}~\citep{biyik2022aprel} and \href{https://github.com/maegant/POLAR}{\texttt{POLAR}}~\citep{tucker2022polar}, focus on the Bayesian setting.
\Citet{biyik2022aprel} provide a specialized framework for preference-based reward learning with a focus on Bayesian methods.
Meanwhile, \citet{tucker2022polar} introduce a framework designed for Bayesian reward learning from multiple feedback types, including pairwise preferences, in MATLAB.
Finally, in the domain of language model fine-tuning, the \href{https://github.com/CarperAI/trlx}{\texttt{trlX}} library~\citep{castricato2023trlx} offers a toolkit specifically designed for language model training.
It specializes in the fine-tuning of transformer-based language models, treating the language model as the policy in an \gls{RLHF} setup.

Due to the many interacting components and the human element in \gls{RLHF} research, implementing new ideas and running experiments can be quite challenging.
The discussed libraries reduce this challenge and make \gls{RLHF} research more approachable to many researchers.

\subsection{Benchmarks}\label{sec:benchmarks}

Due to the difficulty of reproducible evaluations without a ground-truth objective and with humans in the loop, benchmarks play an important role in advancing and evaluating \gls{RLHF} approaches.
Several benchmarks have been proposed, each focusing on different applications and challenges.

One such benchmark is B-Pref~\citep{lee2021bpref}, which focuses on control tasks with synthetic feedback.
B-Pref aims to provide simulated human feedback that captures some irrationalities, thereby coming closer to evaluation with real human feedback than other approaches.
At the same time, by relying entirely on synthetic feedback, the results are reproducible and cost-effective to generate.
In a similar vein, \citet{freire2020derail} propose a set of environments designed to diagnose common problems in reward learning.
These environments help in identifying and addressing the typical challenges that arise in \gls{RLHF} scenarios.

The offline \gls{RLHF} setting is particularly well-suited for benchmarks, as it allows for the use of static datasets.
\Citet{shin2023benchmarks} evaluate pre-existing offline \gls{RL} benchmarks for their suitability for \gls{RLHF} evaluation, and find that many are ill-suited due to the simplicity of the required reward function.
They do, however, identify a subset of these benchmarks together with their own addition for evaluation.
While \citet{shin2023benchmarks} leverage synthetic rewards, \citet{yuan2024unirlhf} propose a dataset and benchmark for offline \gls{RLHF}, including preference data.
This helps to circumvent the challenges of synthetic feedback and benchmark reproducibility with real feedback.

The MineRL BASALT competition~\citep{shah2021minerl,milani2022solving} gives a more application-driven benchmark with a complex environment.
The competition proposes the challenge of solving tasks defined by natural language descriptions in Minecraft based on human feedback.
Writing hand-engineered reward functions is very challenging in that setting, which makes it a good benchmark for methods based on human feedback.
The competition is method-agnostic in principle, and non-RL approaches such as behavioral cloning are also considered.
While the initial dataset consists of human demonstrations, the competition is agnostic for the feedback type, which may include demonstrations, comparisons, and others.
The final evaluation is performed by humans through pairwise comparisons.

In the domain of language modeling, Truthful QA~\citep{lin2022truthfulqa} serves as a benchmark that measures the truthfulness of models.
Also, in the context of language models, \citet{ramamurthy2023reinforcement} introduce a set of pretrained reward models, learned from human feedback, as benchmarks.
These models serve as reference points for evaluating new \gls{RLHF} techniques against established standards.

For evaluating language reward models specifically, \citet{lambert2024rewardbench} introduce RewardBench, a benchmark designed to test reward model performance across chat, safety, reasoning, and preference learning tasks.
The benchmark includes challenging test cases such as adversarial prompts and edge cases where reward models commonly fail, providing standardized evaluation for both classifier-based and DPO-trained reward models.
RewardBench has become a widely-used evaluation framework with a public leaderboard that tracks performance of submitted reward models.

Together, these benchmarks provide a diverse and comprehensive suite of tests that drive the development and refinement of \gls{RLHF} methods, ensuring they are robust, effective, and capable of handling a wide range of real-world scenarios.

\subsection{Datasets}\label{sec:datasets}

Due to its interactive and online nature, \gls{RLHF} research often does not rely on static datasets.
This is because the feedback is generally collected interactively and depends on the current policy.
When the reward model is not refined iteratively, however, as is common practice for the related settings of LLM fine-tuning and offline \gls{RLHF}, static datasets can be used.
Such a static dataset can significantly simplify the development and evaluation of \gls{RLHF} methods.

Since language model fine-tuning is a popular application of \gls{RLHF} and generally does not iteratively refine the reward model, many datasets have been developed for this purpose.
Particularly notable are
\href{https://github.com/anthropics/hh-rlhf}{\texttt{hh-rlhf}} \citep{bai2022training} %
and \href{https://huggingface.co/datasets/PKU-Alignment/PKU-SafeRLHF}{\texttt{PKU-Safe-RLHF}} \citep{ji2023beavertails}, %
two datasets focusing on harmless and helpful responses,
the OpenAssistant datasets (\href{https://huggingface.co/datasets/OpenAssistant/oasst1}{\texttt{oasst1}}, \href{https://huggingface.co/datasets/OpenAssistant/oasst2}{\texttt{oasst2}})~\citep{kopf2023openassistant}, containing not only response rankings but also ratings on various dimensions, %
the \href{https://huggingface.co/datasets/openai/summarize_from_feedback}{\texttt{summarize\_from\_feedback}} dataset \citep{stiennon2020learning} focusing on preferences over text summaries, %
the Stanford Human Preferences Dataset (\href{https://huggingface.co/datasets/stanfordnlp/SHP}{\texttt{SHP}})~\citep{ethayarajh2022understanding},\footnote{This paper introduces the V-usable information framework for measuring dataset difficulty. The SHP dataset was created after the paper's publication, using its techniques to identify learnable preference pairs from Reddit data.} which is based on Reddit responses,
the WebGPT dataset (\href{https://huggingface.co/datasets/openai/webgpt_comparisons}{\texttt{webgpt\_comparisons}})~\citep{nakano2022webgpt}, focused on long-form question answering %
and the \href{https://huggingface.co/datasets/nvidia/HelpSteer}{\texttt{HelpSteer}}~\citep{wang2024helpsteer} dataset, which is not based on preferences but instead gives ratings on for 4 attributes (helpfulness, correctness, coherence, complexity) for each response. %

Although static datasets are used more rarely in the control setting, some datasets have been developed for offline \gls{RLHF} in this domain.
Concretely, \citet{yuan2024unirlhf} propose the \href{https://uni-rlhf.github.io/}{\texttt{Uni-RLHF}} dataset and a benchmark for offline \gls{RLHF} while
\citet{kim2023preference} publish a \href{https://github.com/csmile-1006/PreferenceTransformer/tree/master/human_label}{dataset} of real human preferences for typical offline \gls{RL} tasks (D4RL, Robosuite).

\subsection{Evaluation}

Evaluating \gls{RLHF} poses unique challenges, particularly in scenarios without precise ground-truth task specifications.
Evaluations generally focus on either the learned policy or the reward model, each shedding light on different aspects of system performance.

\paragraph{Policy Evaluation}
Assessing learned behavior is crucial for the evaluation of an \gls{RLHF} system.
In domains with ground-truth rewards, these can be used for policy evaluation~\citep{christiano2017deep}.
However, many \gls{RLHF} applications lack this clarity.
\Citet{ouyang2022training}, for instance, evaluate the quality of language model responses by having labelers rate the output quality on a test set of prompts, highlighting the significance of human judgement in assessing model outputs.
\Citet{jain2015learning} use direct scores on a Likert scale for evaluations, including self-assessments by trainers and cross-evaluations by others.
\Citet{losey2022physical} extend this with a survey based on scores and free-form participant comments, comparing evaluations based on known true rewards with subjective experiences.
Moreover, \citet{abramson2022improving} employ a multi-stage evaluation scheme that includes scripted probe tasks, a standardized test suite evaluated by humans, and full interactive assessments, demonstrating the need for diverse and thorough evaluation methodologies in \gls{RLHF}.

\paragraph{Reward Model Evaluation}
Direct reward model evaluation complements policy assessment.
While reward model accuracy is a more direct measure of preference-learning success, the ultimate goal is inducing effective policies.
A perfectly accurate reward model is often not necessary to induce a good policy, which is the actual goal of \gls{RLHF}.
Therefore, both evaluation methods are ideally used in combination.

\Citet{jain2015learning} also use a ranking loss method for test sets of trajectories, compared against expert evaluations with known scores.
This approach provides quantitative measures of the reward model's fidelity.
In addition, \citet{wilde2022we} compare parameter-based and reward-based evaluation measures for learned reward functions, identifying strengths and weaknesses in both methods and contributing to a more nuanced understanding of reward model assessment in \gls{RLHF}.
These approaches provide a quantitative measure of the reward model's accuracy in reflecting human preferences and expert judgements.
For a detailed discussion of reward model evaluation, also refer to \cref{sec:rewardmodel-utilitylearning-evaluating}.

Policy- and reward model evaluation both offer insights into the performance of an \gls{RLHF} approach.
Ideally, both measures should be combined to enable quick iteration and give insights into both the preference learning performance as well as the quality of the learned behavior.

\section{Discussion and Conclusion}\label{sec:conclusion}
In this survey, we have provided an overview of the current state of \gls{RLHF}, tracing its evolution from \gls{PbRL} and examining its broad applications across various domains like control, natural language processing, and computer vision.
Given the rapid expansion of this field, our coverage necessarily has limitations in addressing every extension and application in full depth.
In particular, given the rapidly evolving nature of \gls{LLM} research, our coverage of techniques specific to this domain is less exhaustive than our treatment of control and robotics applications.
In this concluding section, we discuss key extensions and alternative approaches, identify open questions, and highlight future research directions.

\subsection{Extensions and Related Approaches}
While our survey focuses on \gls{RLHF} methods that learn reward functions online from human feedback, several promising approaches extend beyond this scope while addressing the same core challenge of learning human-aligned objectives.

An increasingly important alternative is \textbf{\gls{RL} from AI feedback} (RLAIF), which replaces human feedback with evaluations from pre-trained AI systems.
This approach leverages foundation models as preference sources and has demonstrated success across diverse applications: language model fine-tuning~\citep{bai2022constitutional,sun2024salmon}, generating intrinsic motivation for text-based games~\citep{klissarov2024motif}, and learning rewards \citep{wang2024rlvlmf} or coding reward functions \citep{ma2024eureka,xie2024text2reward} for control tasks.
A hybrid approach (\emph{assisted evaluation}) involves AI systems assisting rather than replacing human evaluators, such as generating critiques of model outputs \citep{saunders2022selfcritiquing} or enhancing responses with metadata such as citations \citep{nakano2022webgpt,menick2022teaching}.

While most work on \gls{RLHF} implicitly assumes that tasks can be specified by maximization of expected accumulated scalar rewards, this \emph{reward hypothesis}~\citep{silver2021reward} is under active debate in the \gls{RL} community~\citep{lambert2021reward,vamplew2022scalar,bowling2023settling,skalse2022reward}.
\Citet{skalse2024quantifying} investigate the sensitivity of inverse \gls{RL} to reward misspecification, and recent \gls{RLHF} approaches are beginning to move \textbf{beyond scalar rewards} through more complex objective functions, such as multi-objective frameworks with non-linear aggregation of vector rewards~\citep{qian2023learning}.

Finally, the \gls{RLHF} framework has inspired numerous \textbf{extensions across \gls{RL} domains} that revisit classic \gls{RL} topics through the lens of human feedback.
These extensions span three main categories:
fundamental algorithmic concerns such as
exploration~\citep{liang2022reward},
reward feature learning~\citep{katz2021preferencebased},
hindsight experience replay~\citep{zhang2023wisdom},
and reward shaping~\citep{xiao2020fresh};
advanced learning paradigms including
multi-task learning~\citep{ouyang2022training,abramson2022improving,myers2023active},
continual learning~\citep{zhang2024cppo},
and hierarchical \gls{RL}~\citep{pinsler2018sample};
and specialized domains such as
risk-sensitive~\citep{chen2024provably},
safe~\citep{dai2024safe,cosner2022safetyaware},
and fair \gls{RL}~\citep{siddique2023fairness}.
Additionally, as discussed in \cref{sec:challenges-of-human-labeling}, the intersection of \gls{RLHF} with \gls{HCI} offers rich opportunities for advancing feedback collection mechanisms.
It is crucial to keep human psychology in mind when designing these systems and to learn from other related fields that already studied such issues extensively.

\subsection{Current Limitations}

Despite its successes, \gls{RLHF} faces several limitations that constrain its broader applicability.
\Citet{casper2023open} offer a thorough analysis of open problems and fundamental limitations of \gls{RLHF}; beyond that, we highlight several key limitations in the following.

A fundamental limitation concerns \textbf{feedback quality and interpretation}.
Without assistance during feedback, \gls{RLHF} is limited by the tasks humans can reliably judge~\citep{leike2022why,leike2018scalable,wu2021recursively,christiano2018supervising} -- a problem exemplified in language model fine-tuning, where humans often prefer assertive but incorrect responses \citep{hosking2024human}.
Furthermore, current approaches often fail to learn the actual causes of human feedback, learning spurious correlations instead \citep{tien2022study}.

Current approaches also struggle with \textbf{awareness of gaps in preference understanding}.
The common separation between the policy and the reward model limits how the agent can reason about its knowledge of human preferences.
This limitation may be addressed by tighter integration such as in the cooperative inverse \gls{RL} setting (also called assistance games) \citep{hadfield-menell2016cooperative,shah2021benefits}, where the agent simultaneously learns about human preferences and acts to maximize them, enabling behaviors like conditional planning while awaiting feedback and relevance-aware active learning.
More broadly, the limited ability to reason about preference knowledge represents a fundamental constraint of current \gls{RLHF} approaches.

From a \textbf{theoretical perspective}, a primary challenge lies in relaxing underlying assumptions.
This requires balancing assumptions that are not overly restrictive for practical use cases while maintaining feasibility of theoretical guarantees for computationally efficient algorithms.
Key questions include whether algorithms can avoid actively maintaining a policy space and eliminate sub-optimal policies, or relying on computation oracles.
Recent work, such as by \citet{wang2023rlhf} or \citet{wu2024making}, gives hope that this may be possible.

Beyond technical limitations, \gls{RLHF} raises important \textbf{ethical and social considerations} that require careful attention.
These include fundamental questions about whose preferences should guide alignment,
ensuring ethical data collection practices that respect privacy,
mitigating the amplification of human biases,
and addressing incentives for labeler manipulation \citep{armstrong2020pitfalls,carroll2023characterizing} inherent in current training methodologies.
While some technical solutions are emerging, such as \textbf{privacy-preserving alignment} through differential privacy techniques \citep{wu2024privately}, addressing these concerns comprehensively requires coordinated progress across technical research, ethical frameworks, and governance structures.

\subsection{Conclusion}

Although \gls{RLHF} has significantly contributed to advancements in \glspl{LLM} and other areas of \gls{ML}, it remains a domain in its infancy with many unanswered questions and inherent limitations.
These challenges present opportunities for further advancements in theory and practice, potentially resulting in more robust algorithms that make more efficient use of human feedback.
It remains intriguing to what extent \gls{RLHF} will continue to shape natural language processing, \gls{RL}, robotics, AI alignment, and beyond.

\paragraph{Acknowledgements}
We thank
Tom Bewley,
Andreea Bobu,
Adam Gleave,
Erdem Bıyık,
Yannick Metz,
Peter Stone,
and
Banghua Zhu
for their feedback on earlier versions of this survey.
This publication was supported by LMUexcellent, funded by the Federal Ministry of Education and Research (BMBF) and the Free State of Bavaria under the Excellence Strategy of the Federal Government and the Länder as well as by the Hightech Agenda Bavaria.
This work has also been supported in part by the program of National Natural Science Foundation of China (No. 62176154) and a collaborative project funded by NetEase. EH has received funding from the European Union's Horizon Europe research and innovation programme under the Marie Sklodowska-Curie grant agreement No 101073307.

\bibliography{main}

\appendix

\section{RLHF for Language Models}\label{sec:rlhf-for-llms}

Generative models, particularly \glspl{LLM}, have significantly driven recent interest and advances in \gls{RLHF}.
These models are generally trained on vast amounts of data using self-supervised pretraining objectives that are not aligned with their intended usage.
\Gls{RLHF} can be used for preference-based fine-tuning of these models, enabling them to generate more helpful, honest, and harmless outputs \citep{askell2021general}.
This has highlighted the potential of \gls{RLHF} to enhance the performance and adaptability of intelligent systems across a wide range of applications.

The application of \gls{RLHF} to \glspl{LLM} can be traced back to early works focused on summarization \citep{bohm2019better,ziegler2020finetuning,stiennon2020learning}.
This line of work considers a setting where a human provides feedback on language model outputs, such as text summaries, which is then used to train a reward model that guides the language model towards generating better outputs.
In order to apply \gls{RLHF} techniques to this setting, the language model is treated as an \gls{RL} agent, with the reward model providing the feedback that guides the agent's learning.

The corresponding \gls{RL} problem is characterized by an \gls{MDP} where the agent's actions correspond to generating text tokens, the state is the generated text so far, and the reward is provided by the reward model based on the human feedback.
This is most commonly framed as a single-turn interaction with sparse reward, i.e., the episode consists of generating a single output (e.g., summary or dialogue response) with a single reward at the end evaluating the quality of the output.

The process typically starts with supervised fine-tuning on a small dataset labeled with the desired behavior (behavioral cloning).
Empirical evidence shows that supervised fine-tuning alone is often insufficient, however, motivating preference-based fine-tuning \citep{stiennon2020learning}.
This may be due to the limited size and diversity of the supervised fine-tuning data, leading to overfitting and poor generalization.
Preference data is often more diverse and can be collected more efficiently, helping the model learn a more robust and generalizable objective.
This is further supported by \citet{ramamurthy2023reinforcement} who find that training a reward model from preference data which can then be used for policy optimization can be much more data efficient than supervised fine-tuning.
Using a less capable model for the reward model may further improve generalization, as it is less likely to overfit the training data.
Additionally, \gls{RLHF} can train the model to give responses or refuse the response based on its own knowledge boundary, which is tricky to achieve with supervised fine-tuning since the label needs to depend on the model's state of knowledge \citep{schulman2023reinforcement,zhang2023siren,kadavath2022language}.

Regular RL tasks generally require two types of learning:
Learning which states are rewarding and learning how to reach these states.
This is in many ways a simplified setting compared to traditional \gls{RL} problems, as state transitions are entirely deterministic, effectively requiring no exploration of environment dynamics.
An equivalent formulation, possible due to the deterministic nature of the environment, is to consider the model's entire response as a single action, effectively treating the problem as a bandit problem (further discussed in \cref{sec:rlhf-for-llms-bandit}).
Token-level and response-level interpretations can also be mixed \citep{nakano2022webgpt}, as they lend themselves to different parts of the training process.
\Citet{ramamurthy2023reinforcement} find, for example, that token-level discounting can be beneficial.

\paragraph{Practical challenges}
Note that in practice this assumption of single-turn interactions is frequently violated, with the \gls{LLM} interacting with the human in a dialogue setting, occasionally calling out to external systems, or interacting with the environment in other ways.
This is not in conflict with single-turn training, as the dialogue setting can be seen as a sequence of single-turn interactions.
As a consequence, however, the \gls{LLM} has limited planning capabilities and must rely on the human to provide feedback on each generated output.
It generally cannot ask follow-up questions or engage in multi-turn interactions to clarify the human's feedback, except if explicitly trained or prompted to do so~\citep{zhou2024archer}.
In practice, there are many language-model specific implementation details \citep{ramamurthy2023reinforcement,huang2024implementation,dong2024rlhf,xu2024dpo} (see, e.g., Appendix E in the paper by \citet{nakano2022webgpt} for a concrete example) that are not all covered in this work.

\paragraph{Exploration}
While we previously argued that exploration of \emph{environment dynamics} is not necessary, exploration of the reward landscape is crucial.
To reinforce behavior, it is necessary to explore the space of possible responses.
This is particularly challenging due to the large action space and long time horizon, prohibiting random exploration.
Exploration is commonly encouraged by a KL-divergence regularization term in the policy optimization objective, encouraging the agent to stay close to the pretrained model's behavior, which is generally stochastic due to the pretraining objective \citep{stiennon2020learning,nakano2022webgpt}.
This avoids a phenomenon commonly referred to as entropy collapse, where the \gls{LLM} overfits to the reward model and starts to give low-entropy next-token predictions \citep{nakano2022webgpt}.
Nonetheless, the pretrained model may not explore the space of possible responses sufficiently if the reward model encourages responses that are rarely or never generated by the pretrained model, such as abstaining from answering.

\paragraph{Prompt distribution}
\Gls{RLHF} for \glspl{LLM} can be seen as a form of language-conditioned \gls{RL} \citep{luketina2019survey}, where the user's initial prompt specifies the task.
Training a general-purpose model therefore requires a diverse set of prompts for training.
These prompts are sometimes the same as used for preference learning \citep{rafailov2023direct}, sometimes come from the same pool and may overlap partially \citep{ziegler2020finetuning,stiennon2020learning} and sometimes are chosen to be distinct from that dataset \citep{nakano2022webgpt,ouyang2022training}.

\paragraph{Off-policy training}
When studied in small-scale research contexts, \gls{RLHF} is generally applied in an interleaved fashion, iterating or even parallelizing reward model training and reinforcement learning \citep{christiano2017deep}.
This results in a setting where the trajectories used to train the reward model are generated by a recent policy (i.e., semi-on-policy) and updated during the training process, although often not continuously (i.e., semi-online).
For control applications, changes to the policy can quickly accumulate over longer time-horizons, leading to rapid shifts of the distribution of the observed trajectories and requiring on-policy updates to retain a useful reward model.
In contrast, \glspl{LLM} pose a slightly different setting, as the distribution shift is generally less pronounced due to the single-turn interactions and the diversity of the initial training data.
Additionally, the size of the required dataset and the cost of human feedback often prohibit fully online training, necessitating at least some degree of offline or off-policy training (e.g., by using scraped preferences or prior data) \citep{stiennon2020learning,askell2021general}.
In practice, \gls{RLHF} for \glspl{LLM} is often implemented in a semi-online manner as well, introducing both on-policy and off-policy preferences infrequently \citep{stiennon2020learning}.
While most approaches in the literature include some online or on-policy training \citep{ziegler2020finetuning,wu2021recursively,ouyang2022training}, the complexity of the task often leads to ad-hoc data collection methods \citep{stiennon2020learning,nakano2022webgpt} compared to the more principled approach possible for small-scale research efforts \citep{christiano2017deep}.
Although it may be of less importance than for control applications, many works emphasize that at least some degree of online or on-policy data is important for final performance of \gls{LLM} fine-tuning as well \citep{ziegler2020finetuning,dong2024rlhf,xu2024dpo}.
Note that direct methods such as DPO often lack online data, causing issues due to distributional shift \citep{xu2024dpo}.

\paragraph{Extensions}
Note, however, that \gls{RLHF} for \glspl{LLM} does not \emph{always} operate in this simplified setting.
\Citet{zhou2024archer} extend the \gls{RLHF} setting to multi-turn interactions with \glspl{LLM}, where the agent can ask follow-up questions to clarify the human's feedback.
The \gls{LLM} may additionally be given access to tools such as a web-browser \citep{nakano2022webgpt}, leading to further sources of non-determinism and complexity.
Studied under the names of multi-turn \gls{RLHF} or agentic \glspl{LLM} fine-tuning (with the latter setting generally incorporating actions beyond text output), this setting re-introduces many of the complexities of traditional \gls{RL} settings, such as nondeterministic state transitions, partial observability and delayed rewards \citep{ma2024agentboard}.
Another extension is to introduce dense reward signals \citep{chan2024dense}, which can help to guide the agent more effectively.

\paragraph{Pretraining}
The language model, i.e., the policy, is generally pretrained with a self-supervised objective on a large corpus of text.
This is followed by a step of supervised fine-tuning on a small dataset labeled with the desired behavior.
This can be seen as a form of behavioral cloning, and a similar approach is sometimes used in \gls{RLHF} for control.
Similarly, since judging the quality of a response requires a certain level of natural language understanding that is difficult to acquire from the limited training data available for reward model training, the reward model is generally initialized from a language model pretrained on a large corpus of text.
As a side-effect of this initialization, as the pretraining data likely contains examples of the desired behavior as well as natural language descriptions of human values, the reward model likely already has some knowledge of the desired behavior which then needs to be elicited by fine-tuning with the new objective \citep{yang2024capability}.
It is then additionally often pre-trained from offline preferences, e.g., scraped from the web or mined from implicit expressions of preferences \citep{askell2021general}.
Remaining preferences are then often gathered in a somewhat ad-hoc manner, with the reward model trained on a combination of preferences on responses various intermediate models \citep{nakano2022webgpt}.

\paragraph{Inference-time selection}
The single-turn setting allows for a simple inference-time selection strategy, where the model generates several completions and the best one is selected based on the reward model (\emph{best-of-n-sampling}) \citep{nakano2022webgpt,menick2022teaching}.
This is beneficial in some settings, as it reduces the risk of forgetting during fine-tuning and avoids the cost and time of policy optimization.
It can be viewed as a single-step lookahead search, where the reward model acts as the value function.

\paragraph{Direct methods}
While the separation into reward learning and policy training phases is common and useful in the context of \gls{RLHF} for control,
since it enables the \gls{RL} agent to collect more samples for policy training than for reward model training,
effectively learning about the environment dynamics without excessive human supervision,
this separation is less useful in the context of \gls{LLM} fine-tuning.
\Cref{subsec:direct-methods} discusses direct methods for \gls{RLHF} in more detail.
\Citet{xu2024dpo} finds that PPO, when tuned for the purpose of fine-tuning \glspl{LLM} can often outperform direct methods, a phenomenon that can be partially explained by the distributional shift incurred when optimizing for human preferences, which reward-model based methods account for by using online data and a reward model that may partially generalize the human preferences \citep{lambert2024rewardbench}.

\paragraph{Capabilities vs.\ alignment}
Parts of the vast training data will generally capture this type of interaction, so the model likely already has the capability to perform the task \citep{zhou2023lima,gudibande2024false}, but it needs to be elicited by fine-tuning with the new objective \citep{yang2024capability}.
Note that this even extends to capabilities that could be considered part of the aligning process, such as helpfulness \citep{wang2023selfinstruct}, harmlessness \citep{ganguli2023capacity}, and honesty \citep{kadavath2022language}.

\paragraph{Regularization}
In addition to the reward from the reward model, the policy is often regularized with a KL-divergence term \citep{jaques2017sequence,stiennon2020learning,nakano2022webgpt} to the pretrained policy \citep{ramamurthy2023reinforcement} or early-stopping \citep{nakano2022webgpt} to encourage it to stay close to the behavior of the pretrained model.
This has been found to be crucial for preventing the policy from diverging from the behavior of the pretrained model, which can lead to catastrophic forgetting of the pretrained knowledge \citep{nakano2022webgpt,ramamurthy2023reinforcement}.

\section{Prior Surveys}\label{sec:prior-surveys}

Based on the criteria discussed in \cref{sec:scope-of-the-survey}, here we will first differentiate our survey from other surveys in marginally related subject areas sharing the common theme of human-in-the-loop RL.
Then, we will describe the differences between our survey and previous surveys or survey-like articles that exist within the \gls{RLHF} field.

\subsection{Human-in-the-Loop RL}

\begin{table}
	\newcolumntype{F}{>{\centering\arraybackslash}m{\widthof{\textbf{and Online}}}}%
	\caption{%
		An overview of prior surveys of human-in-the-loop RL.
		\cmark{} indicates that the criterion is a main focus of the survey, \acmark{} indicates that the criterion is partially addressed, while \xmark{} indicates that the criterion is not covered.
	}\label{tbl:hil-survey-overview}
	\begin{tabularx}{\textwidth}{L{}@{\hskip8pt}L{}@{\hskip-3pt}F@{\hskip1pt}F@{\hskip1pt}F@{\hskip1pt}F}
		\toprule
		\textbf{Reference} & \textbf{Topic} & \textbf{Reward Modelling} & \textbf{Human Defined} & \textbf{Interactive and Online} & \textbf{Scalable and Async.} \\
		\midrule
		\citet{wu2022survey} & Human-in-the-loop ML & \xmark{} & \xmark{} & \xmark{} & \xmark{} \\[0.3ex]
		\citet{retzlaff2024humanintheloop} & Human-in-the-loop RL & \acmark{} & \acmark{} & \cmark{} & \acmark{} \\[0.3ex]
		\citet{najar2021reinforcement} & RL with human advice & \acmark{} & \acmark{} & \cmark{} & \acmark{} \\[0.3ex]
		\citet{lin2020review} & Social feedback & \xmark{} & \acmark{} & \cmark{} & \xmark{} \\[0.3ex]
		\citet{poole2024interactive} & RL from brain signals & \xmark{} & \cmark{} & \cmark{} & \xmark{} \\[0.3ex]
		\citet{cruz2020survey} & Interactive RL for HCI & \xmark{} & \xmark{} & \cmark{} & \xmark{} \\[0.3ex]
		\citet{osa2018algorithmic} & Imitation learning & \xmark{} & \cmark{} & \xmark{} & \xmark{} \\[0.3ex]
		\citet{arora2021survey} & Inverse RL & \cmark{} & \cmark{} & \xmark{} & \cmark{} \\[0.3ex]
		\citet{bignold2021conceptual} & Assisted RL & \xmark{} & \xmark{} & \acmark{} & \xmark{} \\[0.3ex]
		\citet{luketina2019survey} & Language-informed RL & \xmark{} & \xmark{} & \xmark{} & \xmark{} \\[0.3ex]
		\citet{zhang2021recent} & Human guidance & \xmark{} & \cmark{} & \cmark{} & \xmark{} \\[0.3ex]
		\citet{ji2023ai} & AI Alignment & \cmark{} & \xmark{} & \cmark{} & \cmark{} \\[0.3ex]
		\citet{liu2023summary} & LLM applications & \xmark{} & \xmark{} & \xmark{} & \xmark{} \\[0.3ex]
		Ours & RLHF & \cmark{} & \cmark{} & \cmark{} & \cmark{} \\
		\bottomrule
	\end{tabularx}
\end{table}

Human participation in \gls{ML}, particularly in guiding machine learners, is a well-studied scenario.
This field, commonly referred to as human-in-the-loop \gls{ML}, can be further divided into subfields based on various criteria, e.g., the ones detailed in \cref{sec:scope-of-the-survey}.
Prior surveys of these subfields are compiled in \cref{tbl:hil-survey-overview} and briefly summarized in the following.

\begin{description}
	\item[Human-in-the-Loop]
	Learning from human feedback falls into the domain of human-in-the-loop \gls{ML}.
	\Citet{wu2022survey} survey human-in-the-loop \gls{ML} in general.
	They also cover some applications of \gls{RLHF} (for \glspl{LLM} in particular) but do not provide a detailed overview.
	\Citet{retzlaff2024humanintheloop} provide a similar overview over human-in-the-loop \gls{RL} in particular, focusing on human involvement in \gls{RL} on a more abstract level than our work and not covering \gls{RLHF} in detail.
	Similarly broad in scope, \citet{najar2021reinforcement} study the setting of \gls{RL} with human advice, which they define as `teaching signals that can be communicated by the teacher to the learning system without executing the task'.
	While this setting subsumes \gls{RLHF}, the broad generality limits the depth to which their survey can cover \gls{RLHF} approaches.

	\item[Interactive RL]
	\Gls{RLHF} can be considered a sub-field of interactive \gls{RL}, which studies \gls{RL} algorithms that learn in interaction with humans.
	This interaction can take the form of feedback defining an objective, resulting in the \gls{RLHF} setting, but can also, e.g., be used to drive exploration or speed up the agent's learning process.

	\Citet{cruz2020survey} survey interactive \gls{RL} from an \gls{HCI} viewpoint, exploring various ways humans can influence \gls{RL} agents, with a particular focus on reward definition based on human feedback, without a predefined environmental reward function.
	Due to the breadth of their survey, they do not cover many works in the area.
	The survey by \citet{lin2020review} centers on interactive \gls{RL} using human social cues, like gestures and spoken language, but does not cover the reward modeling aspect.
	Similarly, the study by \citet{poole2024interactive} examines \gls{RL} with direct feedback from human brain signals, such as through brain-computer interfaces, also not focusing on reward modeling.

	\item[Demonstrations]
	Learning from demonstrations, in the form of behavior cloning~\citep{osa2018algorithmic}
	and inverse \gls{RL}~\citep{arora2021survey}, shares the goal of \gls{RLHF} to learn behavior from human input.
	In contrast to \gls{RLHF}, however, it requires demonstrations of
	the desired behavior instead of feedback, and these demonstrations are usually not provided interactively and online.
	This limits their applications and final performance due to the need for near-optimal demonstrations.
	Nonetheless, imitation and demonstration can be a useful component of an \gls{RLHF} system but are not the main focus of this survey.
	However, we will discuss the intersection between these fields in some parts whenever necessary.

	\item[Assisted RL]
	\citet{bignold2021conceptual} review the field of assisted \gls{RL}, where an agent may receive external information (for example, from a human) that aids it in action selection.
	While updates to the reward function are one of the possible effects of advice in this setting (in addition to action selection or modifications of the agent's internal state), it is usually assumed that an initial reward function is given and the extent of the updates is limited to reward shaping or supplementary reward signals.
	In contrast to \gls{RLHF}, the external information does not define the task but only helps the agent achieve it.
	Closely related to this, \citet{luketina2019survey} survey \gls{RL} assisted by natural language.
	In addition to this assistance setting, they also discuss approaches that infer a language-conditioned reward function.
	However, they discuss this setting rather briefly and use techniques from inverse \gls{RL} and not \gls{RLHF}.

	\item[Guidance]
	In their survey on human guidance, \citet{zhang2021recent} explore various aspects related to \gls{RLHF}.
	Although they touch on aspects such as reward learning, it is not the primary emphasis of their work.
	Instead, their main focus lies on exploring more immediate approaches that do not involve the learning of a reward model.

	\item[AI Alignment]
	\citet{ji2023ai} provide a general overview of AI alignment, i.e., the challenge of aligning the objectives of an intelligent system with those of its human operators.
	This survey covers \gls{RLHF} in some detail.
        As AI alignment is a very broad field, however, the article nevertheless does not go into as much depth on the topic of \gls{RLHF} as we do here.

	\item[Applications]
	\citet{liu2023summary} give an overview of current applications of \gls{RLHF} methods for \glspl{LLM} such as ChatGPT and GPT-4.
	Even though it currently enjoys a lot of attention, it is only one specific application area for \gls{RLHF}.
	Our survey adopts a broader perspective, examining the diverse applications and impact of \gls{RLHF} encompassing application areas beyond \glspl{LLM}.
\end{description}

\subsection{PbRL and RLHF}

\begin{table}
	\renewcommand{\arraystretch}{1.2}
	\newcolumntype{G}{>{\centering\arraybackslash}m{\widthof{\textbf{Comparisons}}}}%
	\caption{An overview of prior \gls{RLHF}-specific surveys and articles with substantial review components. The symbols indicate coverage: \cmark{} indicates that the aspect is addressed, \acmark{} indicates that the aspect is partially addressed, while \xmark{} indicates that the aspect is not covered.}\label{tbl:rlhf-review-overview}
	\begin{tabularx}{\textwidth}{L{}@{\hskip-1pt}G@{\hskip1pt}G@{\hskip1pt}G@{\hskip1pt}G@{\hskip1pt}G}
		\toprule
		\textbf{Reference (Focus)} & \textbf{Beyond Comparisons} & \textbf{Label Collection} & \textbf{RM Training} & \textbf{Theory} & \textbf{App.\ and Benchmarks} \\
		\midrule
		\citet{wirth2017survey} \newline{} (preference-based RL) & \xmark{} & \acmark{} & \acmark{} & \xmark{} & \acmark{} \\
		\citet{abdelkareem2022advances} \newline{} (recent advances of PbRL) & \xmark{} & \xmark{} & \acmark{} & \acmark{} & \acmark{} \\
		\citet{jeon2020rewardrational} \newline{} (feedback modelling) & \cmark{} & \xmark{} & \acmark{} & \xmark{} & \xmark{} \\
		\citet{casper2023open} \newline{} (open issues in RLHF) & \cmark{} & \xmark{} & \cmark{} & \xmark{} & \acmark{} \\
    	\citet{fernandes2023bridging} \newline{} (language generation) & \cmark{} & \cmark{} & \xmark{} & \xmark{} & \acmark{} \\
		\citet{metz2023rlhfblender} \newline{} (feedback types) & \cmark{} & \cmark{} & \xmark{} & \xmark{} & \xmark{} \\
		\citet{yuan2024unirlhf} \newline{} (feedback types) & \cmark{} & \cmark{} & \xmark{} & \xmark{} & \cmark{} \\
		Ours (fundamentals, recent advances, and trends) & \cmark{} & \cmark{} & \cmark{} & \cmark{} & \cmark{} \\
		\bottomrule
	\end{tabularx}
\end{table}
Several previous surveys or survey-like articles are closely related to \gls{RLHF}.
\cref{tbl:rlhf-review-overview} gives a brief overview of how these articles differ from ours, which we will explain in more detail below.

\begin{description}
	\item[Preference-Based RL]
	Previous surveys in the domain of \gls{RLHF} often focus on \gls{PbRL}, where feedback is limited to binary preferences (see \cref{sec:origins-of-rlhf}).
	An illustrative example of this is the survey by \citet{wirth2017survey}, which is a direct precursor to our work.
	In contrast to our work, they concentrate on binary preferences for trajectories and primarily survey methods that learn policies without deriving a reward model.
	Since then, the reward-modeling approach has become dominant, and other approaches have extended \gls{RLHF} to new feedback types.
	\citet{abdelkareem2022advances} give another more recent literature review of \gls{PbRL}.
	While this review focuses on reward modeling and includes some recent work, it is far less comprehensive than our review, as many aspects are only touched upon and partly overlap with those of \citet{wirth2017survey}.

	\item[Feedback Types]
	Although not a survey per se, \citet{jeon2020rewardrational} propose reward-rational implicit choice as a unifying framework to comprehend many previous studies in \gls{PbRL} and \gls{RLHF}.
	To illustrate its generality, they overview different feedback types used in previous work and explain how they fit into their framework.
	The concurrent works by \citet{metz2023rlhfblender} and \citet{yuan2024unirlhf}, which are also not strictly surveys, propose frameworks for studying user interaction and interface design for multiple feedback types.
	As part of their work, they provide a classification of feedback types and a brief overview of \gls{RLHF} approaches.
    \citet{metz2023rlhfblender} have a stronger focus on the feedback interface and on learning from multiple feedback types simultaneously, discussing properties of feedback types and proposing a standard encoding for them.
	On the other hand, \citet{yuan2024unirlhf} also include an offline \gls{RLHF} benchmark and have a stronger focus on the reward learning aspect, focusing on the entire learning pipeline.
	Nevertheless, many facets of \gls{RLHF} are not addressed in those studies, as they are not primarily survey articles.
	Our survey has a broader scope and therefore provides more extensive coverage, going beyond the study of feedback types and discussing more recent work.

	\item[Domain-Specific]
	\Citet{fernandes2023bridging} focuses on human feedback for language generation.
	As a result of their focus, their survey is less comprehensive than this work but discusses some language-specific aspects that do not fall into our scope, such as using feedback models at generation time.

	\item[Open Problems]
	\citet{casper2023open} provide a detailed overview of the open questions and limitations of \gls{RLHF} with a particular focus on aspects of security, governance, and transparency.
	In their article, reward modeling is also covered, as is human feedback, which goes beyond preference comparisons, but other aspects, such as theoretical approaches or an overview of existing benchmarks, are not included.
	Thus, it can be seen as a supplementary article that is ideal for further reading once being familiarized with the topic through our survey.
\end{description}

\end{document}